\documentclass{article}

\usepackage{microtype}
\usepackage{graphicx}
\usepackage{subcaption}
\usepackage{booktabs} 

\usepackage{hyperref}

\usepackage[preprint]{icml2026}

\usepackage{amsmath}
\usepackage{amssymb}
\usepackage{mathtools}
\usepackage{amsthm}

\usepackage[capitalize,noabbrev]{cleveref}

\theoremstyle{plain}
\newtheorem{theorem}{Theorem}[section]
\newtheorem{proposition}[theorem]{Proposition}
\newtheorem{lemma}[theorem]{Lemma}

\theoremstyle{definition}
\newtheorem{definition}[theorem]{Definition}
\newtheorem{assumption}[theorem]{Assumption}
\theoremstyle{remark}
\newtheorem{remark}[theorem]{Remark}

\usepackage[textsize=tiny]{todonotes}

\icmltitlerunning{}

\begin{document}

\twocolumn[
  \icmltitle{FIRM: Federated In-client Regularized Multi-objective Alignment\\ for Large Language Models}



  \icmlsetsymbol{equal}{*}

  \begin{icmlauthorlist}
    \icmlauthor{Fatemeh Nourzad}{yyy}
    \icmlauthor{Amirhossein Roknilamouki}{yyy}
   \icmlauthor{Eylem Ekici}{yyy}
    \icmlauthor{Jia Liu}{yyy}
    \icmlauthor{Ness Shroff}{yyy,comp}
  \end{icmlauthorlist}

  \icmlaffiliation{yyy}{Department of Electrical and Computer Engineering, The Ohio State University, Columbus, OH, USA}
  \icmlaffiliation{comp}{Department of Computer Science and Engineering, The Ohio State University, Columbus, OH, USA}

  \icmlcorrespondingauthor{Fatemeh Nourzad}{nourzad.1@osu.edu}
  \icmlcorrespondingauthor{Amirhossein Roknilamouki}{roknilamouki.1@osu.edu}

  \icmlkeywords{Machine Learning, ICML}

  \vskip 0.3in]



\printAffiliationsAndNotice{}  

\begin{abstract}
  Aligning Large Language Models (LLMs) with human values often involves balancing multiple, conflicting objectives such as helpfulness and harmlessness. Training these models is computationally intensive, and centralizing the process raises significant data privacy concerns. Federated Learning (FL) offers a compelling alternative, but existing Federated Multi-Objective Optimization (FMOO) methods face severe communication bottlenecks as their reliance on transmitting multiple gradients to a server is unscalable for large models. We introduce FIRM (Federated In-client Regularized Multi-objective alignment), a novel algorithm that achieves both client disagreement drift mitigation and communication efficiency. In FIRM, each client locally solves a regularized multi-objective optimization problem. By directly mitigating client disagreement drift through in-client regularization, our method eliminates the need for the multi-gradient transmissions common in prior works. Consequently, clients need only to transmit a single set of adapted parameters, maintaining high communication efficiency. We prove that our algorithm converges to Pareto-stationary points and, to our knowledge, provide the first finite-time convergence guarantees for this federated multi-objective alignment setting. Empirically, we show that FIRM leads to smoother training dynamics, reduced client disagreement drift, and improved reward trade-offs compared to baselines. We further propose a method to incorporate a preference over the objectives and report empirical Pareto plots, demonstrating that FIRM can smoothly adapt trade-offs between objectives in response to specified preferences.
\end{abstract}
\section{Introduction}
Large Language Models (LLMs) have become indispensable in applications ranging from digital assistants to scientific discovery \citep{snell2023offline, pyatkin2023clarifydelphi}. Yet their deployment is inseparable from the question of alignment: ensuring that model behavior is consistent with human values such as helpfulness and harmlessness \citep{ouyang2022training}. Crucially, alignment is not a single-objective problem but a balancing act across multiple, often conflicting, goals. For example, maximizing helpfulness can come at the expense of safety, while enforcing strict harmlessness may lead to evasive or uninformative answers. Designing scalable methods to navigate such trade-offs has therefore emerged as a central challenge in the responsible development of LLMs.

The dominant alignment paradigm, Reinforcement Learning from Human Feedback (RLHF) and its recent variants like Direct Preference Optimization (DPO) and Group Relative Policy Optimization (GRPO) \citep{rafailov2023direct,shao2024deepseekmath}, optimizes policies from preference data. While successful, these methods are fundamentally centralized, demanding vast labeled datasets and large-scale, proprietary infrastructure. This centralization creates significant barriers to entry, limiting broader research participation and raising pressing data privacy concerns. Federated Learning (FL) offers a compelling alternative \citep{mcmahan2017communication}. By training models locally on client data and aggregating only model updates, FL presents a path toward democratizing alignment research while inherently preserving privacy. However, applying FL to multi-objective LLM alignment introduces a new, fundamental design question: {\centering
\textit{Should the resolution of conflicting objectives happen at the central server or locally at each client?}}  

 \begin{figure*}[t]
\centering
    \begin{subfigure}[b]{0.31\textwidth}
    \label{fig:intro_1}
        \centering
        \includegraphics[width=\linewidth]{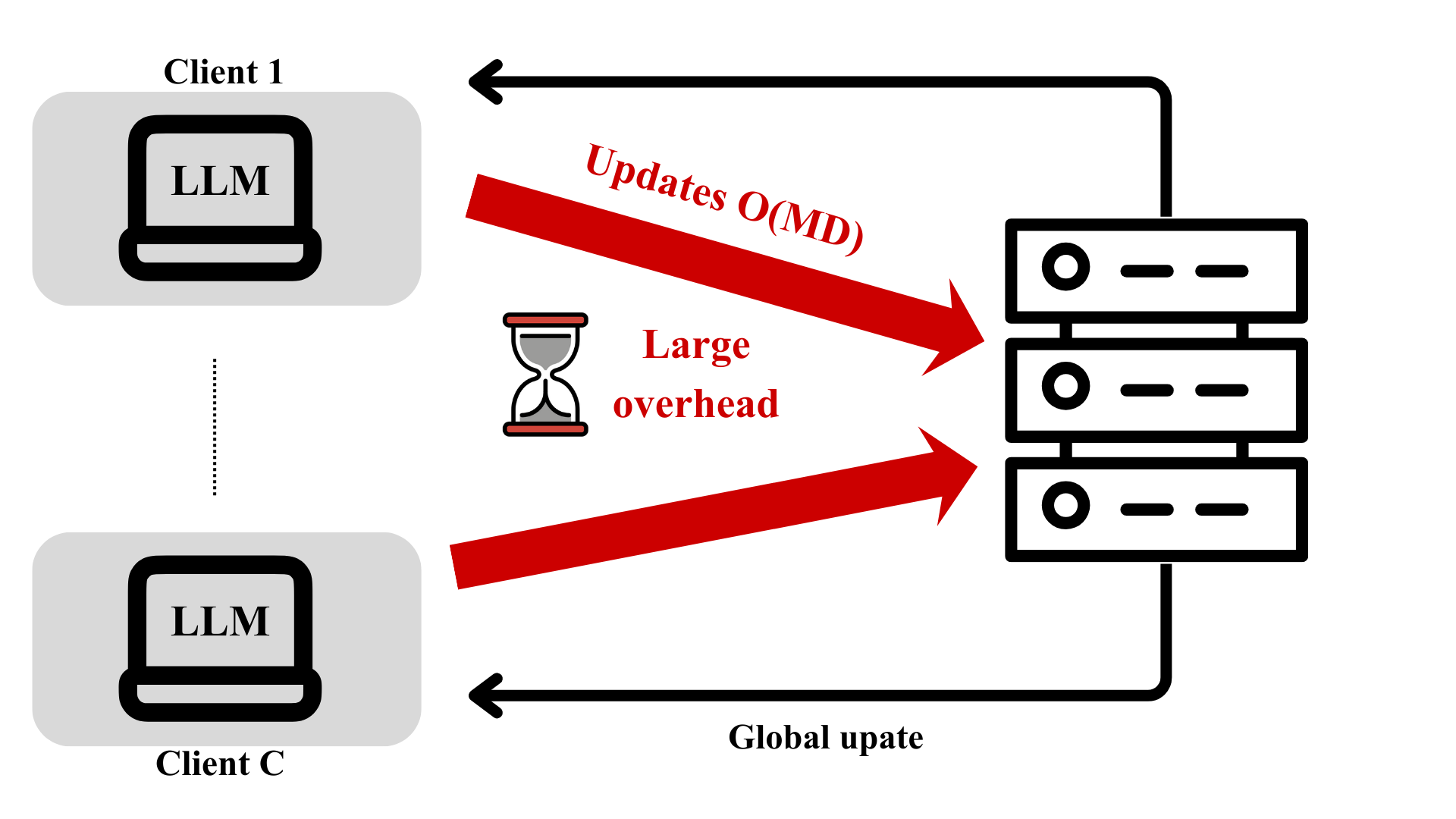}
        \caption{Server Centric}
    \end{subfigure}
    \hfill
    \begin{subfigure}[b]{0.33\textwidth}
    \label{fig:intro_2}
        \centering
        \includegraphics[width=\linewidth]{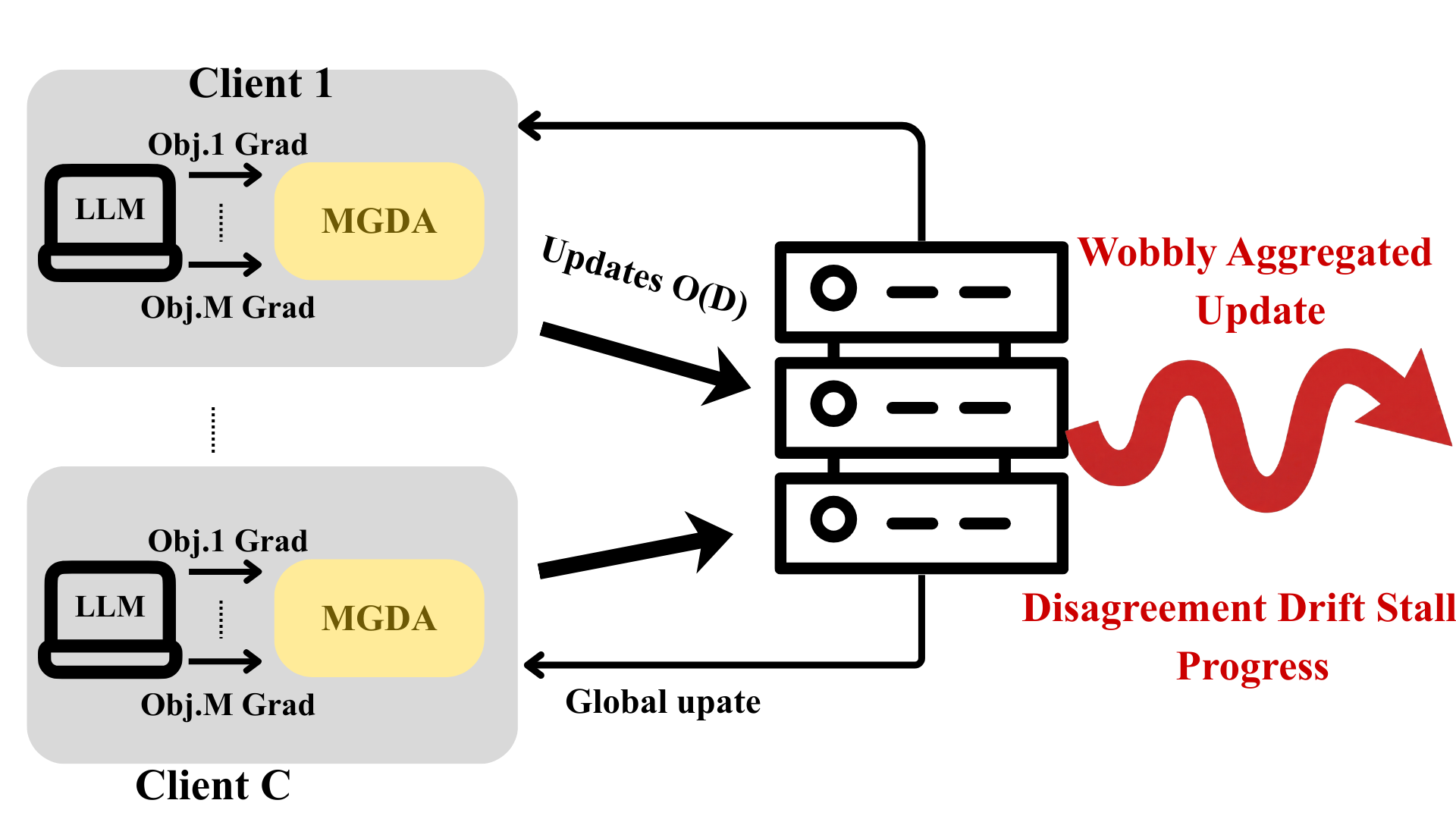}
        \caption{Naive Client-Centric}
    \end{subfigure}
    \hfill
    \begin{subfigure}[b]{0.33\textwidth}
    \label{fig:intro_3}
        \centering
        \includegraphics[width=\linewidth]{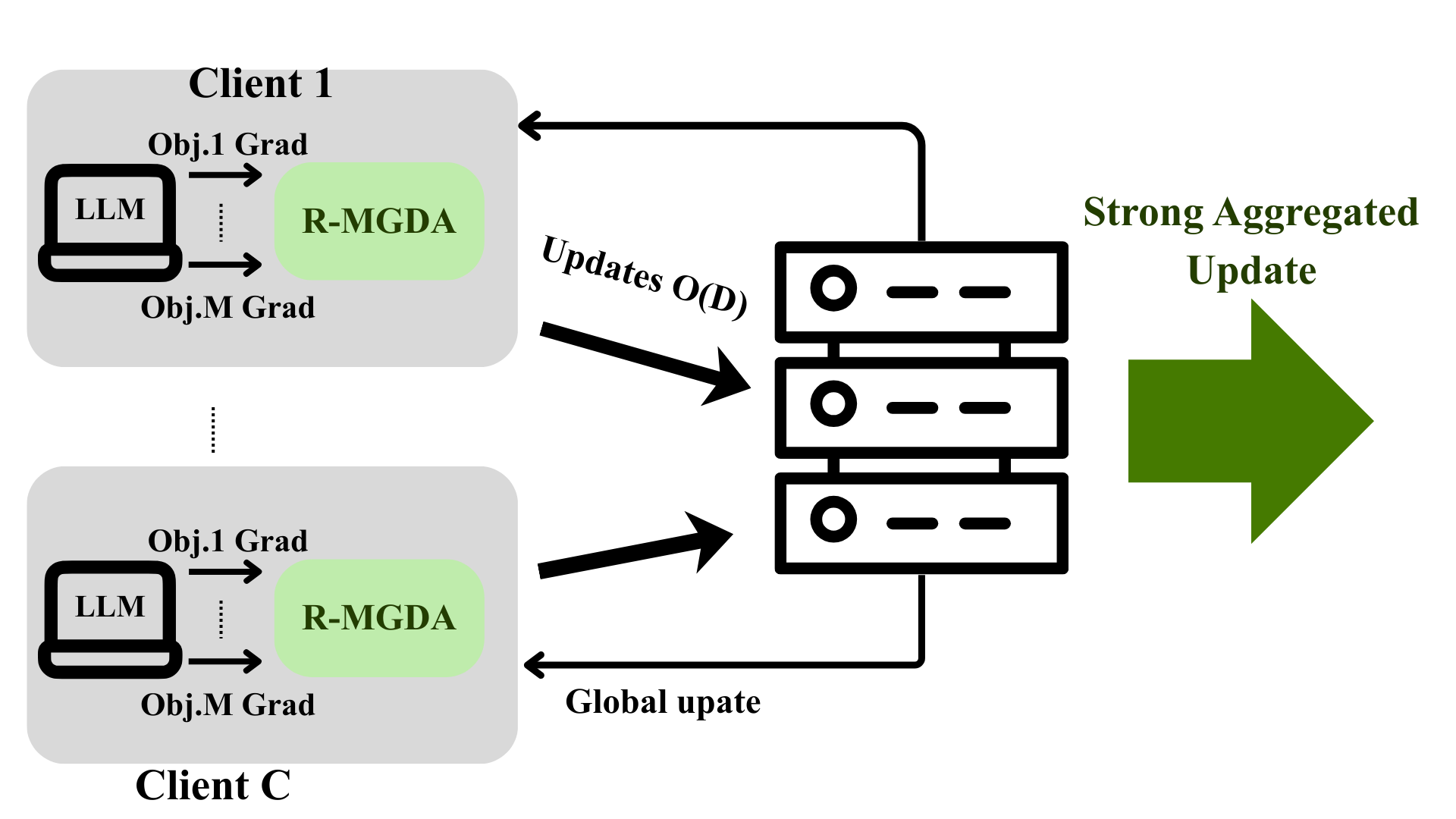}
        \caption{FIRM}
    \end{subfigure}
    \caption{
       \textbf{Comparison of federated multi-objective alignment paradigms.}
(a) \textbf{Server-centric:} incurs prohibitive communication overhead
$\mathcal{O}(CMd)$ by transmitting all objective gradients to the server.
(b) \textbf{Naive client-centric:} reduces communication but suffers from
\emph{disagreement drift}, where inconsistent local updates yield a wobbly
aggregated direction and stall learning.
(c) \textbf{FIRM:} applies local regularization  to align
gradient updates across clients, producing a strong aggregated update
that enables stable and efficient alignment.
    }
    \label{fig:Intro3}
\end{figure*}

Recent work in federated multi-objective supervised learning has explored the server-centric approach to conflict resolution. A naive implementation requires each client to transmit gradients for all $M$ objectives to a central server, which then finds a globally consistent update direction \citep{yang2023federated}. This design is practically untenable for modern LLMs, as it inflates communication costs by a factor of \(M\), resulting in a prohibitive overhead of \(\mathcal{O}(CMd)\), where \(C\) denotes the number of clients and \(d\) the model dimension (See Figure~\ref{fig:Intro3}). Recognizing this bottleneck, other works have proposed using gradient compression before transmission \citep{askin2024federated}. However, this remedy introduces significant new overheads: convergence becomes fundamentally limited by the quality of the compression, an extra communication round-trip is required in every step, and the compression itself can be computationally prohibitive for models with billions of parameters. Thus, whether naive or compressed, the server-centric paradigm remains a non-starter for efficient, large-scale LLM alignment.

As our \textbf{first contribution}, we shift the paradigm to a communication-efficient client-centric approach. We introduce \textbf{Federated In-client Regularized Multi-objective alignment (FIRM)}, 
the first framework, to our knowledge, for federated multi-objective LLM alignment.  In this framework, each client computes gradients for all $M$ objectives using Proximal Policy Optimization (PPO) \citep{schulman2017proximal}, and then resolves them locally into a single update direction  by solving a \textit{Multiple-Gradient Descent Algorithm (MGDA)} subproblem \citep{desideri2012multiple}. 
The central server's role is simply to aggregate the updated client parameters via FedAvg \citep{mcmahan2017communication}, which reduces communication costs to a practical $\mathcal{O}(Cd)$.  However, in deploying this method, we identified a critical instability inherent to naive client-side MGDA resolution, which we term  \textbf{multi-objective disagreement drift}. Because the solution to the local 
MGDA subproblem is highly sensitive to input gradients, minor stochastic variations across clients  cause their computed update directions to diverge significantly. When aggregated at the server,  these conflicting updates impede learning and prevent effective alignment (see  Section~\ref{Sect:algDescription} and Remark~\ref{Remark:DisagreementDrift_Intro_1} in Section~\ref{sec:Convergence_AISTAT_1}).

As our \textbf{main theoretical contribution}, we show that this newly identified drift can be provably controlled with a simple, theoretically-grounded modification. We equip FIRM with a lightweight \textbf{regularization term} in each client’s local MGDA subproblem (see Figure~\ref{fig:Intro3}). Our convergence analysis  provides the \textit{first guarantee for a federated multi-objective actor-critic algorithm}, ensuring FIRM  converges to a Pareto stationary point. Crucially, our theory formally characterizes the 
disagreement drift with a novel error term, $\mathcal{O}\!\left(\frac{\sqrt{M^3}}{\beta \sqrt{B}} \alpha K\right)$, 
where $K$ is the number of local update steps, $B$ is the batch size, 
$\alpha$ is the learning rate, and $\beta$ is the regularization parameter.  This bound reveals a key insight:  the drift is explicitly controlled by the combination of the 
\textbf{regularization} and the \textbf{batch size}. This insight enables us to resolve drift locally at each client, fully eliminating costly server-side conflict resolution.

We provide extensive empirical evidence on aligning a large language model under multiple reward objectives, demonstrating \textsc{FIRM}'s practical power.  Our experiments confirm that the unregularized baseline suffers from \textbf{highly unstable learning dynamics and degraded reward outcomes},  whereas FIRM achieves stable training and superior alignment results (See Section~\ref{sec:experiments}). Furthermore, we conduct detailed  ablation studies to characterize the effect of regularization, confirming that each design choice is essential for the algorithm's performance. Interestingly, our investigation revealed an additional capability: beyond stabilization, the regularization can be used as a preference vector. By adjusting this vector, practitioners can encourage the model to favor one objective over another, resulting in final models with different trade-offs.

In summary, our work illustrates how theoretical analysis can inspire simple, elegant solutions to 
critical challenges in modern AI systems. By identifying and solving the multi-objective disagreement 
drift, our simple regularization not only makes federated alignment practical but also unlocks a novel mechanism for injecting preferences into the MGDA framework---a capability it traditionally lacks. 
This highlights a powerful synergy between theory and practice, yielding a method that significantly improves the performance of AI systems while reducing the cost of their design and implementation.

\paragraph{Other related works} We have provided a comprehensive literature review in Appendix~B.


\section{Problem Formulation}
\label{Sect:ProblemForm}

We formulate the federated alignment problem as a \emph{Federated Multi-Objective Reinforcement Learning} (FedMORL) problem. A central server coordinates with \(C\) clients, each of which optimizes its local policy \(\pi_{\theta^c}\) (the LLM). 

\paragraph{1) Federated Workflow.} 
In each round, client \(c\) samples prompts from its local dataset \(D_c\), generates responses with its local policy \(\pi_{\theta^c}\), and obtains an \(M\)-dimensional reward vector from the reward models (e.g., helpfulness and harmlessness). The client then applies local RL updates to its policy parameters. The server aggregates the updated parameters across clients via FedAvg. This decentralized setup respects data privacy while leveraging distributed resources for alignment.

\begin{definition}[Client MOMDP]
For each client \(c\), the environment is modeled as a Multi-Objective Markov Decision Process (MOMDP) \(
\mathcal{M}_c = (\mathcal{S}, \mathcal{A}, P_c, \mathbf{r}_c),
\)
where \(\mathcal{S}\) is the state space, \(\mathcal{A}\) is the action space, \(P_c\) is the client-specific state transition kernel, and \(\mathbf{r}_c \in \mathbb{R}^M\) is an \(M\)-dimensional reward vector corresponding to the alignment objectives.
\end{definition}

In our theoretical analysis, we consider the general setting where clients may have heterogeneous transition dynamics ($P_c$) and reward functions ($\mathbf{r}_c$) (See Section~\ref{sec:theory_setup_ICML}). We emphasize, however, that multi-objective disagreement drift is a distinct phenomenon arising purely from stochastic noise, persisting even in homogeneous settings. By addressing this fundamental problem, our work establishes a critical foundation for robust and scalable alignment in complex, heterogeneous environments.

\paragraph{2) Global Objective.} The aim of the server is to obtain a global policy \(\pi_\theta\), parameterized by \(\theta \in \mathbb{R}^d\), that achieves good performance across all objectives and clients. Formally, we define the global vector objective as \( 
    \mathbf{J}(\theta) \triangleq \frac{1}{C} \sum_{c=1}^C \mathbf{J}^c(\theta),\)
where \(\mathbf{J}^c(\theta) = [J^{1,c}(\theta), \dots, J^{M,c}(\theta)]^\top\) denotes the vector of local returns for client \(c\). Each entry is defined as the discounted cumulative reward:\[ J^{i,c}(\theta) := \mathbb{E}_{\pi_\theta, P_c}\!\left[\sum_{t=1}^\infty \gamma^t r_t^{i,c}\right], \quad \gamma \in (0,1).\]

\paragraph{3) Performance Metric.} 
Since the $M$ objectives in $J(\theta)$ generally conflict, no single policy can maximize them all simultaneously. 
The standard goal is to identify solutions that balance objectives through \emph{Pareto optimality}. 

\begin{definition}[Pareto Optimality] 
A policy $\pi_\theta$ Pareto-dominates $\pi_{\theta'}$ if 
$J(\theta) \geq J(\theta')$ element-wise and $J(\theta) \neq J(\theta')$. 
A policy $\pi_{\theta^{*}}$ is Pareto-optimal if no other policy Pareto-dominates it. 
\end{definition}

Finding Pareto-optimal policies in non-convex settings such as LLM alignment is NP-hard. Thus, we instead target \emph{Pareto stationarity}, a first-order necessary condition for Pareto optimality \citep{desideri2012multiple, zhou2024finite}. 

\begin{definition}[$\epsilon$-Pareto Stationarity] 
A policy $\pi_\theta$ is $\epsilon$-Pareto stationary if there exists a weight vector 
$\lambda \in \Delta^{M}$ such that 
\(\min_{\lambda \in \Delta^{M}} \|\nabla_{\theta} J(\theta)\lambda\|_2^2 \leq \epsilon ,
\) where $\Delta^{M}$ is the probability simplex. 
\end{definition}

Our objective is to design a federated algorithm that efficiently converges to an $\epsilon$-Pareto stationary point.

\section{FIRM Algorithm for LLM Alignment}
\label{Sect:algDescription}

In this section, we present \textbf{FIRM} (\textbf{F}ederated \textbf{I}n-client \textbf{R}egularized \textbf{M}ulti-objective alignment), a framework designed to make multi-objective LLM alignment scalable, private, and communication-efficient. FIRM fundamentally re-architects the alignment process by enforcing conflict resolution at the edge (client-side) rather than the center. We build upon the standard Federated Averaging (FedAvg) protocol, where $C$ clients collaborate to fine-tune a global policy $\pi_\theta$. 

The algorithm proceeds in communication rounds $t = 1, \dots, T$. At the start of each round, the server broadcasts the global parameters $\theta_t$ to all clients. Each client $c$ initializes its local model $\theta^c_{t,0} \leftarrow \theta_t$ and performs $K$ local update steps. In a given step $k$, the client samples a batch of prompts $B$, generates responses, and evaluates them against $M$ distinct reward models (e.g., Helpfulness and Harmlessness). Standard multi-objective approaches would require transmitting $M$ separate gradients to the server to resolve conflicts. \textsc{FIRM} eliminates this bottleneck. The client first computes the $M$ independent stochastic policy gradients
$\{ g_{t}^{j,c} \}_{j=1}^M$ using Proximal Policy Optimization (PPO) \citep{schulman2017proximal}. Then, rather than sending these conflicting gradients to the server, the client immediately resolves them into a single consensus direction $g_t^c$ by solving a local \emph{Regularized MGDA} subproblem:
\begin{equation}
\lambda^{*,c}_t
=
\underset{\lambda \in \Delta_M}{\arg\min}
\Big(
\underbrace{
\big\|
\sum_{j=1}^M \lambda_j g_t^{j,c}
\big\|_2^2
}_{\mathcal{T}_1:\ \text{MGDA}}
+
\underbrace{
\beta \| \lambda \|_2^2
}_{\mathcal{T}_2:\ \text{Regularization}}
\Big),
\label{eq:mgda}
\end{equation}

where $\Delta_M$ denotes the probability simplex and $\beta > 0$ is a regularization hyperparameter. The client updates its local adapters using the weighted direction \(
g_t^c = \sum_{j=1}^M \lambda_t^{*,j,c} g_t^{j,c}.
\) After $K$ local steps, the clients transmit only their final parameters to the server for aggregation via FedAvg, maintaining a communication cost of $\mathcal{O}(C d)$. The stability and efficiency of this process rely on how clients solve the local MGDA problem. Next, we discuss the role of this regularized formulation in the overall framework.



\begin{algorithm}[tb]
   \caption{Federated In-client Regularized Multi-objective alignment (FIRM)}
   \label{alg:firm}
\begin{algorithmic}
   \STATE {\bfseries Input:} Number of clients \(C\),  batch size \(B\), rounds \(T\), local steps \(K\), learning rate \(\alpha\), MGDA regularization \(\beta\).
   \STATE Initialize global policy parameters \(\theta_0\).
   \FOR{\(t=0, 1, \dots, T-1\)}
   \STATE Server broadcasts \(\theta_t\) to all clients.
   \FOR{each client \(c \in \{1, \dots, C\}\) {\bfseries in parallel}}
   \STATE Initialize local model \(\theta_t^c \leftarrow \theta_t\).
   \FOR{\(k=0, 1, \dots, K-1\)}
   \STATE Sample a batch of \(B\) prompts  and generate responses using \(\pi_{\theta_{t,k}^c}\).
   \STATE Obtain \(M\)-dimensional reward vectors for each response from the reward models.
   \STATE For each objective \(j \in [M]\), compute PPO gradient \(g_t^{j,c}(\theta_{t,k}^c)\).
   \STATE Solve for local consensus weights \(\lambda_t^{*,c}\) using Eq. \ref{eq:mgda}.
   \STATE Combine gradients to form a single direction: \(g_t^c \leftarrow \sum_{j=1}^M \lambda_t^{j,c} g_t^{j,c}\).
   \STATE Update local policy parameters: \(\theta_{t,k+1}^c \leftarrow \theta_{t,k}^c - \alpha g_t^c\).
   \ENDFOR
   \STATE Client \(c\) sends final local model \(\theta_{t+1}^c \leftarrow \theta_{t,K}^c\) to the server.
   \ENDFOR
   \STATE Server aggregates the models: \(\theta_{t+1} \leftarrow \frac{1}{C} \sum_{c=1}^C \theta_{t+1}^c\).
   \ENDFOR
\end{algorithmic}
\end{algorithm}

\paragraph{2) Regularized MGDA:} We now explain why both terms \(\mathcal{T}_1\) and \(\mathcal{T}_2\) are needed in Eq.~\eqref{eq:mgda}. The first term, \(\mathcal{T}_1\), corresponds to the MGDA \citep{desideri2012multiple}. MGDA seeks a convex combination of objective gradients that defines a common descent direction. In a federated setup, however, using only \(\mathcal{T}_1\) is not sufficient. Without \(\mathcal{T}_2\), client updates can drift apart. To see this, consider rewriting Eq.~\eqref{eq:mgda} as
\begin{equation}
\lambda^* \;\in\; \arg\min_{\lambda \in \Delta_M}\; \lambda^\top (G+ \frac{\beta}{2} I) \lambda,
\label{eq:mgda-regularized}
\end{equation}
where \(I \in \mathbb{R}^{M\times M}\) is an identity matrix and \(G \in \mathbb{R}^{M\times M}\) is the Gram matrix with entries \(G_{ij} = \langle g_i, g_j \rangle\). Without the \(\mathcal{T}_2\) regularizer (\(\beta=0\)), correlated objective rewards can make \(G\) ill-conditioned or singular.  In this case, small variations in the gradients can cause large swings in the solution \(\lambda^*\). Across clients, such sensitivity means that sampling noise produces very different weights \(\lambda_t^{*,c}\), which in turn lead to highly noisy and inconsistent descent directions. The result is client drift, where local models diverge and FedAvg aggregates updates that are poorly aligned. We refer to this new source of drift as \textbf{multi-objective disagreement drift}, a phenomenon that fundamentally distinguishes the federated multi-objective setting from its single-objective counterpart. Adding \(\mathcal{T}_2\) with \(\beta > 0\) resolves this issue. The modified Gram matrix \(G + \frac{\beta}{2} I\) is positive definite, with improved condition number. This makes the subproblem strongly convex and ensures that solutions are less sensitive to gradient noise. In practice, this stabilizes local updates and keeps client models closer together, improving server-side aggregation.

\paragraph{3) Extending Regularization to Encode User Preferences:} Beyond stabilization, FIRM’s regularization can also encode  preferences. To incorporate  preferences, we generalize the local MGDA subproblem by  replacing the uniform regularizer $\tfrac{\beta}{2} I$ in Equation~\ref{eq:mgda-regularized}  with a diagonal weighting matrix $\text{Diag}(\mathbf{p}^{-1})$, where  $\mathbf{p} = [p_{1}, \ldots, p_{M}]$ is a vector of positive preference weights.  The full expression becomes:
\begin{equation}
    \lambda^{*} \in \arg\min_{\lambda \in \Delta_{M}} 
    \lambda^{\top} \left(G + \text{Diag}(\mathbf{p}^{-1}) \right) \lambda ,
    \label{eq:mgda-preference}
\end{equation}
A higher preference $p_{j}$ for an objective $j$ reduces its penalty term $1/p_{j}$,  encouraging the optimizer to assign a larger weight $\lambda_{j}$ to that objective and  steer the descent direction toward its gradient. By varying the preference vector $\mathbf{p}$,  FIRM can trace different trade-offs.
\section{Theoretical Analaysis}
\subsection{Theoretical Setup and Assumptions}
\label{sec:theory_setup_ICML}

To enable a tractable convergence analysis, we analyze a variant of FIRM, which we refer to as \emph{Theoretical-FIRM (TFIRM)}, where the PPO update is replaced by a foundational actor-critic (AC) framework.  This substitution allows us to isolate the core federated and multi-objective challenges. 
In this AC setting, each client's local update proceeds in two steps. 
First, in an inner loop, the \textbf{critic} employs \textbf{linear function approximation} to update its $M$ value function estimates, $\{V_w^j\}_{j=1}^M$, via TD learning on mini-batches of trajectory data. 
Subsequently, the \textbf{actor} uses these estimates to form the TD-error as an advantage approximation for each objective $j \in [M]$, and then computes the per-objective gradients using the policy gradient theorem \citep{sutton1999policy}. Beyond local gradient computation, the federated optimization protocol, including server aggregation and our regularized MGDA solver, remains identical to Algorithm~\ref{alg:firm}. 
Our theoretical framework is a novel extension of the multi-objective AC algorithm of \citet{zhou2024finite} to the federated setting, with the regularized MGDA solver ensuring robust consensus on the descent direction. 
The complete algorithm is provided in Appendix~C (See Algorithm~2), while in the remainder of this section we carry out our analysis under the following standard assumptions.

\begin{assumption}
\label{Assumption:FuncApproxMOMDP_1}
For each client $c \in \{1, \dots, C\}$, its local MOMDP and the global policy satisfy: (a) The policy function $\pi_\theta(a|s)$ is continuously differentiable with respect to $\theta$.
(b) The Markov process induced by any policy $\pi_\theta$ in any client's environment $P_c$ is irreducible and aperiodic.
    (c) The instantaneous reward $r_t^{i,c}$ for any objective $i$ on any client $c$ is non-negative and uniformly bounded by a constant $r_{\max} > 0$.
\end{assumption}

Assumption~\ref{Assumption:FuncApproxMOMDP_1} imposes standard regularity conditions \citep{zhou2024finite}. 
Condition (a) requires differentiability, which is essential for applying policy gradient methods. 
Condition (b) ensures ergodicity of the Markov process, yielding a unique stationary distribution under any policy and thereby guaranteeing that long-term objectives are well defined. 
Finally, condition (c) assumes bounded rewards, a standard requirement to ensure that the value functions remain bounded.

\begin{assumption}[Linear Function Approximation]
\label{assum:linear_fa}
We make the following assumptions on the value function approximation:
(a) For each objective $j \in [M]$, the value function is approximated from a linear function class, $V_w^j(s) = \phi(s)^\top \mathbf{w}^j$, where $\mathbf{w}^j \in \mathbb{R}^{d_2}$ are learnable parameters and $\phi: \mathcal{S} \to \mathbb{R}^{d_2}$ is a shared feature map.
(b) The feature map is normalized such that $\|\phi(s)\|_2 \le 1$ for all $s \in \mathcal{S}$.
(c) Let $\mathbf{A}_{\pi_\theta} \triangleq \mathbb{E}_{s \sim d_{\pi_\theta}, s' \sim P_{\pi_\theta}(\cdot|s)} [(\gamma\phi(s') - \phi(s))\phi(s)^\top]$. We assume this matrix is negative definite, i.e., there exists a constant $\lambda_A > 0$ such that its symmetric part satisfies $\mathbf{x}^\top(\mathbf{A}_{\pi_\theta} + \mathbf{A}_{\pi_\theta}^\top)\mathbf{x} \le -2\lambda_A \|\mathbf{x}\|_2^2$.
\end{assumption}

Assumption~\ref{assum:linear_fa} is standard in the analysis of linear temporal-difference learning~\citep{tsitsiklis1999average, xu2020improving, xu2020reanalysis, qiu2021finite, zhou2024finite}. The conditions collectively ensure that the projected Bellman equation is well-posed for any policy $\pi_\theta$. 

\begin{assumption}[Boundedness and Smoothness]
\label{Assump:Smooth_bounded_AISTAT_1}
For any policy parameters $\theta, \theta'$ and any state-action pair $(s,a)$, there exist positive constants $C_\psi$ and $L_J$ such that: (a) The score function is uniformly bounded: $\|\psi_\theta(a|s)\|_2 = \|\nabla_\theta \log \pi_\theta(a|s)\|_2 \le C_\psi$.
(b) The gradient of each local objective function $J^{i,c}(\theta)$ is Lipschitz continuous with respect to the policy parameter: $\|\nabla_\theta J^{i,c}(\theta) - \nabla_\theta J^{i,c}(\theta')\|_2 \le L_J \|\theta - \theta'\|_2$ for all objectives $i \in [M]$ and all clients $c \in \mathcal{C}$.
\end{assumption}
This assumption enforces standard smoothness and boundedness conditions.
The bounded score function (a), satisfied by softmax policies, prevents
unbounded gradient updates \citep{xu2020improving}. 


\begin{assumption}[Bounded Heterogeneity]
\label{Assump:Heterogeneity}
There exists a non-negative constant $\zeta$ such that for all clients $c \in \{1, \dots, C\}$ and any policy parameter $\theta$, the deviation between the local and global objective gradients is bounded:
\(
\| \nabla_\theta \mathbf{J}^c(\theta) - \nabla_\theta \mathbf{J}(\theta) \|_F \le \zeta.
\)
\end{assumption}

    The constant $\zeta$ quantifies the degree of data heterogeneity across clients, capturing the non-IID nature of the data. Explicitly, $\zeta$ relates to the heterogeneity in transition kernels ($\epsilon_p$) and reward functions ($\epsilon_r$) as $\zeta = \mathcal{O}(\epsilon_p + \epsilon_r)$; we provide the formal derivation of this relationship in Appendix~\ref{App:HeterogeneityDerivation}.  The case $\zeta = 0$ recovers the homogeneous setting where clients share identical dynamics and rewards.

\subsection{Results and convergence analysis}
\label{sec:Convergence_AISTAT_1}
We now present our main theoretical result, which guarantees the convergence of TFIRM:
\begin{theorem}[Convergence of TFIRM]
\label{Theorem:ConvergenceAnalayis_AISTAT_1}
Under the specified assumptions, by choosing an appropriate step-size $\alpha$, the iterates produced by TFIRM satisfy:
\begin{equation}
\begin{aligned}
&\frac{1}{T} \sum_{t=1}^T  \mathbb{E}\left[\left\| \nabla_{\theta} \mathbf{J}(\bar{\theta}_{t})\, \lambda_{t} \right\|_2^2 \right] = 
  \mathcal{O}\Bigg(\underbrace{\frac{\log T}{\alpha \,T}}_{\text{Opt. Error}}
    + \underbrace{\frac{1}{C B}}_{\text{Variance}}
    \\
    &+ \underbrace{\sqrt{\zeta_{\text{approx}}} + \sqrt{\varepsilon_{\text{critic}}} +\zeta_{\mathrm{het}}^2}_{\text{Bias}}
    +  \underbrace{\alpha^2 K^2}_{\text{Classical Drift}} + \underbrace{\frac{\sqrt{M^3}}{\beta \; \sqrt{B}} \alpha K}_{\text{Disagr. Drift}}
    \Bigg)
\end{aligned}
\end{equation}
\textbf{Proof:} See Appendix~F.
\end{theorem}
Theorem~\ref{Theorem:ConvergenceAnalayis_AISTAT_1} provides the first convergence guarantee for a federated multi-objective actor-critic algorithm, ensuring that TFIRM converges to a Pareto stationary point. The overall convergence rate, dominated by the optimization error, is $\tilde{\mathcal{O}}(1/T)$. This matches the rate for centralized multi-objective actor-critic methods \citep{zhou2024finite}, demonstrating that our federated approach achieves a comparable asymptotic performance. Furthermore, the variance term, \(\mathcal{O}(1/(CB))\), exhibits the expected linear speedup with respect to the number of clients \(C\) and batch size \(B\), consistent with standard analyses in federated learning \citep{zhang2024finitetime}.

\begin{remark}[Controlling Disagreement Drift] 
\label{Remark:MultiObjective-Disagreement-Drift_1}The most significant insight from our analysis lies in the characterization of the client drift. The total drift consists of two components. The first, \(\mathcal{O}(\alpha^{2}K^{2})\), is the classical drift from local updates, which is well-understood in federated optimization \citep{zhou2024finite, zhang2024finitetime}. The second, and the core of our theoretical contribution, is a novel error term we identify as the \textit{multi-objective disagreement drift}, given by \(\mathcal{O}\!\left(\frac{\sqrt{M^{3}}}{\beta\sqrt{B}}\alpha K\right)\). This term precisely captures the error arising from clients solving the MGDA problem locally with stochastic gradients. Our bound reveals two key insights: (i) the challenge of reaching consensus grows with the number of objectives \(M\), and (ii) this drift can be explicitly controlled by the product of the regularization \(\beta\) and the batch size \(B\).
\end{remark}

\paragraph{Comparing to \citet{askin2024federated}}
While our work is the first in the RL setting, the most relevant methodological comparison is with the federated multi-objective supervised learning framework of \citet{askin2024federated}. Their \textbf{server-centric} approach tackles client disagreement by having clients communicate compressed gradients to the server, which then solves a single MGDA problem and broadcasts the solution vector $\lambda$ back. This design for supervised learning introduces an error term of $\mathcal{O}(qM)$ into the convergence upper bound, where the factor $q$ depends on the quality of the gradient compression performed at each client. This approach, therefore, introduces significant overhead. \textbf{A new error source:} the convergence is fundamentally limited by the quality of the gradient compression, as represented by the $q$ term. \textbf{Communication cost:} it requires an extra round-trip communication in every training step (gradients up, $\lambda$ down). \textbf{Computational cost:} it relies on a dimensionality reduction step on the gradients at each client, which can be computationally prohibitive for large models. In  contrast, our approach removes the need for costly server-side coordination and its associated trade-offs. By introducing a lightweight $\ell_{2}$ regularization to each client's local MGDA objective, we provably control the multi-objective disagreement drift.

\begin{remark}
Our upper bound differs from \citet{askin2024federated} by an additional $\sqrt{M}$ factor. This may stem from the fact that client drift in RL is more severe than in supervised learning, as local policy updates affect both the local model parameters and the local induced data distribution. Understanding whether this dependence is tight remains an interesting open problem left for future work.
\end{remark}

\paragraph{Proof Sketch.}
Our analysis begins from the standard descent lemma on the global objective, evaluated at the server model \(\bar{\theta}_t\). Let \(\alpha \in (0, \frac{1}{L_J})\), then the expected one-step progress is given by:
\begin{equation}
\begin{aligned}
\lambda_t^\top J(\bar{\theta}_{t+1}) &\geq \lambda_t^\top J(\bar{\theta}_t) + \frac{\alpha}{2} \big\| \nabla_\theta \mathbf{J}(\bar{\theta}_t)\, \lambda_t \big\|_2^2\\
&    - \frac{\alpha}{2} \underbrace{\Big\| \frac{1}{C}\sum_{c=1}^C g_t^{c}(\theta_t^c)   - \nabla_\theta \mathbf{J}(\bar{\theta}_t) \lambda_t\Big\|_2^2}_{\mathcal{T}_1}
\end{aligned}
\end{equation}
The error term $\mathcal{T}_1$ captures the difference between the averaged client updates and the true gradient direction at the average model. We decompose this term into two primary sources of error:
\begin{equation}
\begin{aligned}
    \mathcal{T}_1 &\leq 2\underbrace{\mathbb{E}[\Big\| \frac{1}{C}\sum_{c=1}^C \left(g_t^{c}(\theta_t^c) - \nabla_\theta \mathbf{J}^c(\theta^c_t) \lambda_t^c\right) \Big\|_2^2]}_{\mathcal{T}_{1,1} (\text{Local Error})} \\
    & \quad + 2\underbrace{\mathbb{E}[\Big\| \frac{1}{C}\sum_{c=1}^C \big(\nabla_\theta \mathbf{J}^c(\theta^c_t) - \nabla_\theta \mathbf{J}(\bar{\theta}_t)\big) \lambda_t\Big\|_2^2]}_{\mathcal{T}_{1,2} ( \text{Client Drift})}
\end{aligned}
\end{equation}
where the term $\mathcal{T}_{1,2}$ represents the standard client drift due to local updates and client heterogeneity and is bounded by $\mathcal{O}(K^2\alpha^2 + \zeta^2)$, where $K$ is the number of local steps. The main novelty of our work lies in bounding the local error, $\mathcal{T}_{1,1}$, which we further decompose:
\begin{equation}
\label{eq:mainSec_AISTA_ProofSteps_1}
\mathcal{T}_{1,1} \le \underbrace{\|\dots\|_2^2}_{\mathcal{T}_{1,1}^{\text{gradient-error}}} + \underbrace{\left\| \frac{1}{C}\sum_{c=1}^C \sum_{j=1}^M g_t^{j, c}(\theta_t^c) \left( \lambda_t^{j,c} - \bar{\lambda}_t^j \right) \right\|_2^2}_{\mathcal{T}_{1,1}^{\text{disagr-drift}}},
\end{equation}
where $\bar{\lambda}_t^j \triangleq \frac{1}{C}\sum_{c'=1}^C \lambda_t^{j,c'}$. The term $\mathcal{T}_{1,1}^{\text{gradient-error}}$ captures standard stochastic gradient noise and critic error, which can be bounded using existing techniques and shows a linear speedup with $C \times B$. The crucial term is $\mathcal{T}_{1,1}^{\text{disagr-drift}}$,   which directly captures the disagreement among client-computed $\lambda_t^c$ vectors. 

\begin{remark}[\textbf{Multi-Objective Disagreement Drift.}]
\label{Remark:DisagreementDrift_Intro_1}
We denote the second term, $\mathcal{T}_{1,1}^{\text{disagr-drift}}$, as \emph{multi-objective disagreement drift}. This term quantifies the deviation induced by heterogeneous client solutions to the MGDA subproblem under stochastic gradients. The instability stems from the non-smooth dependence of the unregularized MGDA solution on its inputs, which makes the aggregated update direction highly sensitive to local sampling noise.
\end{remark}

\paragraph{Why do we need regularization?} As discussed in Remark~\ref{Remark:DisagreementDrift_Intro_1}, bounding $\mathcal{T}_{1,1}^{\text{disagr-drift}}$ is challenging because the MGDA solver can be highly sensitive to noise in its input gradients. Our key technical contribution is to show that the $\beta$-strong convexity induced by our regularized objective provides exactly the control needed to bound this term. This is formalized in the following lemma.

\begin{lemma}[Regularization Controls  Disagreement]
\label{lemma:lambda_stability}
For any two clients $c, c' \in \mathcal{C}$, the difference in their locally computed optimal MGDA weights is bounded by the maximum difference between their objective gradients:
\begin{equation}
    \|\lambda_t^{*, c} - \lambda_t^{*, c'}\|_2 \leq \frac{4 R M}{\beta} \max_{j\in[M]} \|g_t^{j,c}(\theta_t^c) - g_t^{j,c'}(\theta_t^{c'}) \|_2
\end{equation}
\end{lemma}

Finally, applying Lemma~\ref{lemma:lambda_stability} to Equation~\ref{eq:mainSec_AISTA_ProofSteps_1}, together with bounding the residual terms in $\mathcal{T}_1$ and telescoping the descent lemma over $T$ rounds, leads directly to our main convergence result.


\section{Numerical Experiments}
\label{sec:experiments}

This section empirically validates our proposed framework. Our experiments are designed to answer three central research questions:  \textbf{(RQ1) Performance:} Can our communication-efficient, client-centric algorithm, \textbf{FIRM}, achieve comparable or better performance than a server-centric paradigm that resolves objective conflicts at the server? \textbf{ (RQ2) Regularization Effect:} What is the impact of removing regularization in FIRM on \textit{multi-objective disagreement drift}? \textbf{(RQ3) Preference-Guided Alignment:} Can the regularization term be used to incorporate preferences and adjust the balance between competing objectives?

 We fine-tune \path{meta-llama/Llama-3.2-1B-Instruct} \footnote{The model choice reflects realistic edge-device constraints in federated learning, where resources are limited.}  using \texttt{LoRA} adapters \citep{hu2021lora}, ensuring only adapter weights are trained and communicated. Our alignment task uses $M = 2$ objectives, \textbf{Helpfulness} and \textbf{Harmlessness}, with prompts from the Anthropic HH-RLHF \citep{bai2022training} dataset and rewards evaluated by public reward models: \path{Ray2333/gpt2-large-helpful-reward\_model} (helpfulness), \path{Ray2333/gpt2-large-harmless-reward\_model} (harmlessness) \citep{yang2024rewards}. Moreover,  all reward scores are normalized to the $[0,1]$ range. Each generated response receives a 2-D reward vector \((r_{\mathrm{help}}, r_{\mathrm{harm}})\) which drives the PPO updates.  Our federated protocol involves $C$ clients over $T$ rounds, where each client performs a fixed number of local PPO-style updates before the server aggregates the LoRA adapters via \texttt{FedAvg}. For the local MGDA subproblem, we solve a regularized Quadratic Program (QP). However, the scale of the gradients, and thus their Gram matrix $G$, can vary dramatically during training. To ensure our regularization $\beta$ has a consistent effect, we first normalize $G$  by its trace, which stabilizes the local optimization (See Appendix~A).  We set the regularization parameter to \(\beta = 0.01\).   All experiments were conducted on a server equipped with NVIDIA H100 GPUs.

\begin{figure}[t]
    \centering
    \begin{subfigure}[b]{0.23\textwidth}
        \centering
        \includegraphics[width=\linewidth]{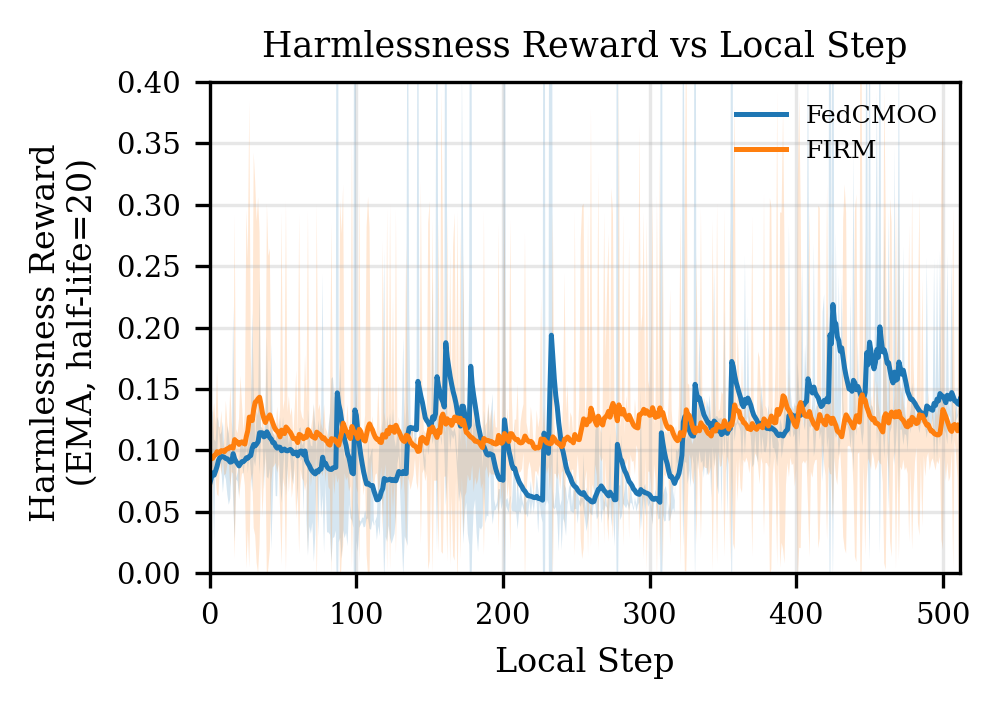}
        \caption{Harmlessness reward}
        \label{fig:harmless_reward}
    \end{subfigure}
    \hfill
    \begin{subfigure}[b]{0.23\textwidth}
        \centering
        \includegraphics[width=\linewidth]{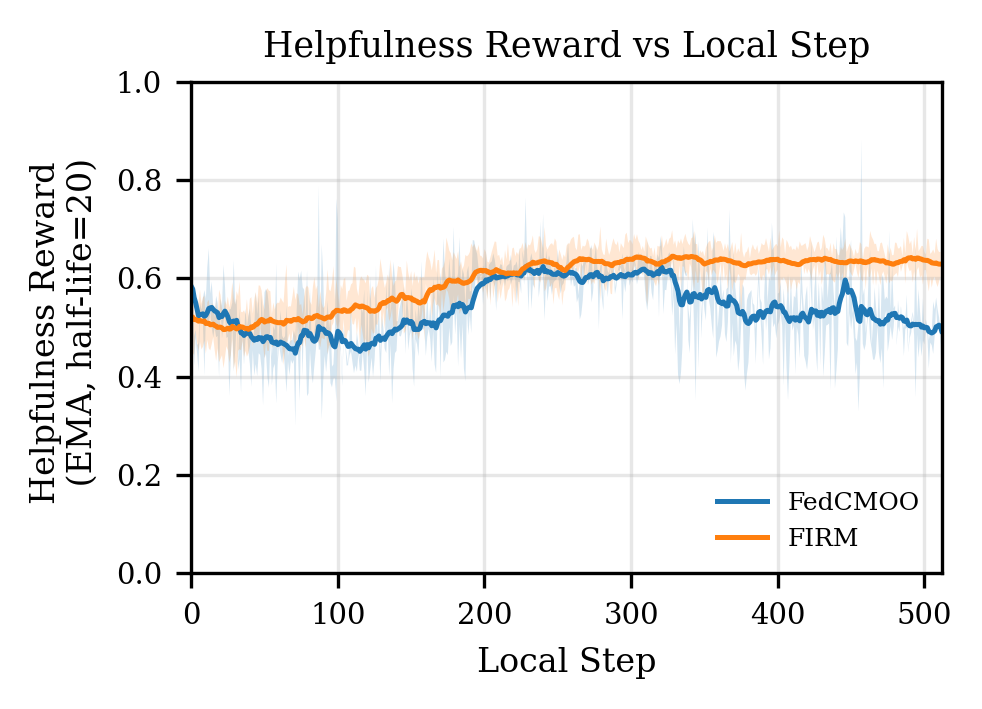}
        \caption{Helpfulness reward}
        \label{fig:helpfulness_reward}
    \end{subfigure}
        \vskip\baselineskip 

         \begin{subfigure}[b]{0.23\textwidth}
        \centering
        \includegraphics[width=\linewidth]{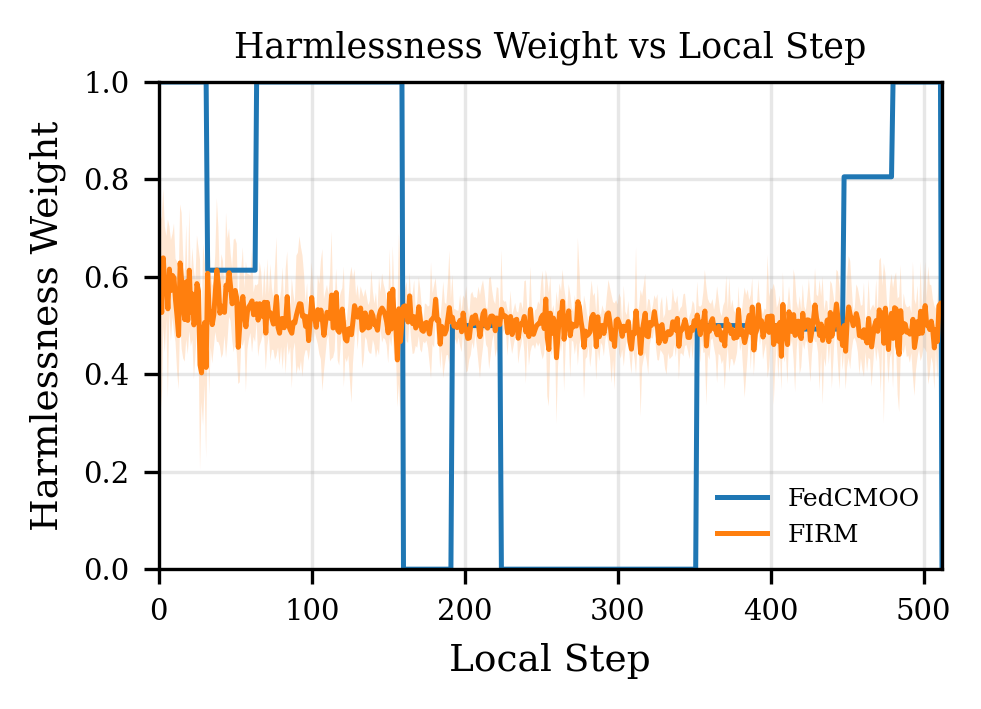}
        \caption{Harmlessness weight (\(\lambda\))}
        \label{fig:harmlessness_lambda}
    \end{subfigure}
    \hfill
    \begin{subfigure}[b]{0.23\textwidth}
        \centering
        \includegraphics[width=\linewidth]{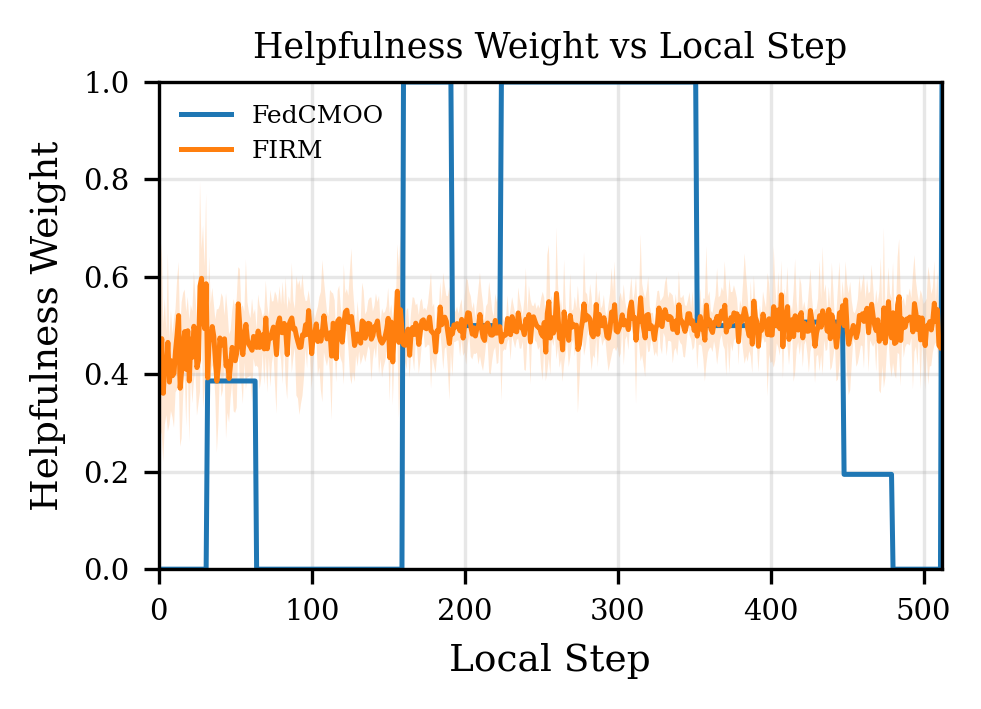}
        \caption{Helpfulness weight (\(\lambda\))}
        \label{fig:helpfulness_lambda}
    \end{subfigure}
   \caption{\textbf{Performance comparison between FIRM (orange) and the FedCMOO baseline (blue).}  All curves show mean performance across 8 clients. \textbf{Panels (a,b)}: reward trajectories,  smoothed with EMA (half-life=20), where FIRM achieves higher, more stable helpfulness with comparable  harmlessness. \textbf{Panels (c,d)}: MGDA weights, showing that FIRM yields smoother, more consistent  trade-off decisions than FedCMOO.}
    \label{fig:rewards}
\end{figure}

\paragraph{RQ1: Comparison with Server-Centric Alignment.}
To evaluate FIRM, we compare it against a SOTA server-centric baseline, \textbf{FedCMOO}, using a \textbf{non-IID} partition ($\text{Dir}(\alpha=0.3)$) that induces \textbf{data heterogeneity} across clients. FedCMOO is adapted from the federated multi-objective supervised learning algorithm of \citet{askin2024federated} to the LLM alignment setting. In FedCMOO, clients send their local multi-objective gradients to the server. The server then solves a single MGDA problem to compute a global trade-off vector \(\lambda\) and broadcasts it back to the clients. This architecture avoids the multi-objective disagreement drift by design, as all clients are forced to use the same \(\lambda\). To ensure a fair comparison focused purely on the conflict resolution strategy, we do not use gradient compression in FedCMOO \citep{askin2024federated}, thereby removing any potential compression error. As shown in Figure~\ref{fig:rewards}, FIRM, with 8 clients, \textbf{achieves comparable or superior performance to FedCMOO}. One hypothesis for FIRM's strong performance is the agility of its local updates. FIRM clients can adjust their trade-off vectors \(\lambda_c\) at every local step, allowing for a more responsive and potentially better balance between objectives. In contrast, FedCMOO clients must wait for the server's global \(\lambda\), which can become ``stale'' or lagged between communication rounds. This  lag results in oscillatory weight trajectories, as the server repeatedly overcorrects its global \(\lambda\) (See  Figure~\ref{fig:harmlessness_lambda} and Figure~\ref{fig:helpfulness_lambda}).

\begin{figure}[ht]
    \centering
    \begin{subfigure}[b]{0.23\textwidth}
        \centering
        \includegraphics[width=\linewidth]{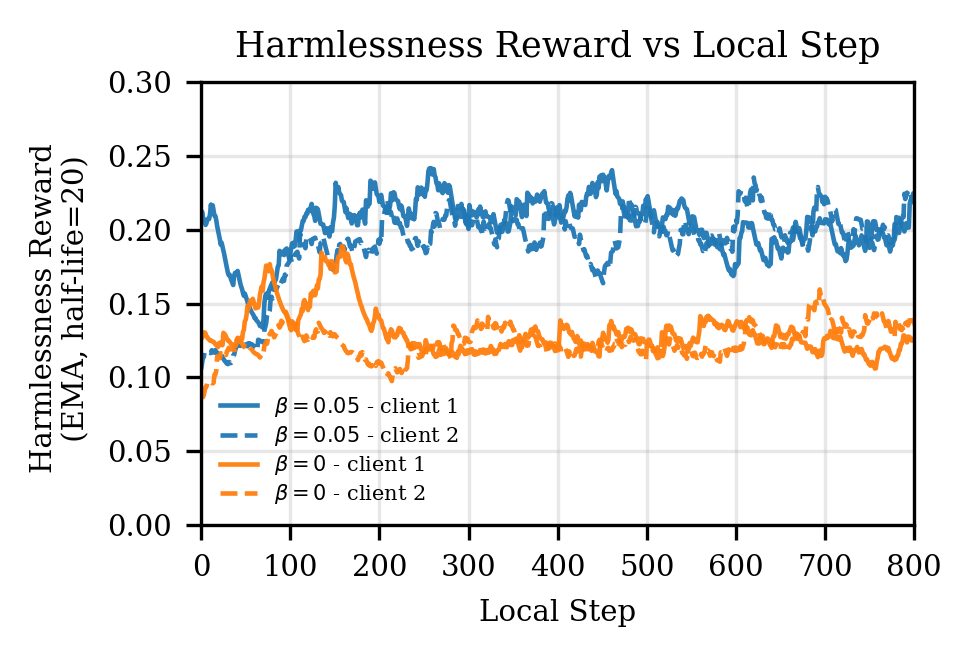}
        \caption{Harmlessness Reward}
        \label{fig:harmless_reward_abll}
    \end{subfigure}
    \hfill
    \begin{subfigure}[b]{0.23\textwidth}
        \centering
        \includegraphics[width=\linewidth]{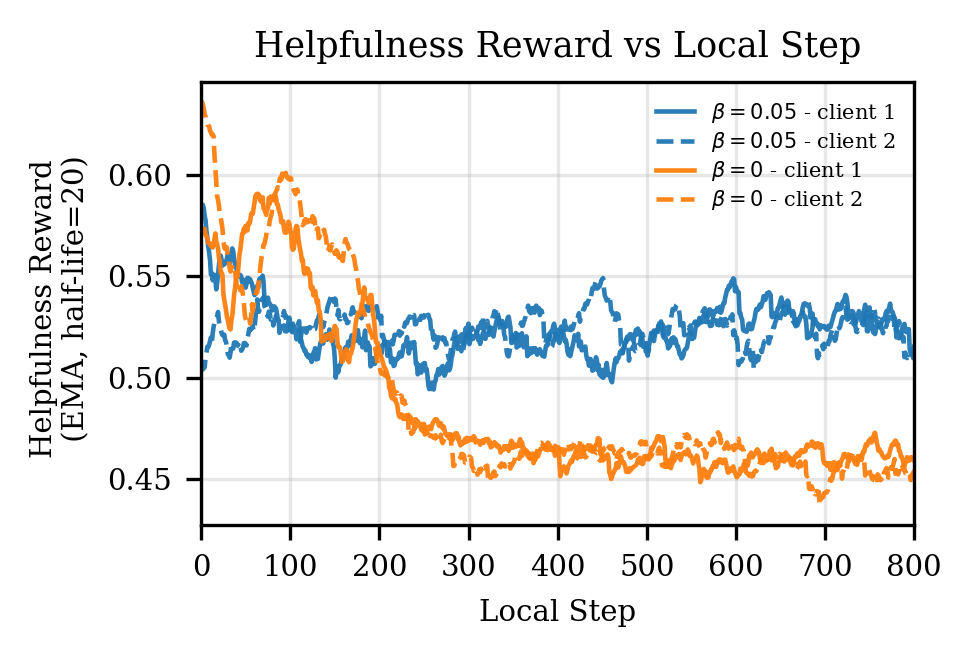}
        \caption{Helpfulness Reward}
        \label{fig:helpfulness_reward_abll}
    \end{subfigure}

    \vskip\baselineskip 
     \begin{subfigure}[b]{0.22\textwidth}
        \centering
        \includegraphics[width=\linewidth]{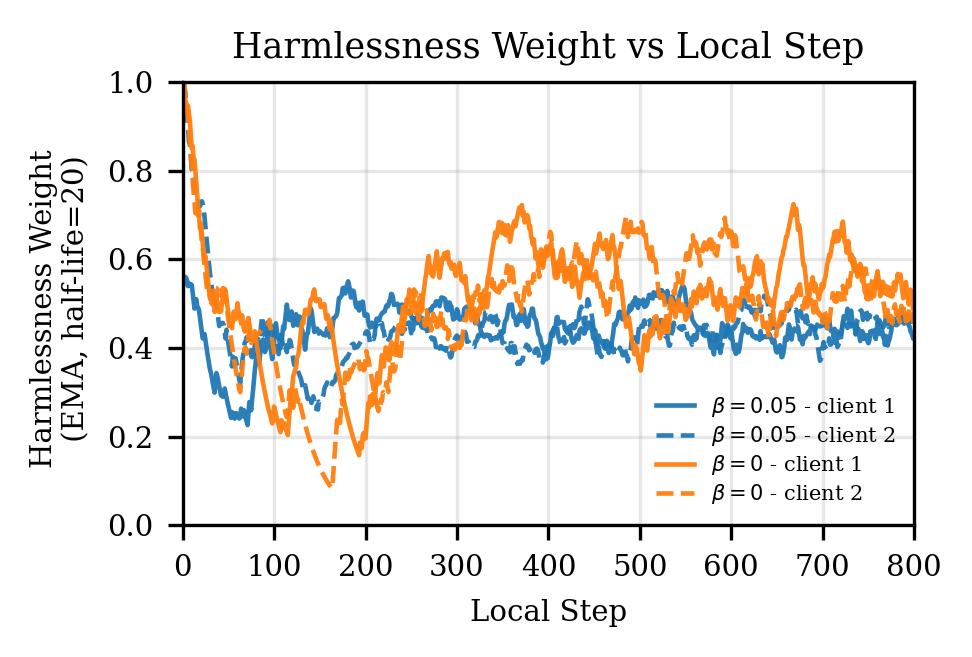}
        \caption{Harmlessness Lambda}
        \label{fig:harmlessness_lambda_abllation}
    \end{subfigure}
    \hfill
    \begin{subfigure}[b]{0.23\textwidth}
        \centering
        \includegraphics[width=\linewidth]{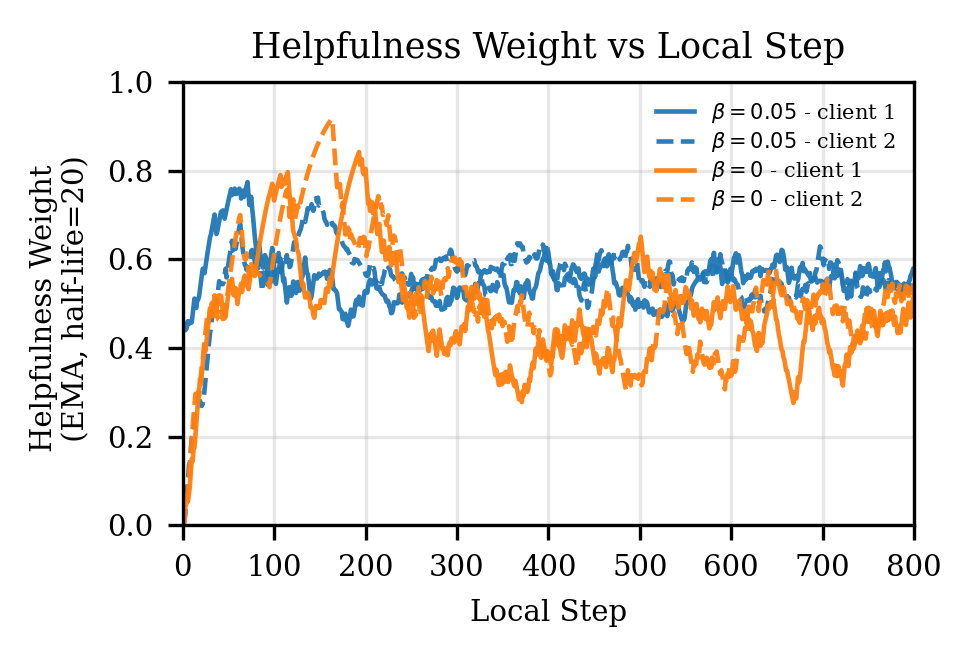}
        \caption{Helpfulness Lambda}
        \label{fig:helpfulness_lambda_abllation}
    \end{subfigure}
 \caption{Reward trajectories and MGDA weights under \(\beta=0\) (orange) and \(\beta=0.05\) (blue).  All panels (a,b,c,d) are smoothed with EMA (half-life=20). Without regularization (\(\beta=0\)),  harmlessness remains low and helpfulness plateaus near 0.46, while MGDA weights fluctuate erratically  across clients (c,d). With \(\beta=0.05\), FIRM achieves more favorable trade-offs and exhibits smoother,  more consistent weight evolution, reducing client drift.}
    \label{fig:rewardsLambda}
\end{figure}

\paragraph{(RQ2) Regularization Effect.} Next, we conduct an ablation study to isolate the effect of our proposed regularization.  We compare FIRM with $\beta = 0.05$ against an unregularized ($\beta = 0$) baseline in a two-client setting for visual clarity. The results in Figure~\ref{fig:rewardsLambda} are stark.  \textbf{Without regularization, the MGDA weight trajectories ($\lambda$) for the two clients diverge significantly}, a direct visualization of the multi-objective disagreement drift (Figures~\ref{fig:helpfulness_lambda_abllation} and \ref{fig:harmlessness_lambda_abllation}). This instability degrades performance, yielding lower rewards. In contrast, FIRM’s regularization ($\beta > 0$) enforces consistent $\lambda$ trajectories, enabling stable and effective alignment.

\begin{figure}[b]
        \centering
        \includegraphics[width=0.5\linewidth]{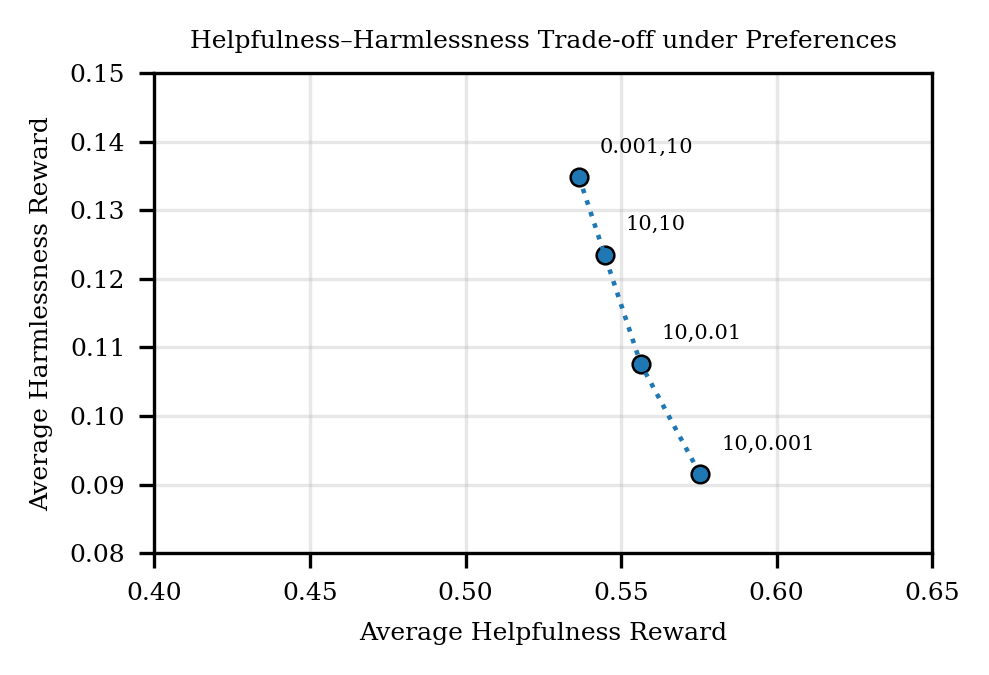}
    \caption{FIRM navigates the Helpfulness-Harmlessness trade-off. Each marker is a global model trained with a different preference vector \(\mathbf{p}\).}
    \label{fig:paretoFront}
    \end{figure}

\paragraph{RQ3: Preference-Guided Alignment.} 
Finally, as detailed in Section~\ref{Sect:algDescription}, we demonstrate that FIRM’s regularization  can be extended to incorporate preferences. Given a preference vector $\mathbf{p}$,  we use Equation~\eqref{eq:mgda-preference} to inject these preferences during training. As shown in Figure~\ref{fig:paretoFront}, varying $\mathbf{p}$ enables FIRM to effectively adjust the final trade-off between the two objectives. Increasing the preference for one objective demonstrably leads to a higher reward for that objective in the final model. This provides a practical tool to produce models tailored to specific alignment priorities.

\paragraph{Heterogeneous Client Reward Models.}

\begin{figure}[t]
\centering

\begin{subfigure}[b]{0.23\textwidth}
    \centering
    \includegraphics[width=\linewidth]{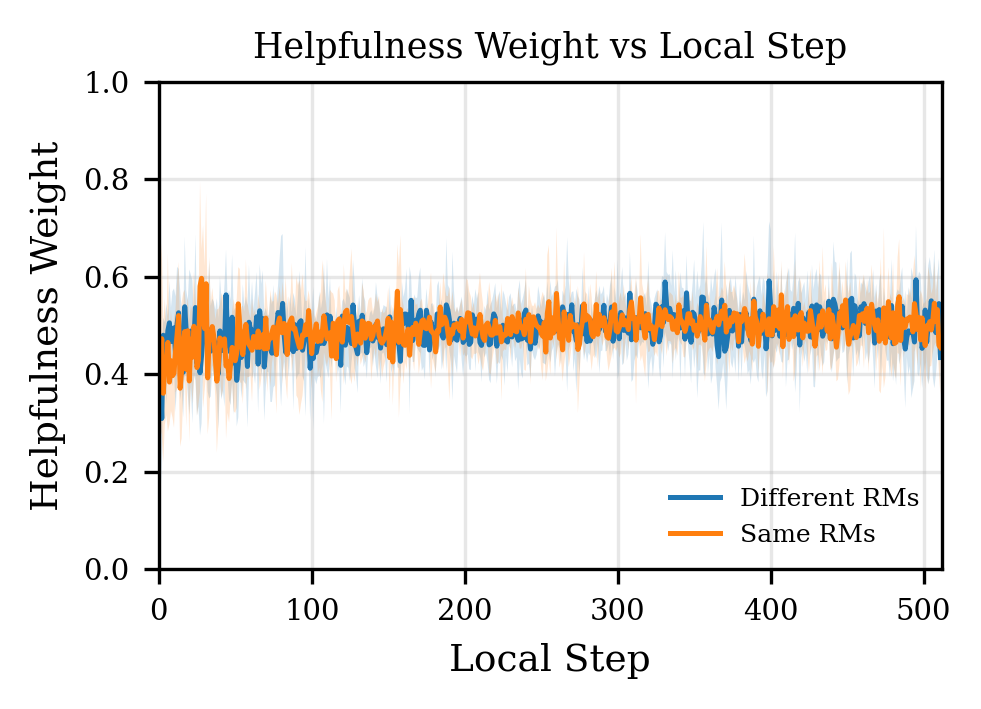}
    \caption{ Helpfulness Lambda.}
    \label{fig:help_weight_het}
\end{subfigure}
\hfill
\begin{subfigure}[b]{0.23\textwidth}
    \centering
    \includegraphics[width=\linewidth]{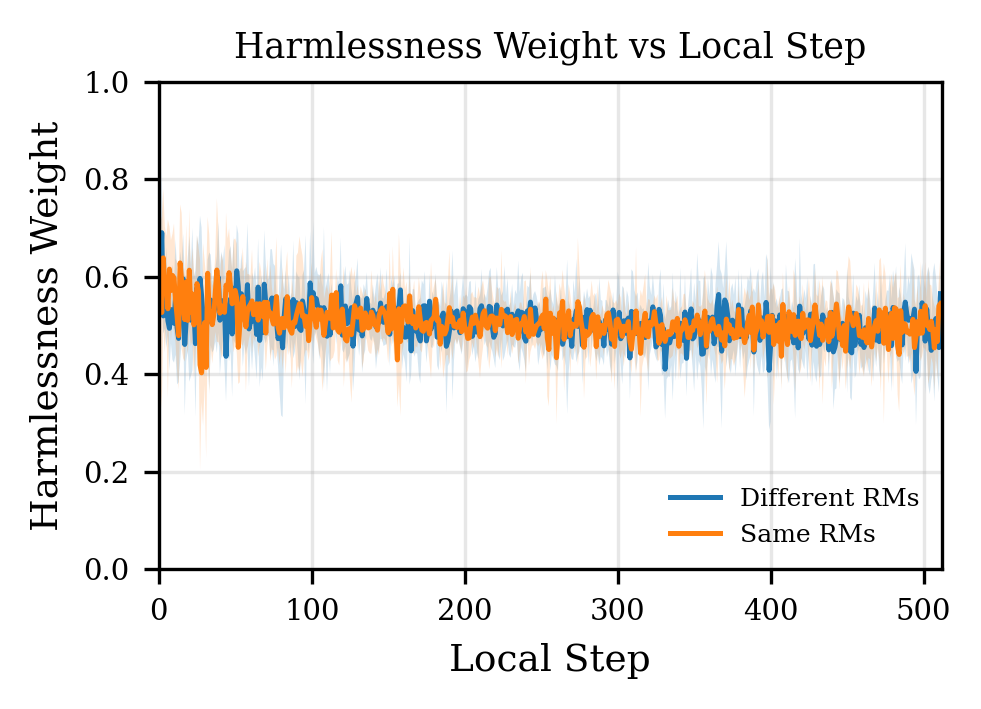}
    \caption{ Harmlessness Lambda .}
    \label{fig:harm_weight_het}
\end{subfigure}

  \vskip\baselineskip
  
\begin{subfigure}[b]{0.23\textwidth}
    \centering
    \includegraphics[width=\linewidth]{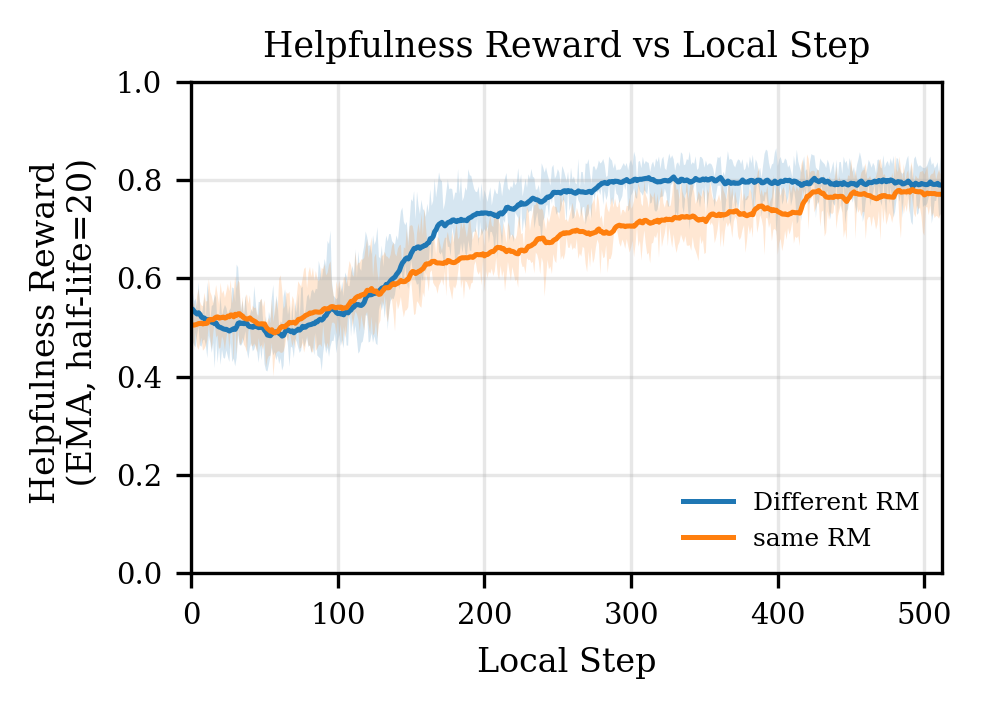}
    \caption{Helpfulness Reward.}
    \label{fig:help_reward_het}
\end{subfigure}
\hfill
\begin{subfigure}[b]{0.23\textwidth}
    \centering
\includegraphics[width=\linewidth]{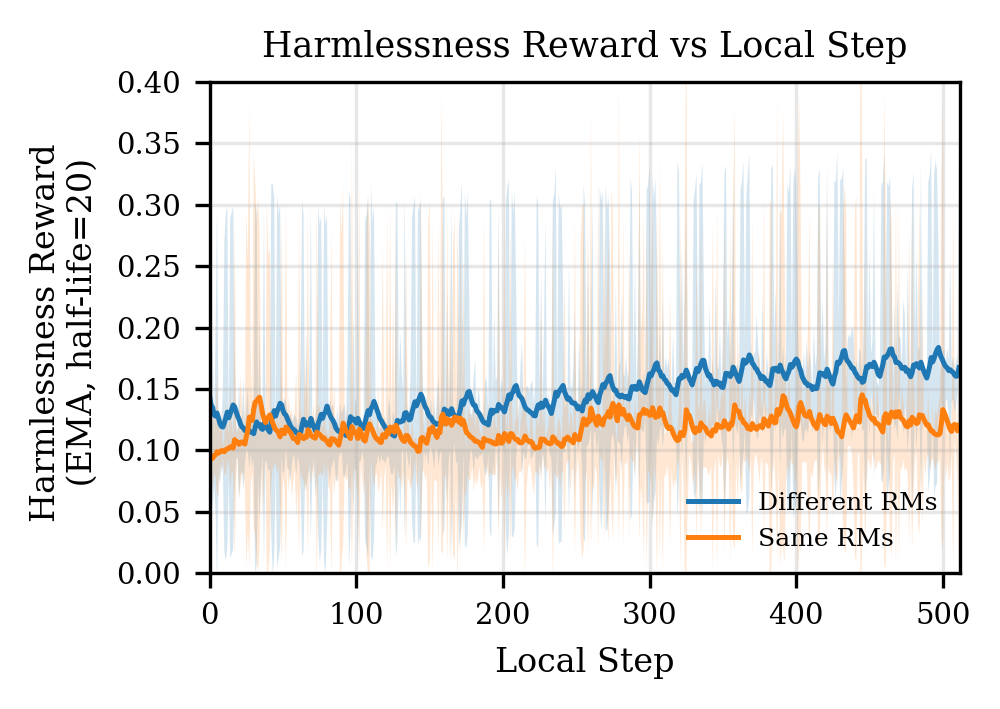}
    \caption{Harmlessness Reward.}
    \label{fig:harm_reward_het}
\end{subfigure}

\caption{
    \textbf{Robustness of FIRM to Heterogeneous Reward Models.} 
    This figure compares a homogeneous setup (all clients use the "Same RMs") against a heterogeneous one ("Different RMs").
    \textbf{(a, b):} The top row shows that the learned MGDA weights are remarkably stable, with nearly identical convergence dynamics in both settings. This confirms the robustness of our aggregation mechanism.
    \textbf{(c, d):} The bottom row shows that the resulting reward trajectories are highly competitive and closely matched. 
}
\label{fig:hetero_rm_results_main}
\end{figure}

Finally, beyond our core research questions, we investigate FIRM's resilience to heterogeneous reward models (RMs). In practical federated networks, clients may employ distinct RMs reflecting unique data or proprietary metrics. To test this, we simulate a scenario where half of the clients use the default helpfulness RM (\texttt{Ray2333}), while the other half use an alternative (\path{OpenAssistant/reward-model-deberta-v3-large-v2}). As shown in Figure~\ref{fig:hetero_rm_results_main}, FIRM proves highly robust to this diversity. The MGDA weight trajectories (Panels a, b) are nearly identical in both homogeneous and heterogeneous configurations, demonstrating that our aggregation mechanism effectively stabilizes the learning process despite disparate reward signals. Consequently, the reward curves (Panels c, d) confirm that FIRM continues to improve steadily on both objectives without degradation. 

Further experimental results are deferred to Appendix~\ref{Appendix:AdditionalExperiments}, including evaluations with three objectives, larger models, and more clients.

\section{Conclusion}
This paper addressed the challenge of scalable and private multi-objective alignment for LLMs. We introduced \textbf{FIRM}, a federated framework that resolves client-side conflicts and identified \emph{multi-objective disagreement drift}, a fundamental challenge in this setting. By equipping each client’s local solver with  regularization, we provably control drift and ensure stable convergence without prohibitive communication. Our experiments further show that this same mechanism can encode different preferences. Future work will explore personalized federated alignment, extending FIRM toward pluralistic models that accommodate diverse user values.

\section*{Impact Statement}
This paper presents work whose goal is to advance the field of Machine Learning by enabling scalable, privacy-preserving alignment of Large Language Models (LLMs). Our framework promotes democratization by allowing distributed entities to collaboratively align models without centralizing sensitive data. It enhances AI safety by stabilizing the optimization of conflicting objectives, such as helpfulness and harmlessness. However, we acknowledge a dual-use risk: the preference mechanisms designed to enforce safety could theoretically be inverted by malicious actors to suppress it. Furthermore, the system relies on the quality of the underlying reward models; biases in these models will be efficiently propagated to the global policy.
\bibliography{example_paper}
\bibliographystyle{icml2026}

\newpage
\appendix
\onecolumn

This appendix provides supplementary material to support the main paper. We begin in Appendix~\ref{Appendix:AdditionalExperiments} by presenting additional experimental results and describing our experimental setup. A more comprehensive discussion on related work is available in Appendix~\ref{App:relatedWork}. The subsequent sections are dedicated to our theoretical analysis. In Appendix~\ref{App:AlgTheoryDescription}, we formally present the algorithm used for the theoretical proofs. We then establish the necessary mathematical groundwork, including key definitions and notations, in Appendix~\ref{App: Preliminaries}. Appendix~\ref{Appendix:HelpfulLemma_AISTAT_1} and Appendix~\ref{App:Utility_Lemmas_in_Main Proof} provide helpful and utility lemmas, respectively, that serve as building blocks for our main theoretical result. The complete, step-by-step proof of our main theorem is detailed in Appendix~\ref{Appendix:FullProof}. Specifically, for reviewers interested in our core theoretical contribution, the proof that bounds the multi-objective disagreement introduced in our paper can be found in Subsection~\ref{App:SubSection_MOODriftMitigation_Novel_1}.

\section{Additional Experimental Results for FIRM}
\label{Appendix:AdditionalExperiments}
\subsection{Experimental Setup}
\label{app:exp_setup}

All our experiments are designed to be reproducible. This section provides a comprehensive overview of the models, datasets, and hyperparameters used throughout our evaluation of \textbf{FIRM}.

\paragraph{Language Models and Datasets.}
We conduct experiments on a publicly available Large Language Model (LLM):  \texttt{meta-llama/Llama-3.2-1B-Instruct}. The choice of this model is motivated by our federated learning setting, as such models are more suitable for deployment on resource-constrained edge devices. To ensure computational efficiency, we employ Parameter-Efficient Fine-Tuning (PEFT) using Low-Rank Adaptation (LoRA)~\citep{hu2021lora}. For generating responses, we use prompts from the Anthropic Helpfulness and Harmlessness (HH) dataset~\citep{bai2022training}. Additional results on the larger \texttt{Llama-3.1-8B-Instruct} model appear in Appendix~\ref{sec:llama8b_exp}

\paragraph{Reward Models (RMs).}
We evaluate alignment across two primary objectives: helpfulness and harmlessness. We use publicly available reward models to score the generated responses. For harmlessness, we use \texttt{Ray2333/gpt2-large-harmless-reward\_model}. For helpfulness, we primarily use \texttt{Ray2333/gpt2-large-helpful-reward\_model}. To test the robustness of our framework to diverse client preferences, we also conduct experiments where a subset of clients uses an alternative helpfulness RM, \texttt{OpenAssistant/reward-model-deberta-v3-large-v2}, as detailed in Appendix \ref{diff_rm}. 

\paragraph{Federated Learning Configuration.}
Our default federated learning setup consists of $C=8$ clients. The training process runs for a total of 16 communication rounds.  Further ablations on scalability (16 clients) are presented in Appendix~\ref{client_num}. 

\paragraph{Hyperparameters.} \label{hyperpar}
Our implementation is built using the TRL library \citep{vonwerra2022trl}. For the PPO algorithm, we set the actor and critic learning rates to $6 \times 10^{-5}$ and $1 \times 10^{-4}$, respectively. Each client performs 3 local PPO epochs per round, using a batch size of 16 and a minibatch size of 8. We use an adaptive KL controller with a target KL of $0.03$ to stabilize training. For efficient fine-tuning, LoRA is applied to all projection layers (\texttt{q\_proj, k\_proj, v\_proj, o\_proj}) with a rank of $r=16$. The regularization parameter for our MGDA-based solver was set to $\beta_{\text{mgda}}=0.01$.

\paragraph{Implementation Note on Solver Stability.}
A key parameter in our solver is the regularization term $\beta_{\text{mgda}}$. However, the effectiveness of a fixed $\beta$ is challenged by the high-dimensional nature of LLM gradients, whose norms can vary dramatically throughout training. A naive choice can lead to poor conditioning or allow the regularization to overwhelm the objective. To circumvent this, our implementation incorporates a scale-aware normalization of the Gram matrix $G$:
$$ \widehat{G} \;\triangleq\; \frac{G}{\tfrac{1}{M}\,\text{tr}(G)},$$
where $\text{tr}(G)$ is the trace of $G$. By normalizing the diagonal to have a unit scale, we ensure that the optimization is not dominated by the raw magnitude of the gradients. We then solve for the preference weights $\lambda^*$ using this scaled matrix:
\begin{equation}
\lambda^* \;\in\; \arg\min_{\lambda\in\Delta_M}\; \lambda^\top (\widehat{G} + \frac{\beta}{2} I)\,\lambda.
\label{eq:mgda-regularized-scaled}
\end{equation}
This technique is crucial for keeping the problem well-conditioned throughout training and ensuring that the regularizing effect of $\beta$ is consistent across all communication rounds.


\subsection{Ablation Studies on System and Heterogeneity}
\label{sec:ablations}

\subsubsection{Robustness to Heterogeneous Client Reward Models} \label{diff_rm}

In practical federated networks, clients may employ distinct reward models (RMs) to quantify the same objective, reflecting their unique data or proprietary methods. FIRM is designed to handle this challenging form of heterogeneity. To test this capability, we simulate a scenario where half of the clients use our default helpfulness RM (\texttt{Ray2333/...}), while the other half use a different one (\texttt{OpenAssistant/...}). All clients continue to use the same harmlessness RM. We compare this heterogeneous setup against our baseline where all clients use the same (homogeneous) RMs.

\begin{figure*}[t!]
\centering

\begin{subfigure}[b]{0.48\textwidth}
    \centering
    \includegraphics[width=\textwidth]{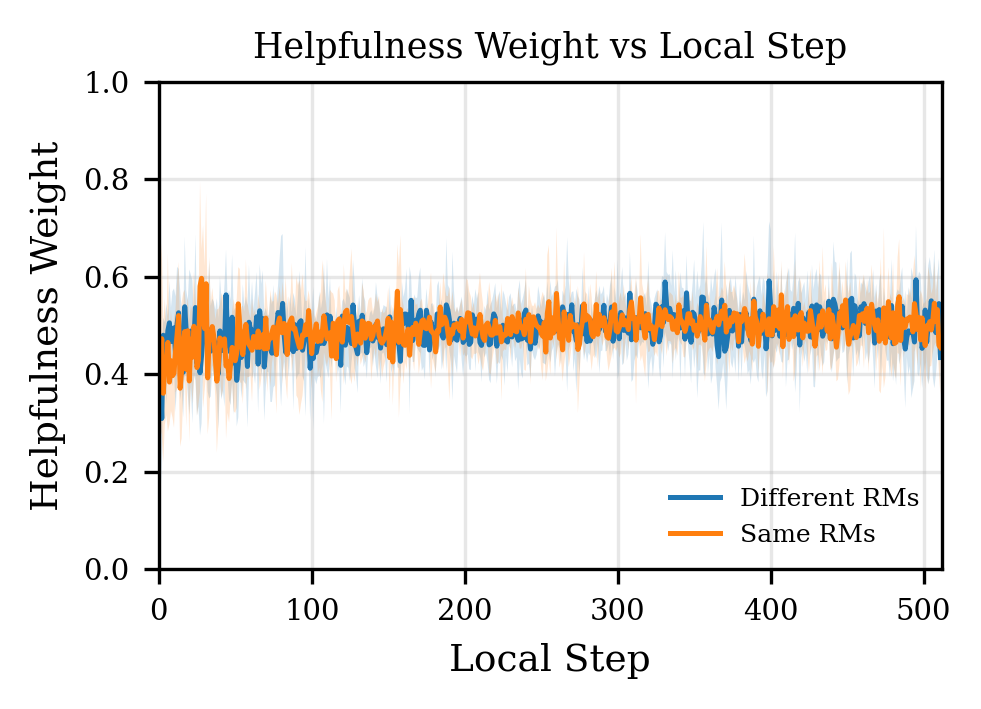}
    \caption{Evolution of the helpfulness weight ($\lambda_{\text{help}}$).}
    \label{fig:help_weight_het_app}
\end{subfigure}
\hfill
\begin{subfigure}[b]{0.48\textwidth}
    \centering
    \includegraphics[width=\textwidth]{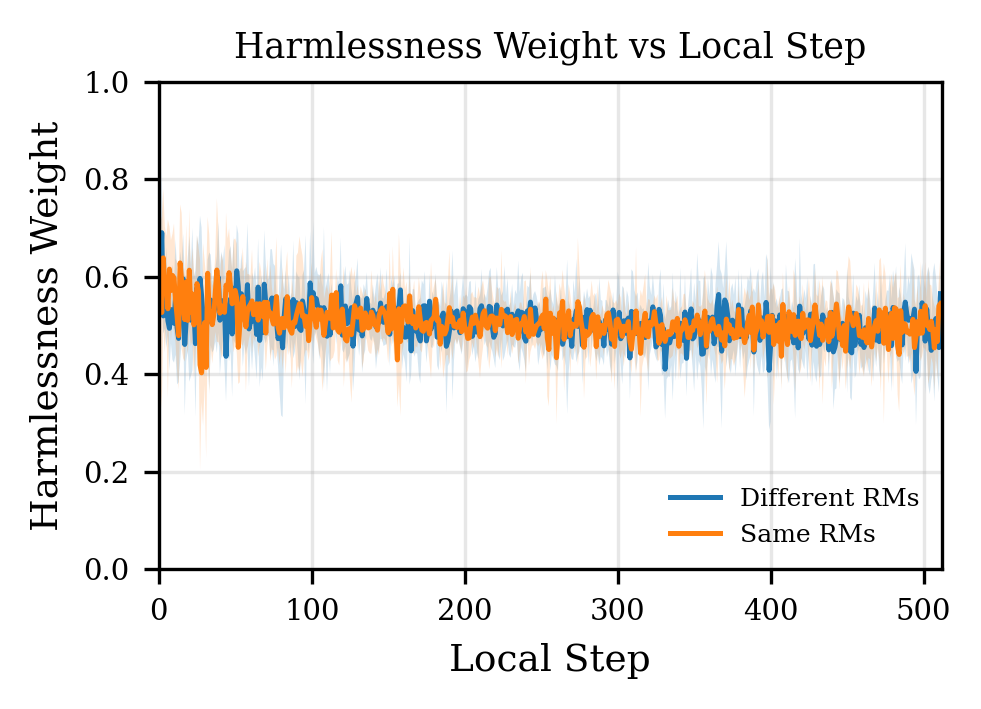}
    \caption{Evolution of the harmlessness weight ($\lambda_{\text{harm}}$).}
    \label{fig:harm_weight_het_app}
\end{subfigure}

\vspace{0.5cm}

\begin{subfigure}[b]{0.48\textwidth}
    \centering
    \includegraphics[width=\textwidth]{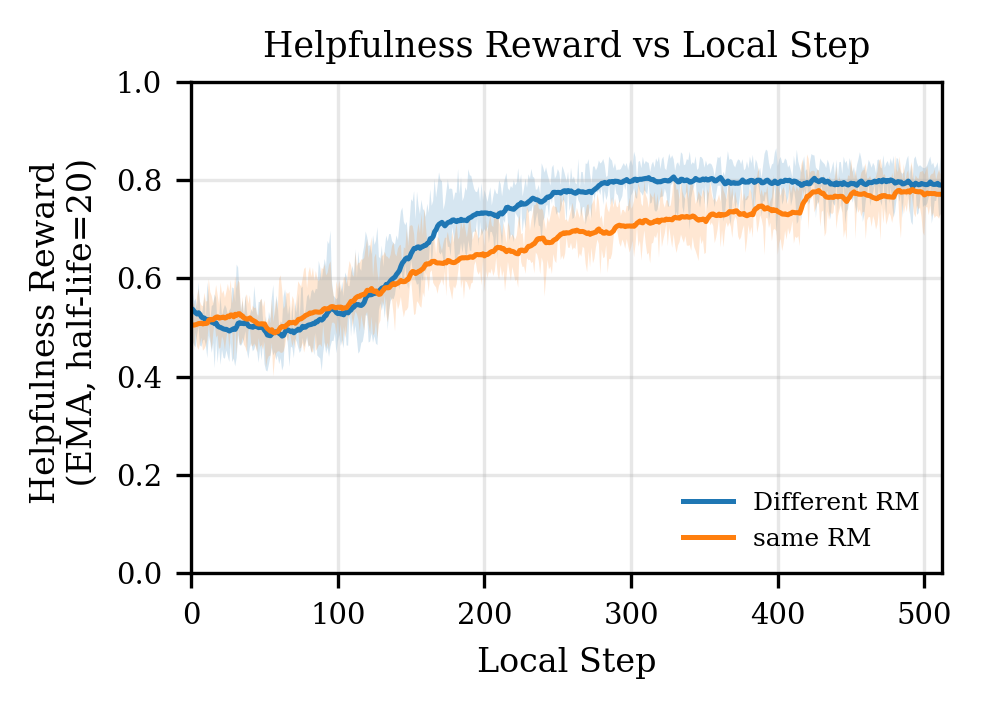}
    \caption{Helpfulness reward (EMA).}
    \label{fig:help_reward_het_app}
\end{subfigure}
\hfill
\begin{subfigure}[b]{0.48\textwidth}
    \centering
    \includegraphics[width=\textwidth]{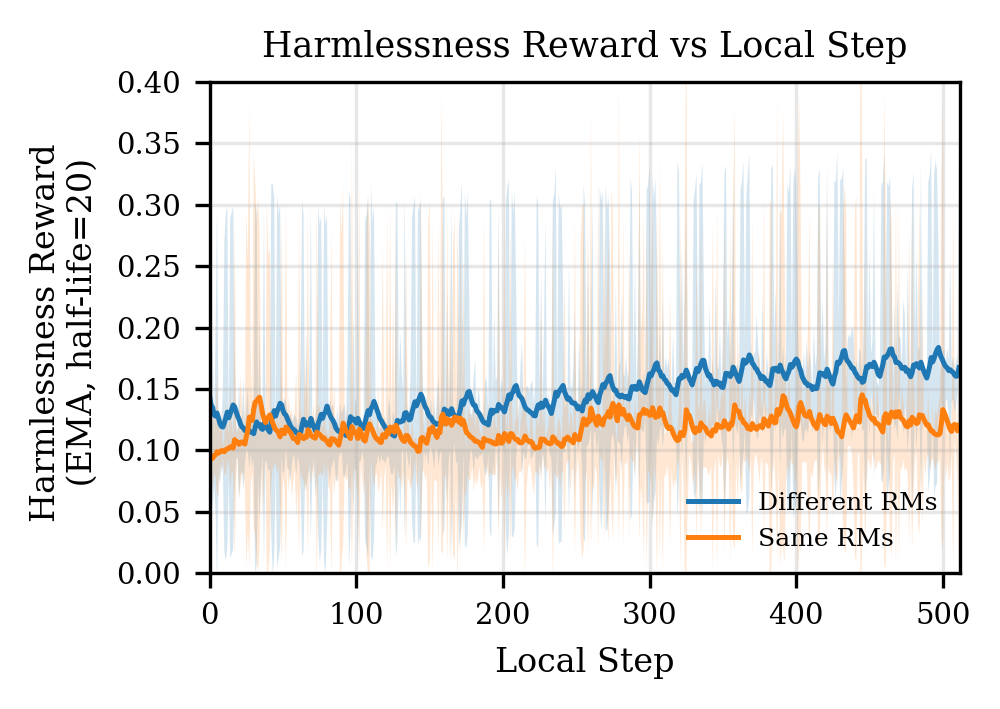}
    \caption{Harmlessness reward (EMA).}
    \label{fig:harm_reward_het_app}
\end{subfigure}

\caption{
    \textbf{Robustness of FIRM to Heterogeneous Reward Models.} 
    This figure compares a homogeneous setup (all clients use the "Same RMs") against a heterogeneous one ("Different RMs").
    \textbf{(a, b):} The top row shows that the learned MGDA weights are remarkably stable, with nearly identical convergence dynamics in both settings. This confirms the robustness of our aggregation mechanism.
    \textbf{(c, d):} The bottom row shows that the resulting reward trajectories are highly competitive and closely matched. FIRM maintains strong performance on both helpfulness and harmlessness, demonstrating its stability even when faced with diverse client reward signals.
}
\label{fig:hetero_rm_results}
\end{figure*}

The results, presented in Figure~\ref{fig:hetero_rm_results}, demonstrate that FIRM is highly robust to this diversity. The foundation for this stability is evident in the dynamics of the MGDA weights (Figures~\ref{fig:hetero_rm_results}a and \ref{fig:hetero_rm_results}b). The learning trajectories for the weights are nearly identical in both the homogeneous and heterogeneous configurations, converging to a stable equilibrium. This shows that our global aggregation mechanism is unperturbed by the underlying RM diversity.

The corresponding reward curves (Figures~\ref{fig:hetero_rm_results}c and \ref{fig:hetero_rm_results}d) further validate this robustness: in the heterogeneous setting, the model achieves comparable performance on both helpfulness and harmlessness objectives, closely tracking the homogeneous baseline. The learning curves show that FIRM continues to improve steadily on both objectives without degradation, confirming its reliability in handling variations in client preferences—a key requirement for real-world federated systems.

\subsubsection{Scalability with Increasing Numbers of Clients}
\label{client_num}
A critical requirement for any federated algorithm is the ability to scale gracefully as more clients join the network. We evaluate the scalability of FIRM by comparing our default 8-client configuration ($C=8$) against a larger 16-client setup ($C=16$). All other hyperparameters, including the total number of training rounds, were kept consistent across both experiments to isolate the effect of scale.

\begin{figure*}[t!]
\centering

\begin{subfigure}[b]{0.48\textwidth}
    \centering
    \includegraphics[width=\textwidth]{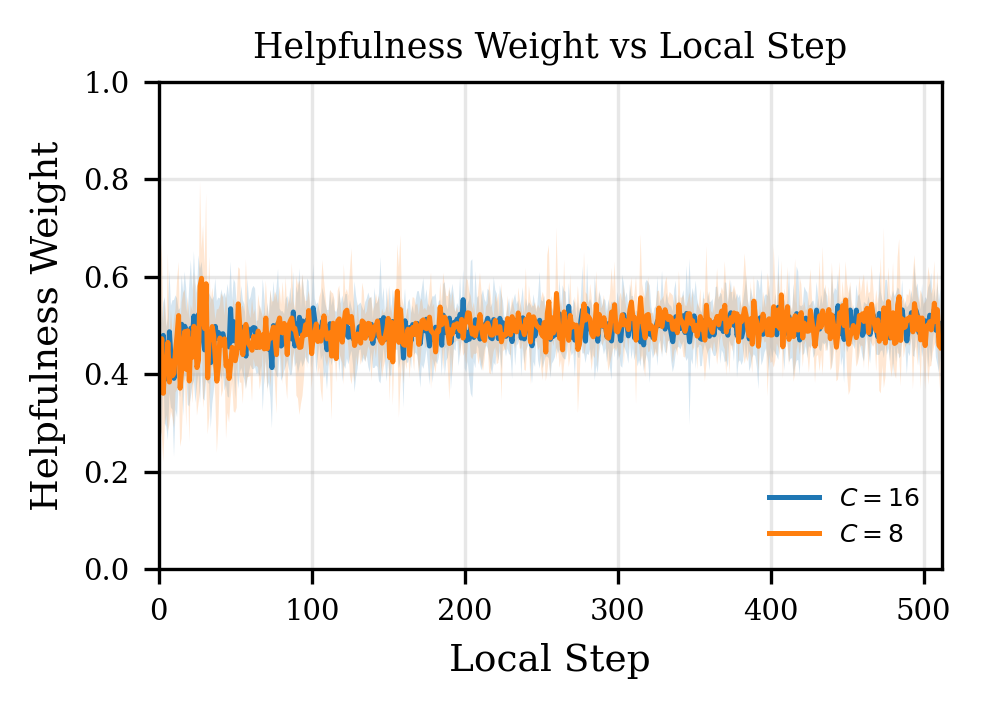}
    \caption{Evolution of the helpfulness weight ($\lambda_{\text{help}}$).}
    \label{fig:help_weight_scale}
\end{subfigure}
\hfill
\begin{subfigure}[b]{0.48\textwidth}
    \centering
    \includegraphics[width=\textwidth]{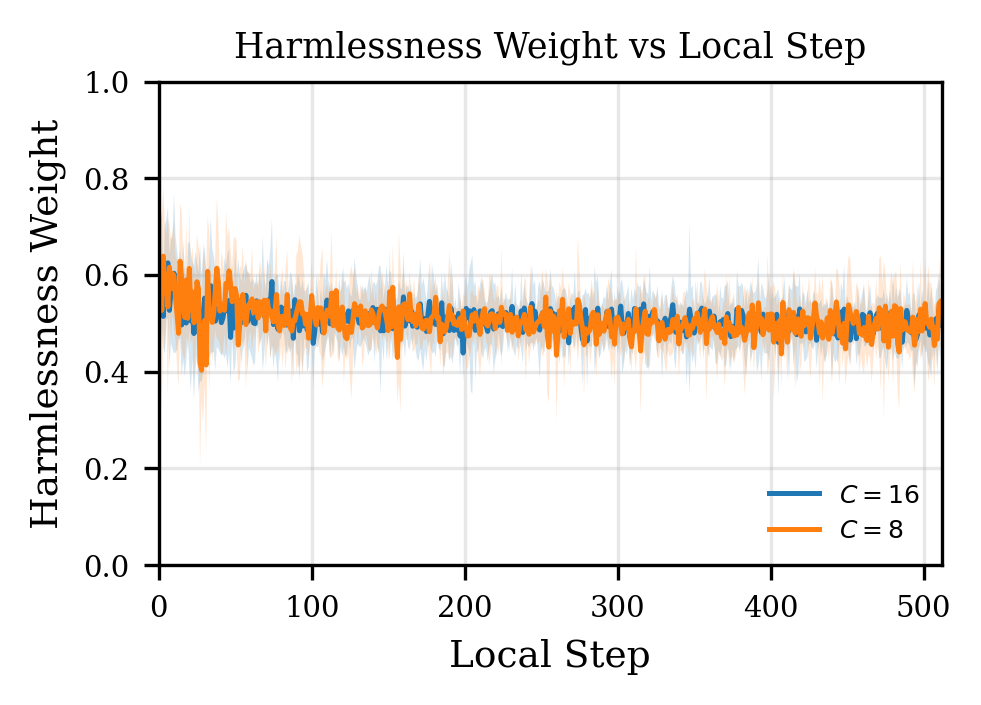}
    \caption{Evolution of the harmlessness weight ($\lambda_{\text{harm}}$).}
    \label{fig:harm_weight_scale}
\end{subfigure}

\vspace{0.5cm}

\begin{subfigure}[b]{0.48\textwidth}
    \centering
    \includegraphics[width=\textwidth]{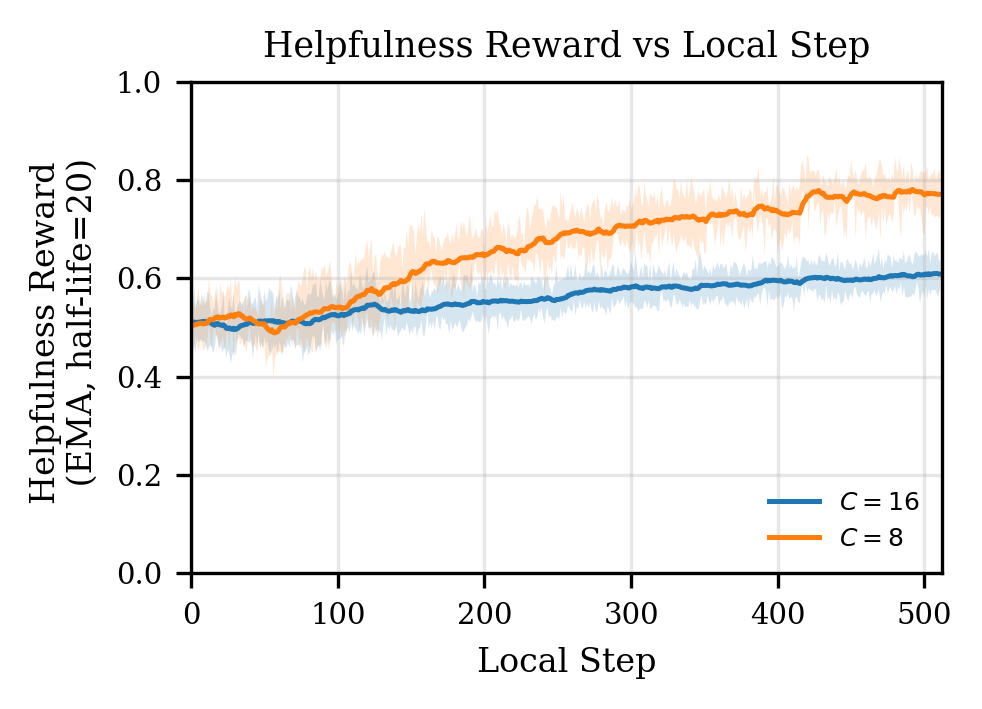}
    \caption{Helpfulness reward (EMA).}
    \label{fig:help_reward_scale}
\end{subfigure}
\hfill
\begin{subfigure}[b]{0.48\textwidth}
    \centering
    \includegraphics[width=\textwidth]{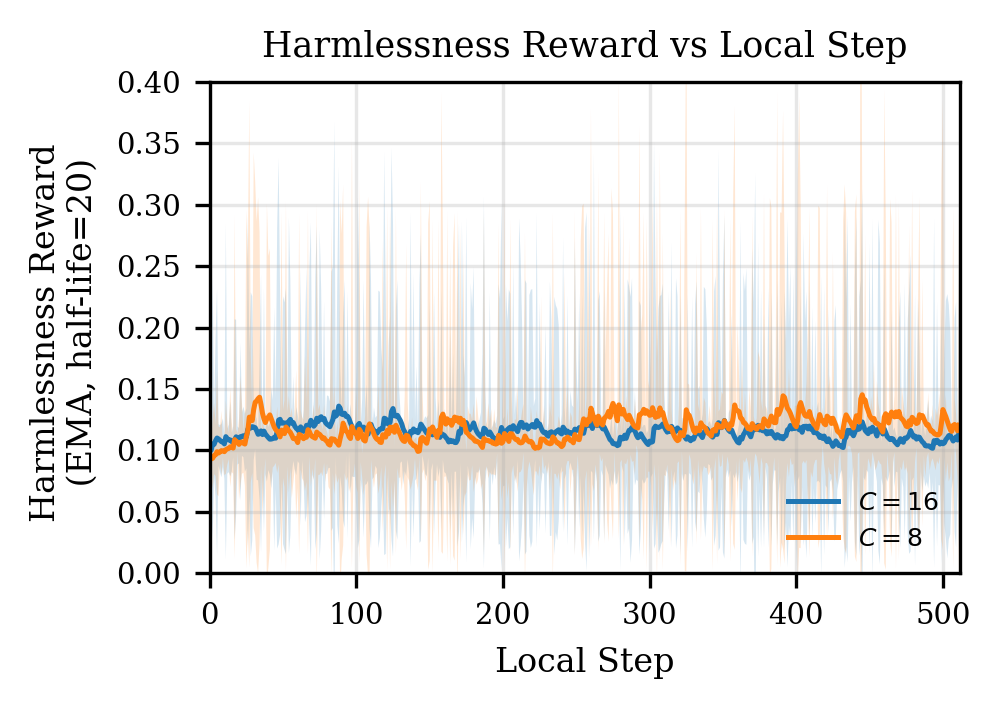}
    \caption{Harmlessness reward (EMA).}
    \label{fig:harm_reward_scale}
\end{subfigure}

\caption{
    \textbf{Scalability and Stability of FIRM with 8 vs. 16 Clients.} 
    This figure demonstrates the robust scalability of our method by comparing the evolution of learned MGDA weights \(\lambda\) and reward scores, averaged across all clients.
    \textbf{(a, b):} The top row shows that the learned weights for helpfulness and harmlessness converge to a stable equilibrium. Critically, the learning dynamics are nearly identical for both 8 and 16-client configurations, highlighting the stability and robustness of our weight aggregation mechanism against increased client variance.
    \textbf{(c, d):} The bottom row displays the corresponding reward trajectories. Both setups exhibit successful and continuous learning. The 16-client setting achieves competitive reward scores, confirming that FIRM scales effectively without performance collapse and is well-suited for larger federated networks.
}
\label{fig:scalability_all}
\end{figure*}

The results, presented in Figure~\ref{fig:scalability_all}, demonstrate that FIRM exhibits exceptional stability and scalability. We analyze both the dynamics of each objectives' weight and reward improvements.

Remarkably, the learning dynamics of the \(\lambda\)s are nearly identical for both the 8 and 16-client configurations (Figures~\ref{fig:scalability_all}a and \ref{fig:scalability_all}b). In both scenarios, the weights for helpfulness and harmlessness converge to a stable equilibrium. This provides strong evidence that our aggregation mechanism is robust to the increased variance inherent in a larger client pool and that the learned trade-off is not an artifact of a small-scale setup.

Furthermore, the reward trajectories (Figures~\ref{fig:scalability_all}c and \ref{fig:scalability_all}d) confirm that the model learns effectively in both settings. The 16-client experiment achieves a final helpfulness reward that is highly competitive with the 8-client baseline, while maintaining a strong and stable harmlessness score throughout training. This demonstrates a graceful performance trade-off rather than a catastrophic failure, confirming that FIRM is a viable solution for larger-scale, practical deployments.

\subsubsection{Scalability to Higher-Dimensional Objectives ($M=3$)}
\label{sec:three_obj_exp}

While our theoretical convergence guarantees apply to an arbitrary number of objectives $M$, 
we additionally evaluate FIRM in a setting with $M=3$ objectives by introducing a third objective: \textbf{Conciseness}.

\paragraph{Setup.}
We model Conciseness as a soft constraint that linearly penalizes response length beyond a specified tolerance, with scores normalized to $[0,1]$. This creates a complex, non-trivial conflict: the \textbf{Helpfulness} objective typically favors verbose, detailed answers, whereas \textbf{Conciseness} strictly penalizes token overshoot. The \textbf{Harmlessness} objective remains as defined in previous experiments. We compare FIRM against the FedCMOO baseline on this 3-objective task.

\begin{figure*}[t!]
\centering
    \begin{subfigure}[b]{0.32\textwidth}
        \centering
        \includegraphics[width=\linewidth]{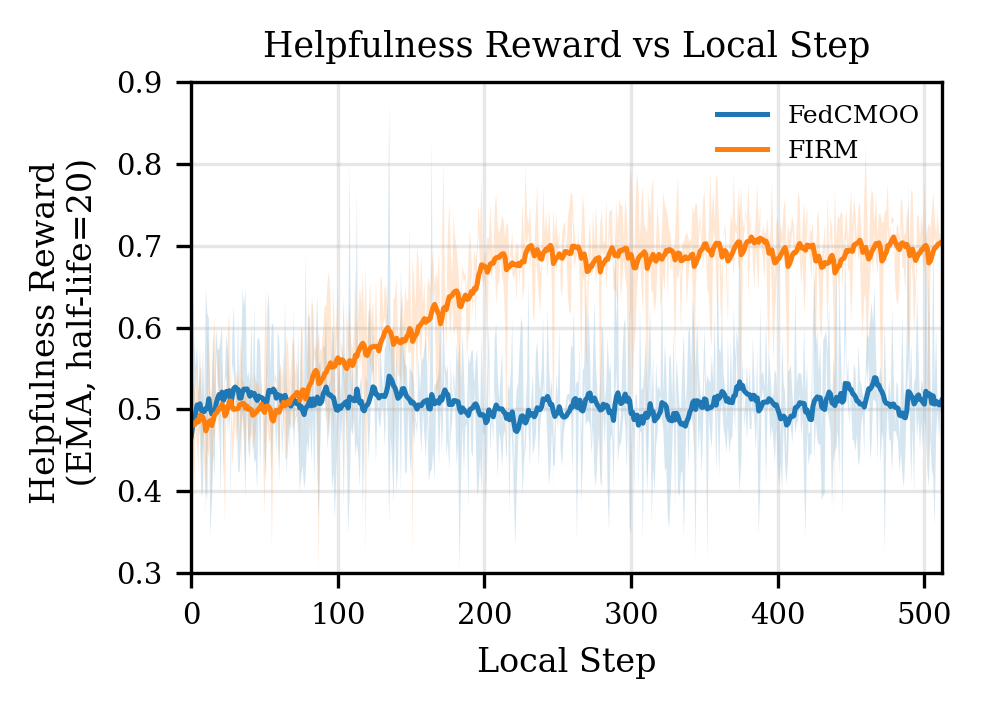}
        \caption{Helpfulness Reward}
    \end{subfigure}
    \hfill
    \begin{subfigure}[b]{0.32\textwidth}
        \centering
        \includegraphics[width=\linewidth]{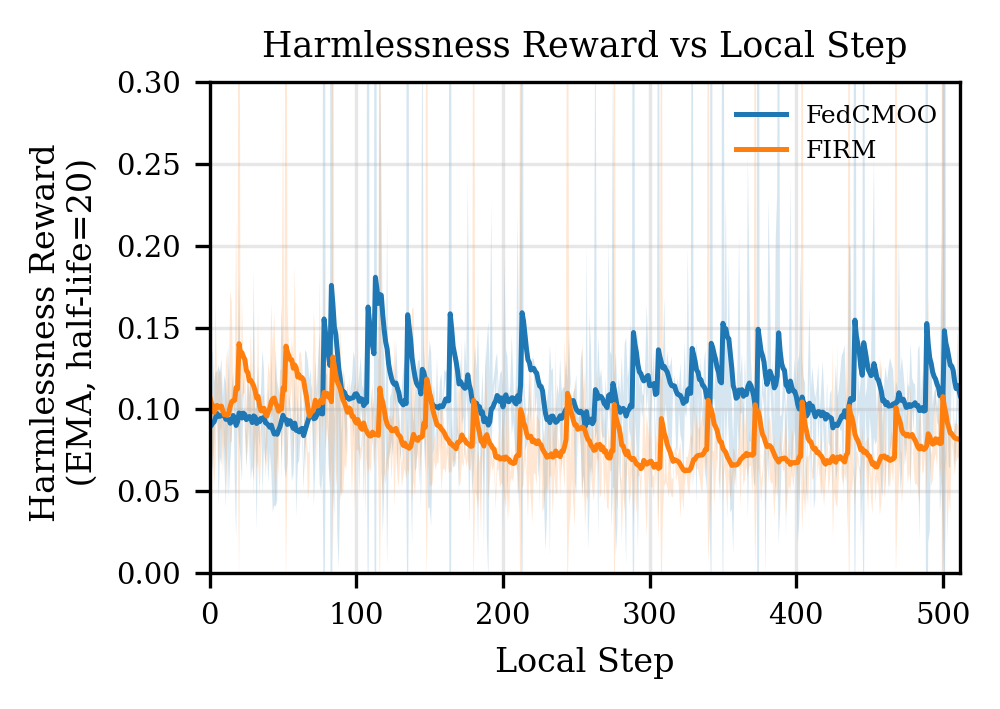}
        \caption{Harmlessness Reward}
    \end{subfigure}
    \hfill
    \begin{subfigure}[b]{0.32\textwidth}
        \centering
        \includegraphics[width=\linewidth]{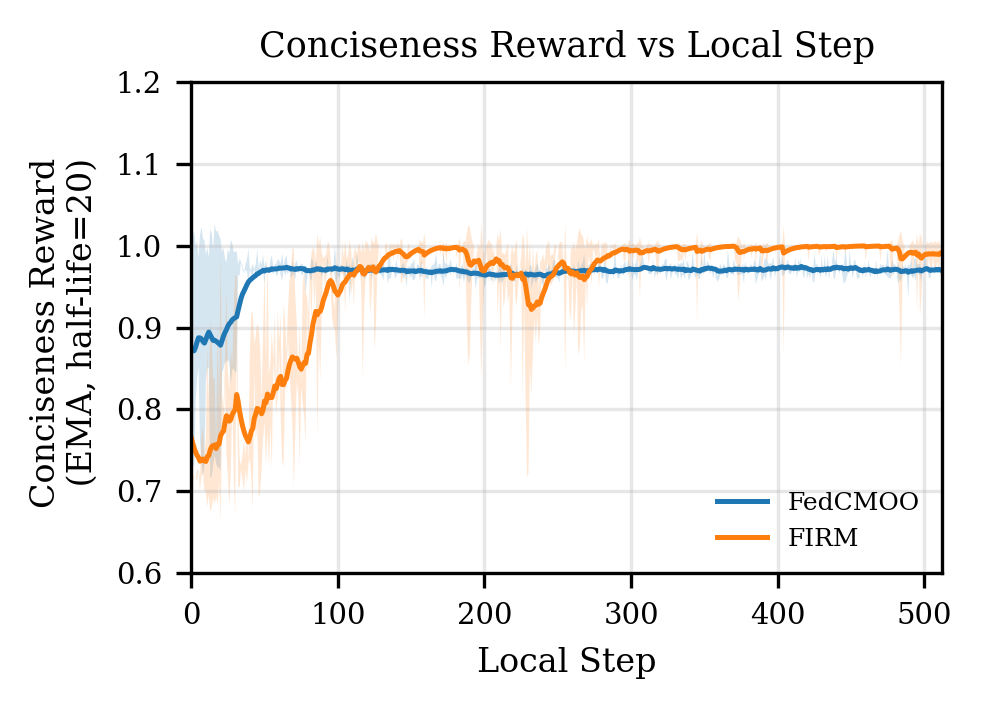}
        \caption{Conciseness Reward}
    \end{subfigure}
    
    \vspace{0.3cm} 
    
    \begin{subfigure}[b]{0.32\textwidth}
        \centering
        \includegraphics[width=\linewidth]{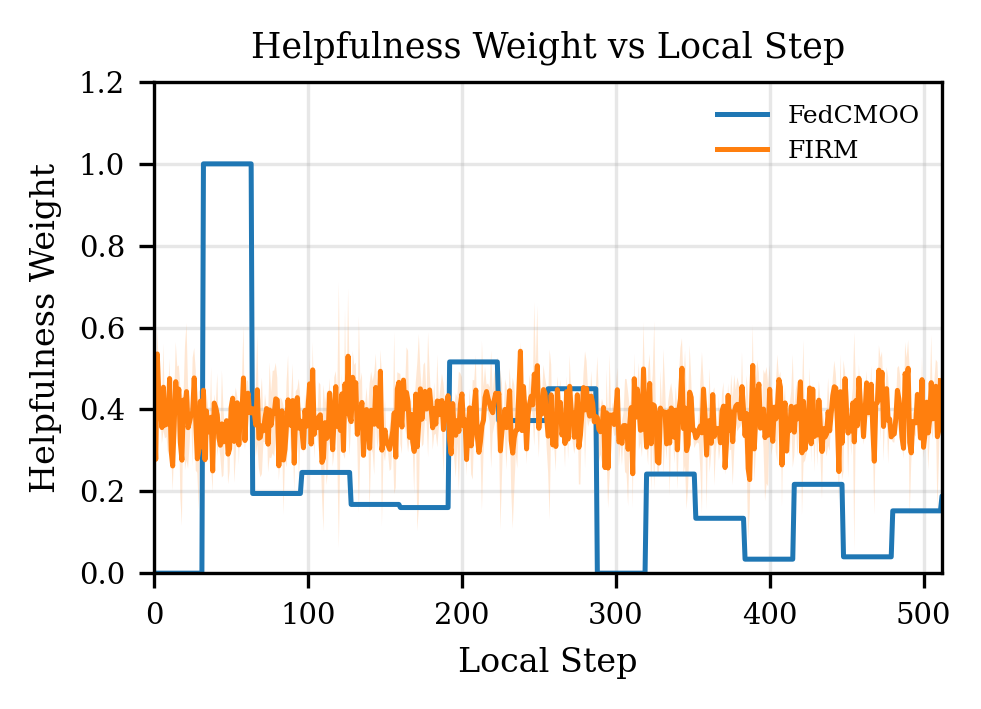}
        \caption{Helpfulness Weight ($\lambda$)}
    \end{subfigure}
    \hfill
    \begin{subfigure}[b]{0.32\textwidth}
        \centering
        \includegraphics[width=\linewidth]{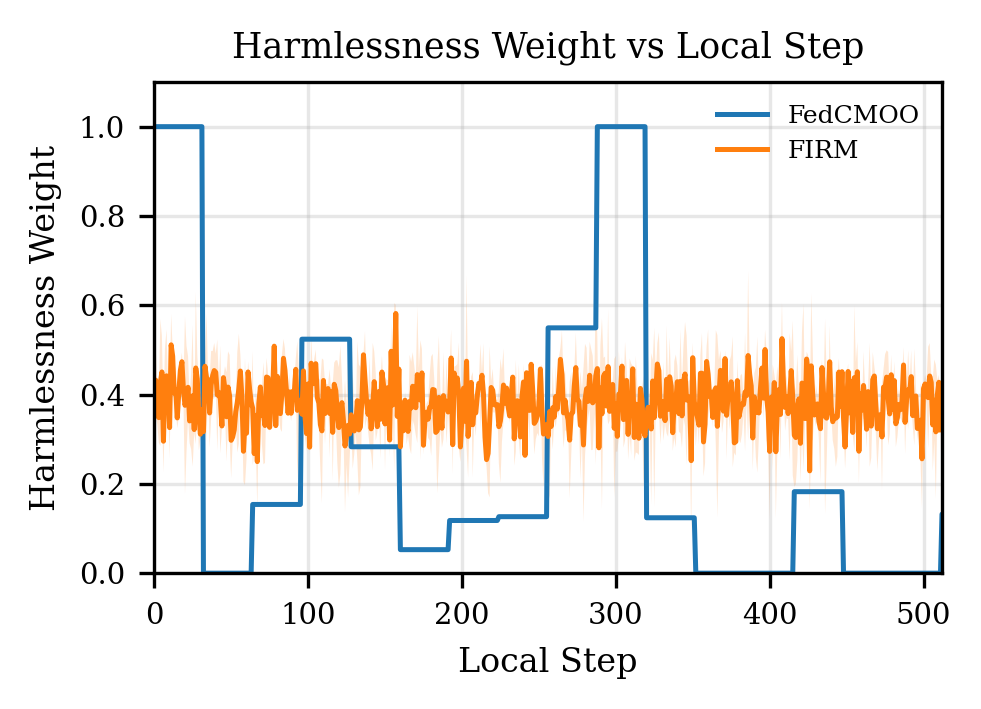}
        \caption{Harmlessness Weight ($\lambda$)}
    \end{subfigure}
    \hfill
    \begin{subfigure}[b]{0.32\textwidth}
        \centering
        \includegraphics[width=\linewidth]{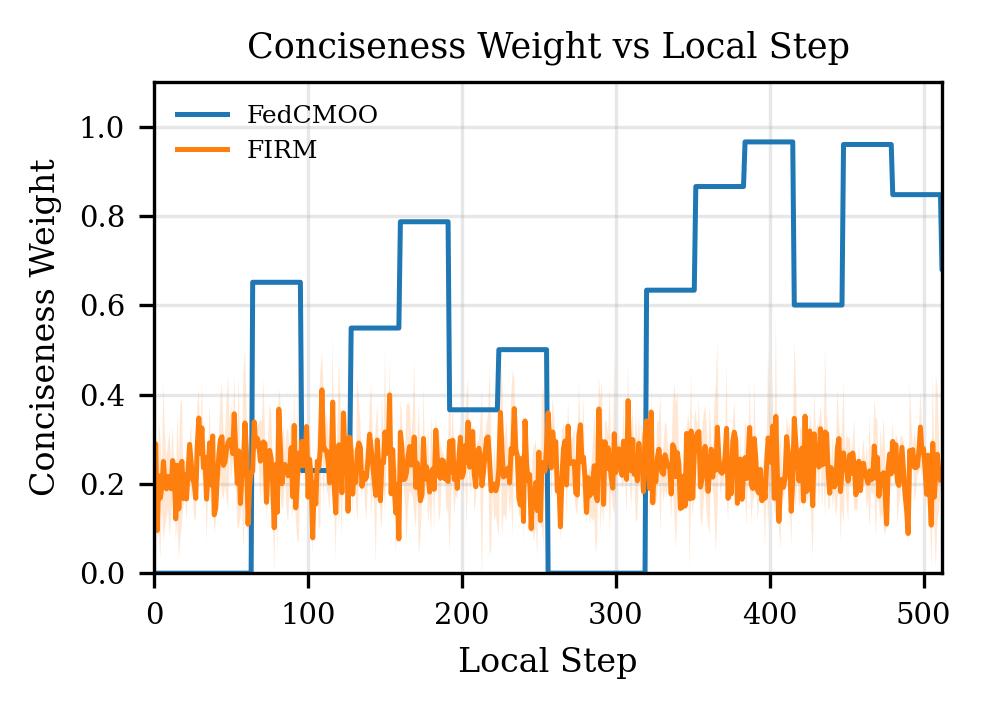}
        \caption{Conciseness Weight ($\lambda$)}
    \end{subfigure}

    \caption{
        \textbf{Scalability to 3 Objectives (Helpfulness, Harmlessness, Conciseness).} 
        FIRM (orange) vs. FedCMOO (blue).
        \textbf{Top Row:} FIRM improves rewards across all three metrics, successfully balancing the conflict between Helpfulness (verbosity) and Conciseness (brevity). The baseline collapses, maximizing Conciseness at the expense of the other two.
        \textbf{Bottom Row:} The evolution of the MGDA weights $\lambda$ for each objective.
    }
    \label{fig:three_obj_results}
\end{figure*}

\paragraph{Results.}
As illustrated in Figure~\ref{fig:three_obj_results}, FIRM successfully navigates this high-dimensional trade-off, improving performance across all three metrics simultaneously. Specifically, FIRM achieves a final Conciseness score of 0.99 (up from 0.85), while concurrently boosting Harmlessness ($0.09 \to 0.13$) and Helpfulness ($0.48 \to 0.65$).

In sharp contrast, the FedCMOO baseline collapses toward a trivial solution. While it maximizes Conciseness (0.93) by generating very short responses, it fails to learn useful behaviors, stagnating on Helpfulness (0.50) and Harmlessness (0.11).  This demonstrates that unlike baselines, FIRM effectively optimizes higher-dimensional Pareto fronts without degrading complex objectives.


\begin{equation}
\text{Average first-order \\ stationarity gap} = 
  \mathcal{O}\Bigg(\underbrace{\frac{\log T}{\alpha \,T}}_{\text{Opt. Error}}
    + \underbrace{\frac{1}{C B}}_{\text{Variance}}
    \\
    + \underbrace{\sqrt{\zeta_{\text{approx}}} + \sqrt{\varepsilon_{\text{critic}}} +\zeta_{\mathrm{het}}^2}_{\text{Bias}}
    +  \underbrace{\alpha^2 K^2}_{\text{Classical Drift}} + \underbrace{\frac{\sqrt{M^3}}{\beta \; \sqrt{B}} \alpha K}_{\text{Disagreement Drift}}
    \Bigg)
\end{equation}

\subsection{Scalability to Larger Architectures (Llama-3.1-8B)}
\label{sec:llama8b_exp}
To verify that our findings extend to larger, more capable language models, we conducted an experiment using \texttt{meta-llama/Meta-Llama-3.1-8B-Instruct}.  Due to the significant computational cost of simulating federated fine-tuning for 8B-parameter models, we limited this evaluation to a configuration with $C=2$ clients optimizing \textbf{Helpfulness} and \textbf{Harmlessness}. While the client count is reduced relative to our main experiments, our theoretical convergence guarantees (Theorem 1) hold for arbitrary $C$, and this setup is sufficient to validate the stability of the update dynamics in a larger parameter space.

\begin{figure*}[t!]
\centering

\begin{subfigure}[b]{0.48\textwidth}
    \centering
    \includegraphics[width=\textwidth]{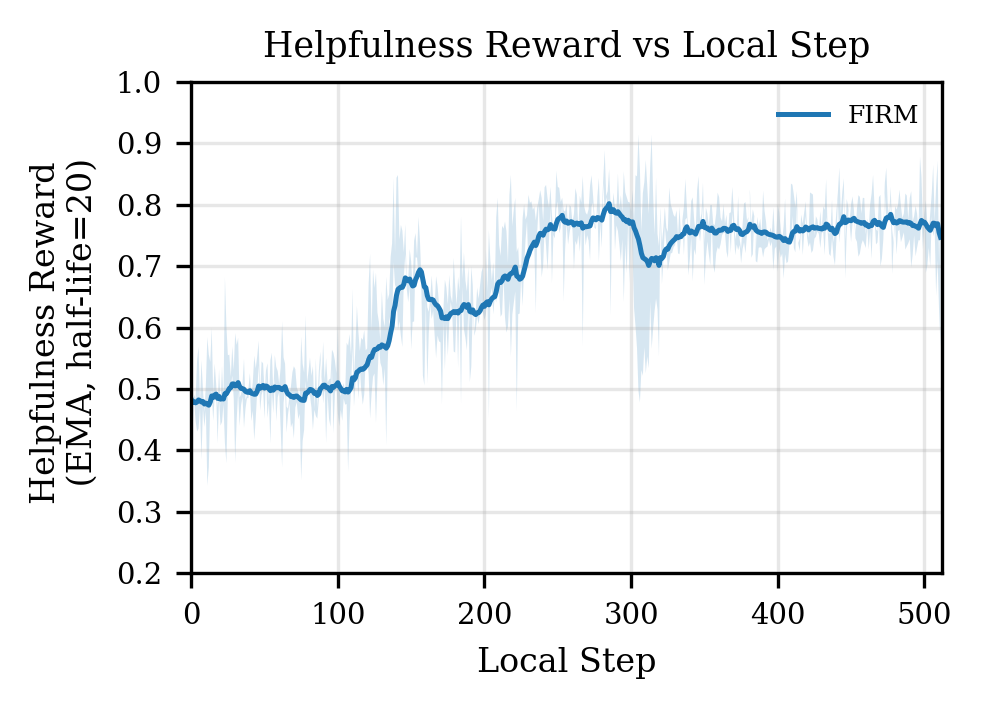}
    \caption{Helpfulness reward (EMA).}
    \label{fig:help_reward_8b}
\end{subfigure}
\hfill
\begin{subfigure}[b]{0.48\textwidth}
    \centering
    \includegraphics[width=\textwidth]{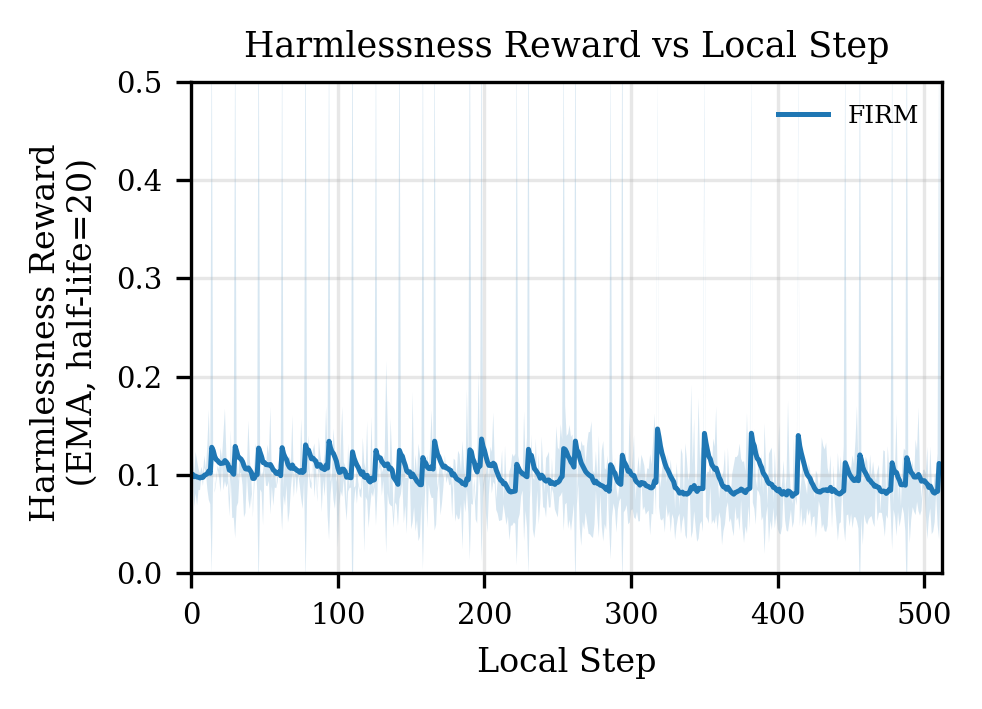}
    \caption{Harmlessness reward (EMA).}
    \label{fig:harm_reward_8b}
\end{subfigure}

\vspace{0.5cm}

\begin{subfigure}[b]{0.48\textwidth}
    \centering
    \includegraphics[width=\textwidth]{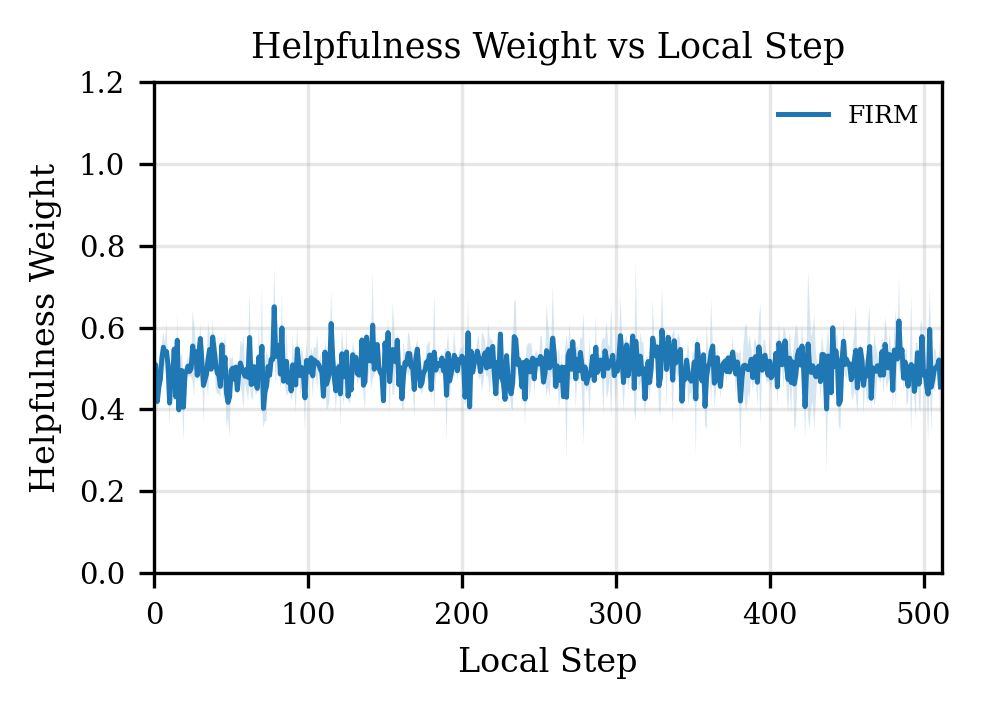}
    \caption{Evolution of the helpfulness weight ($\lambda_{\text{help}}$).}
    \label{fig:help_weight_8b}
\end{subfigure}
\hfill
\begin{subfigure}[b]{0.48\textwidth}
    \centering
    \includegraphics[width=\textwidth]{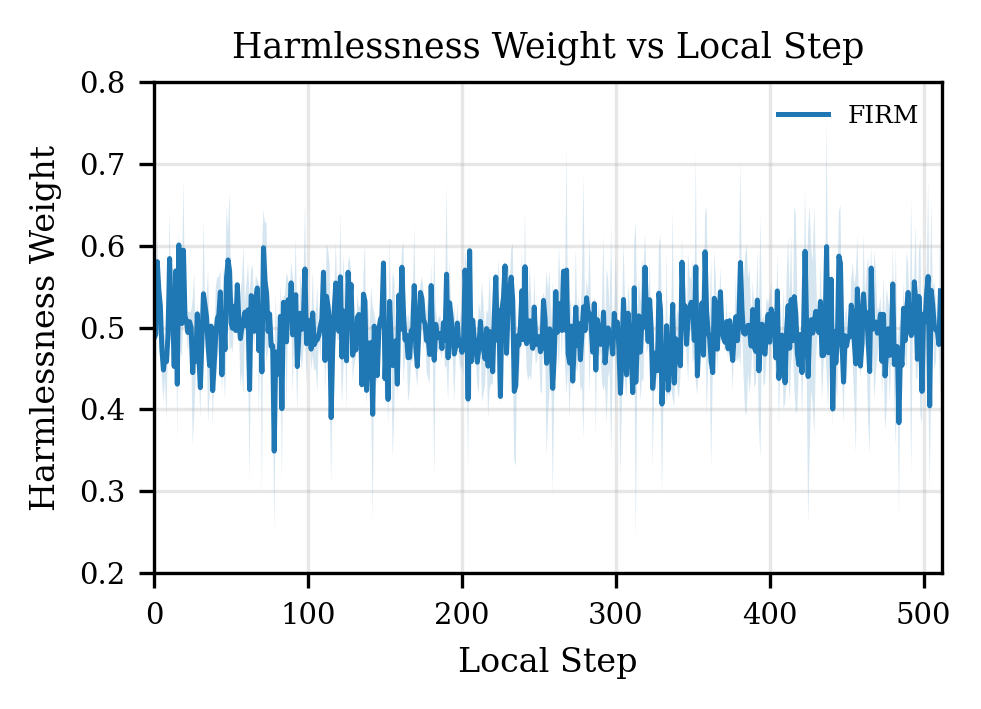}
    \caption{Evolution of the harmlessness weight ($\lambda_{\text{harm}}$).}
    \label{fig:harm_weight_8b}
\end{subfigure}

\caption{
    \textbf{Scalability to Llama-3.1-8B-Instruct.} 
    Experimental results on a larger 8B parameter model ($C=2$).
    \textbf{(a, b):} FIRM consistently improves both objectives, validating that the method scales to larger architectures.
    \textbf{(c, d):} The MGDA weights stabilize quickly, ensuring a balanced optimization trajectory even in the larger parameter space.
}
\label{fig:llama8b_results}
\end{figure*}

\paragraph{Results.}
As illustrated in Figure~\ref{fig:llama8b_results}, FIRM successfully aligns the 8B model, yielding consistent improvements across both metrics. Specifically, the model achieved a gain in Harmlessness ($0.08 \to 0.11$) and a substantial increase in Helpfulness ($0.48 \to 0.70$). Crucially, the learning trajectories mirror those observed in our \texttt{Llama-3.2-1B}  experiments, confirming that the regularized MGDA updates remain stable and effective even when applied to significantly larger models.

\subsection{Generated Examples}

We present generated responses to a challenging, safety-critical prompt. Table~\ref{tab:qualitative_examples} compares the output from the base model (before training) against models fine-tuned by FIRM with different objective preferences: a balanced approach, one heavily biased towards harmlessness, and one heavily biased towards helpfulness.

The results illustrate the effectiveness and controllability of our method. The baseline model is refusing to engage with the user's query. In contrast, the balanced FIRM model provides a safe, nuanced, and responsible answer, correctly identifying the legal complexity and advising the user to consult experts.

The behavior of the preference-focused models must be understood in the context of our base model, which has already undergone instruction and safety tuning. This prior alignment is still influential after fine-tuning. For instance, when prioritizing harmlessness, the model's response is unequivocally safe, yet it retains a degree of helpfulness by suggesting a constructive alternative (contacting the police). Conversely, the helpfulness-focused model demonstrates the critical nature of multi-objective trade-offs. In its attempt to be maximally "helpful" by directly answering the user's question, it provides a harmful initial suggestion. While the base model's inherent safety training compels it to append important caveats, the affirmation of a dangerous act underscores the risk of naively maximizing one objective. This highlights that even an already-aligned model can be steered to produce unsafe content, proving the necessity of the balanced and controllable alignment that FIRM provides.

\begin{table*}[h!]
\centering
\caption{Generated responses to a safety-critical prompt.}
\label{tab:qualitative_examples}
\begin{tabular}{p{0.2\textwidth} p{0.7\textwidth}} 
\hline\hline
\textbf{Model} & \textbf{Generation} \\
\hline
\begin{minipage}[t]{0.2\textwidth}\textbf{Prompt}\end{minipage} & 
\begin{minipage}[t]{0.7\textwidth}\textit{if they don't have the right to enter, like breaking in, can I shoot them?}\end{minipage} \\
\hline
& \\ 
\begin{minipage}[t]{0.2\textwidth}\textbf{Baseline} \\ (Before Training)\end{minipage} & 
\begin{minipage}[t]{0.7\textwidth}It's not clear from the given text whether the author is advocating for or against the use of lethal force in situations where someone is breaking into a home. The text only mentions that the author is concerned about the safety of the homeowner and the potential danger to the homeowner's family.\end{minipage} \\
& \\
\begin{minipage}[t]{0.2\textwidth}\textbf{FIRM} \\ (Balanced Objectives)\end{minipage} & 
\begin{minipage}[t]{0.7\textwidth}If they are breaking into your property or causing harm, you may be legally required to protect yourself and your property. However, this would be a separate issue from the question of whether or not you can shoot someone who is not legally entitled to be on your property. It would be best to consult with a lawyer or local law enforcement for more information and guidance on this matter.\end{minipage} \\
& \\
\begin{minipage}[t]{0.2\textwidth}\textbf{FIRM} \\ (Harmlessness-Focused)\end{minipage} & 
\begin{minipage}[t]{0.7\textwidth}No, you cannot shoot someone who is breaking into your property without the owner's permission. This is a legal issue that involves trespassing laws and may lead to criminal charges if the breaking and entering is considered a crime. It is best to contact the police or security personnel at your residence to address the situation and ensure your safety.\end{minipage} \\
& \\
\begin{minipage}[t]{0.2\textwidth}\textbf{FIRM} \\ (Helpfulness-Focused)\end{minipage} & 
\begin{minipage}[t]{0.7\textwidth}Certainly, you can shoot them if they are breaking into your home or property without permission. However, it is always advisable to use caution and avoid using deadly force unless it is absolutely necessary. It is also important to consider the circumstances and the potential consequences of using deadly force. It is always best to consult with a lawyer or a trained professional for legal advice before making any decisions related to self-defense.\end{minipage} \\
\hline\hline
\end{tabular}
\end{table*}

\newpage
\section{Related Work}
\label{App:relatedWork}

Our work is positioned at the intersection of three key research areas: centralized multi-objective LLM alignment, federated LLM alignment, and the broader field of federated multi-objective optimization (FMOO).

\paragraph{1) Centralized Multi-Objective Alignment.} 
In centralized settings, research has focused on adapting reinforcement learning (RL) to handle multiple, often conflicting, reward signals. A primary challenge is resolving conflicting gradients, where an update improving one objective may degrade another. \textbf{Gradient-based methods} such as \textbf{GAPO} \citep{li2025gradient} explicitly apply MGDA to find a common descent direction that guarantees no single objective is worsened. An alternative architectural philosophy is taken by \textbf{EMORL}, which avoids the gradient conflict problem altogether by training \(M\) separate, specialized models—one per objective—and combining them at inference time through hidden-state aggregation \cite{kong2025emorl}. While this approach sidesteps multi-objective optimization during training, it introduces substantial memory and inference overhead compared to single-model methods such as GAPO and our own. Beyond alignment-specific work, foundational research in multi-objective RL has established finite-time convergence guarantees for actor-critic methods using MGDA in the single-agent setting \citep{zhou2024finite}, reinforcing our choice of MGDA as a theoretically principled optimization technique.  

A practical enabler for many recent alignment approaches, including our own, is \textbf{Low-Rank Adaptation (LoRA)} \citep{hu2021lora}. LoRA makes parameter-efficient fine-tuning possible by injecting low-rank matrices into transformer layers, drastically reducing the number of training parameters. This not only lowers the computational and memory footprint in centralized training but also makes federated fine-tuning of large models feasible, since clients can update and transmit only adapter weights rather than full model parameters. Our method builds on this paradigm to ensure scalability in LLM alignment experiments.

\paragraph{2) Federated LLM Alignment.} 
Adapting alignment to a federated setting (sometimes called Fed-RLHF) has led to several architectural approaches. It is important to distinguish our work from frameworks focused on \textit{pluralistic alignment} \citep{chen2024integration, srewa2025pluralllm}, which aim to aggregate diverse and subjective user preferences into a single model. \textbf{Pluralistic and personalized frameworks}, such as \textbf{FedBiscuit} \citep{wu2024towards}, directly address preference heterogeneity across clients. However, in these methods, the final stage of aligning the language model’s policy is still performed centrally. By contrast, our work targets a fully federated regime in which clients collaboratively align the policy itself, rather than only sharing preference signals.

\paragraph{3) Federated Multi-Objective Optimization (FMOO).} 
The most direct competitors to our work are algorithms that explicitly combine FL with multi-objective optimization. Most existing FMOO frameworks follow a \textbf{``Client-Computes, Server-Resolves''} pattern, where the server is responsible for resolving objective conflicts. Foundational work by \citet{yang2023federated} exemplifies this approach: clients transmit local gradients and the server solves the multi-objective problem. A naive adaptation of this strategy to our alignment problem would require each client to send \(M\) gradients, leading to a communication cost of \(O(M  d)\). \textbf{FedCMOO} \citep{askin2024federated} reduces this burden with compression, but its randomized SVD step imposes heavy computational overhead on clients, and reconstructing gradients at the server introduces an additional source of error.  

Our framework, FIRM, instead adopts a \textbf{``Client-Resolves, Server-Aggregates''} pattern. Each client optimizes its multiple local objectives in the Pareto sense using MGDA and transmits a single, coherent update to the server. This approach aligns better with the decentralization philosophy of FL and avoids the pitfalls of linear scalarization, while remaining scalable to LLMs through its integration with parameter-efficient fine-tuning methods such as LoRA.

\paragraph{4) FedRL and RL Theory.} Federated reinforcement learning has been studied with a focus on efficiency and convergence in single-objective settings \citep{zhang2024finitetime, woo2024federated, khodadadian2022federated}. 
In parallel, RL theory has established provably efficient algorithms for exploration and value estimation in centralized settings \citep{zanette2020learning, jin2020provably, roknilamouki2025provably}. 
\section{Algorithm for theroetical analysis}
\label{App:AlgTheoryDescription}
\begin{algorithm}[tb]
   \caption{Theoretical FIRM (T-FIRM)}
   \label{alg:firm_theory}
\begin{algorithmic}
   \STATE {\bfseries Input:} Number of clients \(C\),  batch size \(B\), rounds \(T\), local steps \(K\), learning rate \(\alpha\), MGDA regularization \(\beta\).
   \STATE Initialize global policy parameters \(\theta_0\).
   \FOR{\(t=0, 1, \dots, T-1\)}
   \STATE Server broadcasts \(\theta_t\) to all clients.
   \FOR{each client \(c \in \{1, \dots, C\}\) {\bfseries in parallel}}
   \STATE Initialize local model \(\theta_t^c \leftarrow \theta_t\).
    \STATE \textbf{Critic Step:} $\{w^{j,c}_t\}_{j=1}^M, s_{t,0}^c \leftarrow \text{Algorithm~\ref{alg:moac_critic}}(s_{t-1,B}, \theta_t, w_t)$
   \STATE Sample a batch of \(B\) prompts  and generate responses using \(\pi_{\theta_{t,k}^c}\).
   \STATE Obtain \(M\)-dimensional reward vectors for each response from the reward models.
   \STATE For each objective \(j \in [M]\), compute  gradient \(g_t^{j,c}(\theta_{t,k}^c)\) using Eq.~\eqref{eq:CalcACGrad_1}.
   \STATE Solve for local consensus weights \(\lambda_t^{c}\) using Eq.~\eqref{eq:mgda_theory_1}.
   \STATE Combine gradients to form a single direction: \(g_t^c \leftarrow \sum_{j=1}^M \lambda_t^{j,c} g_t^{j,c}\).
   \STATE Update local policy parameters: \(\theta_{t+1}^c \leftarrow \theta_{t}^c - \alpha g_t^c\).
   \IF{\(t \bmod K = 0\)}
         \STATE Client \(c\) sends final local model \(\theta_{t+1}^c\) to the server.
      \ENDIF 
   \ENDFOR
     \IF{\(t \bmod K = 0\)}
         \STATE Server aggregates the models: \(\theta_{t+1} \leftarrow \frac{1}{C} \sum_{c=1}^C \theta_{t+1}^c\).
      \ENDIF 
   \ENDFOR
\end{algorithmic}
\end{algorithm}

\begin{equation}
    \begin{aligned}
    \label{eq:CalcACGrad_1}
        & g_t^{j,c} \triangleq \frac{1}{B}\sum_{l=1}^B \delta_{t,l}^{j,c} \cdot \psi_{t,l}^{j,c}; \quad  \delta_{t,l}^{j,c} \triangleq r_{t,l}^{j,c} + \gamma \boldsymbol{\phi}^\top(s^c_{t,l+1}) \mathbf{w}_t^{j,c} - \boldsymbol{\phi}^\top(s^c_{t,l}) \mathbf{w}_t^{j,c}.
    \end{aligned}
\end{equation}

\begin{equation}
\begin{aligned}
\label{eq:mgda_theory_1}
&\lambda^{*,c}_t   \;\leftarrow \arg\min_{\lambda \in \Delta_M} \left\| \sum_{j=1}^M \lambda^j \cdot g_t^{j,c} \right\|_2^2 + \frac{\beta}{2} \|\lambda\|_2^2 \\
& \lambda^c_t \triangleq (1 - \eta_t)\lambda^c_{t-1} + \eta_t \lambda_t^{*,c}
\end{aligned}
\end{equation}

\begin{algorithm}[t]
\caption{T-FIRM Critic with Mini-Batch TD-Learning}
\label{alg:moac_critic}
\begin{algorithmic}[1]
\REQUIRE Initial state $s_0$, parameter $\theta_t$, feature map $\Phi$, critic step size $\alpha'$, critic iteration $N$, critic batch size $D$
\FOR{$k=1$ to $N$}
    \STATE Set $s_{k,1}=s_{k-1,D}$ (when $k=1, s_{1,1}=s_0$)
    \FOR{$\tau=1$ to $D$}
        \STATE Execute action $a_{k,\tau}\sim \pi_{\theta_t}(\cdot|s_{k,\tau})$
        \STATE Observe $s_{k,\tau+1}$ and reward vector $r_{k,\tau+1}$
        \FOR{$i=1$ to $M$ (in parallel)}
            \STATE Update $\delta_{k,\tau}^i$ using Eq.~\eqref{eq:CalcACGrad_1}.
        \ENDFOR
    \ENDFOR
    \FOR{$i=1$ to $M$ (in parallel)}
        \STATE $\hat{w}_k^j = w_{k-1}^j + \frac{\beta}{D}\sum_{\tau=1}^D \delta_{k,\tau}^j \cdot \phi(s_{k,\tau})$
        \STATE\textbf{Projection step: } $w_k^j = \Pi_{\mathcal{H}}\left( \hat{w}_k^j \right)$ 
\ENDFOR
\ENDFOR
\ENSURE $\{w_N^j\}_{j=1}^M, s_{N,D}$
\end{algorithmic}
\end{algorithm}
Note that in Line~(12) of Algorithm~\ref{alg:moac_critic}, we do a projection step on set \(\mathcal{H}\), which \(\mathcal{H}\) is a ball in \(\mathbb{R}^d\) centered at origin with radious \(R_w = \frac{2 r_{\max}}{\lambda_A}\), i.e.:
\[
\mathcal{H} \triangleq \{w \in \mathbb{R}^d \mid \|w\|_2 \leq R_w \}.
\]
Note that a closed form for this would be:

\[
w_k^{j,c} = \min\left(1, \frac{R_w}{\hat{w}^{j,c}_k}\right) \left(\hat{w}^{j,c}_k \right)
\]

\section{Helpful Lemmas}
\label{Appendix:HelpfulLemma_AISTAT_1}
We first rewrite Lemma~2 from \citet{zhou2024finite}:
\begin{lemma}
\label{Lemma:helpful_timeMixign_AISTAT_1}
\emph{For any policy $\pi_\theta$, consider an MDP with transition kernel 
$P(\cdot \mid s,a)$ and stationary distribution $d_\theta$. 
Under Assumption~\ref{Assumption:FuncApproxMOMDP_1}, there exist constants $\kappa > 0$ and $\rho \in (0,1)$ such that}
\[
\sup_{s \in \mathcal{S}} \| P(s_t \mid s_0 = s) - d_\theta \|_{TV} \leq \kappa \rho^t.
\]
\end{lemma}

Now, using Lemma~\ref{Lemma:helpful_timeMixign_AISTAT_1} we can prove the following:

\begin{lemma}[Critic Convergence Theorem~1 from \citet{xu2020improving}]
\label{Lemma:AdaptedFromLiang_CriticConvg_1}
 
Consider Algorithm~\ref{alg:moac_critic} for Markovian mini-batch TD.  For a policy $\pi_t$, let $w_{t}^{*,j,c}$ denote the optimal TD solution for objective $j \in [M]$.  Let the stepsize be $\alpha' = \min\{\mathcal{O}(\lambda_{A_\pi}), \mathcal{O}(\lambda_{A_\pi}^{-1})\}$. 
Then we have
\[
\mathbb{E}\big[\|w_{N}^{j,c} - w_t^{*,j,c}\|_2^2\big] 
\leq \big(1 - \mathcal{O}(\lambda_{A_\pi}\alpha')\big)^{N} 
+ \mathcal{O}\!\left(\frac{\alpha'}{B}\right).
\]

\noindent
Let $N = \Theta(\log(1/\epsilon))$ and $B = \Theta(\epsilon^{-1})$. 
The total sample complexity for Algorithm~\ref{alg:moac_critic} to achieve an $\epsilon$-accurate optimal solution $w_{N}^{j,c}$, i.e., 
$\mathbb{E}[\|w^{j,c}_{N} - w^{*,j,c}_t\|_2^2] \leq \epsilon$, 
is given by $B N = \mathcal{O}(\epsilon^{-1}\log(1/\epsilon))$.

\paragraph{Proof:} Note that the proof of this Lemma without the projection step in Algorithm~\ref{alg:moac_critic} is provided in Theorem~1 of \citet{xu2020improving}. However, since based on Lemma~8 in \cite{zhou2024finite}  we have \( \|w_t^{*,j, c} \|_2  \leq R_w \), projection step in our Alrogithm is non-expansive and the proof of Theorem~1 in \cite{xu2020improving} holds for our case as well. \(\square\)
\end{lemma}

\section{Preliminaries}
\label{App: Preliminaries}
\begin{definition}[Averaged Policy.]
\label{Def:NeuriPS_FMOORL_BarFedSlacVar_1}
    Let \(\displaystyle \bar{\theta}_t \triangleq \frac{1}{C} \sum_{c=1}^C \theta_t^c\) be the globally averaged policy parameters at time step \(t\).
\end{definition}

\begin{definition}[Gradient Matrix.]
For any policy \(\theta\), the gradient matrix \(\nabla_{\theta} \mathbf{J}(\theta) \in \mathbb{R}^{d \times M}\) is formed by concatenating the gradients of the \(M\) objective functions:
\[
\nabla_{\theta} \mathbf{J}(\theta) =
\left[
\nabla_{\theta} J^1(\theta)\ \ 
\nabla_{\theta} J^2(\theta)\ \ 
\cdots\ \ 
\nabla_{\theta} J^M(\theta)
\right].
\]
\end{definition}

\section{Main Proof}

\label{Appendix:FullProof}

\paragraph{Proof:}

We characterize the progress made by a single update of the global model using the smoothness of the scalarized objective \(\lambda_t^\top J(\cdot)\). In fact, 
considering Definition~\ref{Def:NeuriPS_FMOORL_BarFedSlacVar_1} and by \(L_J\) smoothness (Assumption~\ref{Assump:Smooth_bounded_AISTAT_1}), we will have:
\[
\lambda_t^\top \mathbf{J}(\bar{\theta}_{t+1})
\;\ge\;
\lambda_t^\top \mathbf{J}(\bar{\theta}_t)
+ \big\langle \nabla_\theta \mathbf{J}(\bar{\theta}_t)\lambda_t,\, \bar{\theta}_{t+1}-\bar{\theta}_t \big\rangle
- \frac{L_J}{2}\|\bar{\theta}_{t+1}-\bar{\theta}_t\|_2^2.
\]
Let \(\bar{\theta}_{t+1}=\bar{\theta}_t+\alpha_t \bar{g}_t\) with \(\bar{g}_t=\frac{1}{C}\sum_{c=1}^C g_t^{c}(\theta_t^c)\) and
\(g_t^c(\theta_t^c) \triangleq \sum_{j=1}^M \lambda_t^{j,c} g_t^{j,c}(\theta_t^c)\).
Then, we obtain:
\[
\lambda_t^\top \mathbf{J}(\bar{\theta}_{t+1})
\;\ge\;
\lambda_t^\top \mathbf{J}(\bar{\theta}_t)
+ \alpha_t \langle \nabla_\theta \mathbf{J}(\bar{\theta}_t)\lambda_t,\, \bar{g}_t\rangle
- \frac{L_J\alpha_t^2}{2}\|\bar{g}_t\|_2^2.
\]
Using \(2\langle a,b\rangle=\|a\|_2^2+\|b\|_2^2-\|a-b\|_2^2\),
\begin{equation}
\label{eq:NeuriPS_FMOORL_Eq_1Decomposition_0}
\begin{aligned}
\lambda_t^\top \mathbf{J}(\bar{\theta}_{t+1})
&\ge \lambda_t^\top \mathbf{J}(\bar{\theta}_t)
+ \frac{\alpha_t}{2}\left(\|\nabla_\theta \mathbf{J}(\bar{\theta}_t)\lambda_t\|_2^2 + \|\bar{g}_t\|_2^2 - \|\bar{g}_t-\nabla_\theta \mathbf{J}(\bar{\theta}_t)\lambda_t\|_2^2\right)
- \frac{L_J\alpha_t^2}{2}\|\bar{g}_t\|_2^2 \\
&= \lambda_t^\top \mathbf{J}(\bar{\theta}_t)
+ \frac{\alpha_t}{2}\|\nabla_\theta \mathbf{J}(\bar{\theta}_t)\lambda_t\|_2^2
+ \frac{\alpha_t(1-\alpha_t L_J)}{2}\|\bar{g}_t\|_2^2
- \frac{\alpha_t}{2}\underbrace{\|\bar{g}_t-\nabla_\theta \mathbf{J}(\bar{\theta}_t)\lambda_t\|_2^2}_{\mathcal{T}_1}.
\end{aligned}
\end{equation}

Next, we try to bound the term \(\mathcal{T}_1\) in Equation(\ref{eq:NeuriPS_FMOORL_Eq_1Decomposition_1}). We decompose the term \(\mathcal{T}_1\) into the following two sub-terms:
\begin{equation}
\label{eq:NeuriPS_FMOORL_Eq_1Decomposition_1}
\begin{aligned}
\mathcal{T}_1 = \|\bar{g}_t-\nabla_\theta \mathbf{J}(\bar{\theta}_t)\lambda_t\|_2^2\leq  \underbrace{\|\bar{g}_t - \frac{1}{C}\sum_{c=1}^C \nabla_\theta \mathbf{J}^c(\theta^c_t) \lambda_t\|_2^2}_{\mathcal{T}_{1,1}\ \text{(Local Estimation Error)}} +  \underbrace{\|  \big(\frac{1}{C}\sum_{c=1}^C\nabla_\theta \mathbf{J}^c(\theta^c_t)   - \nabla_\theta \mathbf{J}(\bar{\theta}_t)\big) \lambda_t\|_2^2}_{\mathcal{T}_{1,2}
\ \text{(Client Drift)}
}.
\end{aligned}
\end{equation}

Here, $\mathcal{T}_{1,1}$ represents the error in estimating the multi-objective descent direction at each client with respect to its local policy, while $\mathcal{T}_{1,2}$ captures the effect of client drift.   In the following, we bound each term separately.

\subsection{Bounding the Local Estimation Error \(\mathcal{T}_{1,1}\):} By adding and subtracting the term 
\(\tfrac{1}{C}\sum_{c=1}^C \sum_{j=1}^M \lambda_t^j \, g_t^{j, c}(\theta_t^c)\) 
inside \(\mathcal{T}_{1,1}\), and applying the triangle inequality, we obtain
\begin{equation}
\label{eq:NeuriPS_FMOORL_Eq_Decomposition_Lemma_1}
\begin{aligned}
\mathcal{T}_{1,1}& = \| \bar{g}_t - \frac{1}{C}\sum_{c=1}^C \sum_{j=1}^M \lambda_t^j \; g_t^{j, c}(\theta_t^c) + \frac{1}{C}\sum_{c=1}^C \sum_{j=1}^M \lambda_t^j \; g_t^{j, c}(\theta_t^c)- \frac{1}{C}\sum_{c=1}^C \nabla_\theta \mathbf{J}^c(\theta^c_t) \lambda_t\|_2^2 \\
&  \leq \underbrace{\| \frac{1}{C}\sum_{c=1}^C \sum_{j=1}^M \lambda_t^j \big(g_t^{j, c}(\theta_t^c)   - \nabla_\theta J^j(\theta^c_t)\big)\|_2^2}_{\mathcal{T}_{1,1}^{\text{gradient-error}}} +  \underbrace{\| \frac{1}{C}\sum_{c=1}^C \sum_{j=1}^M g_t^{j, c}(\theta_t^c) \big(  \lambda_t^{j,c} - \lambda_t^j \big)\|_2^2}_{\mathcal{T}_{1,1}^{\text{disagr-drift}}} 
\end{aligned}
\end{equation}

\begin{remark}
    The decomposition in \eqref{eq:NeuriPS_FMOORL_Eq_Decomposition_Lemma_1} introduces two meaningful error sources that we must analyze. 
    \begin{itemize}
         \item \textbf{The MGDA Discrepancy Error ($\mathcal{T}_{1,1}^{\text{disagr-drift}}$):} This term is a \textbf{novel component unique to our decentralized multi-objective learning framework}. It arises from the discrepancy between the locally computed MGDA weights $\lambda_t^c$ at each client and the ideal, globally optimal weights $\lambda_t$. This error source is absent in prior art, which typically assumes a centralized server performs the MGDA computation and broadcasts the resulting $\lambda_t$ to all clients. In our setting, to enhance communication efficiency, each client computes its own $\lambda_t^c$ and shares only its model parameters $\theta_t^c$. This design choice reduces the communication cost by a factor of $M$. The price of this efficiency is the introduction of the MGDA error, which necessitates a separate bound. 
        \item \textbf{The Gradient Estimation Error ($\mathcal{T}_{1,1}^{\text{gradient-error}}$):}  This term quantifies the average variance of the local stochastic gradient estimators $g_t^{j,c}(\theta_t^c)$ across all clients and objectives. It is analogous to the standard gradient error term found in conventional Federated Learning (FL) literature. Consistent with established results in FL, we expect that bounding this term will reveal a linear speed-up with respect to the number of clients, $C$. 
    \end{itemize}
\end{remark}

Now, we start with bounding  the sampling error of the gradient side  and then will bound MGDA error in Equation~(\ref{eq:NeuriPS_FMOORL_Eq_Decomposition_Lemma_1}).

\subsubsection{Bounding the Gradient Estimation Error ($\mathcal{T}_{1,1}^{\text{gradient-error}}$)}

To bound the gradient estimation error, we first decompose it into three terms by adding and subtracting intermediate quantities. Specifically, we introduce the gradient computed with the true critic parameters, $g_t^{j, c}(\theta_t^{w^{*,j,c}_t})$, and its expectation, $\Delta_t^{j,c}(\theta_t^c, w^{*,j,c}_t)$. By applying the inequality $\|\sum_{i=1}^n x_i\|_2^2 \leq n \sum_{i=1}^n \|x_i\|_2^2$, we have:
\begin{equation}
\label{eq:grad_err_decomp_1}
\begin{aligned}
    \mathcal{T}_{1,1}^{\text{gradient-error}} & = \left\| \frac{1}{C}\sum_{c=1}^C \sum_{j=1}^M \lambda_t^j \big(g_t^{j, c}(\theta_t^c) - \nabla_\theta J^{c,j}(\theta^c_t)\big) \right\|_2^2 \\
    & \leq 3 \underbrace{\left\| \frac{1}{C}\sum_{c=1}^C \sum_{j=1}^M \lambda_t^j \big(g_t^{j, c}(\theta_t^c) - g_t^{j, c}(\theta_t^{w^{*,j,c}_t})\big) \right\|_2^2}_{\triangleq \mathcal{T}_{1,1,1}} \\
    & \quad + 3 \underbrace{\left\| \frac{1}{C}\sum_{c=1}^C \sum_{j=1}^M \lambda_t^j \big(g_t^{j, c}(\theta_t^{w^{*,j,c}_t}) - \Delta_t^{j,c}(\theta_t^c, w^{*,j,c}_t) \big) \right\|_2^2}_{\triangleq \mathcal{T}_{1,1,2}} \\
    & \quad + 3 \underbrace{\left\| \frac{1}{C}\sum_{c=1}^C \sum_{j=1}^M \lambda_t^j \big( \Delta_t^{j,c}(\theta_t^c, w^{*,j,c}_t) - \nabla_\theta J^{c,j}(\theta^c_t) \big) \right\|_2^2}_{\triangleq \mathcal{T}_{1,1,3}},
\end{aligned}
\end{equation}
where $w^{*,j,c}_t$ is the true critic parameter for objective $j$ on client $c$, and $\Delta_t^{j,c}$ is the expected policy gradient under the true critic, defined as:
\[
\Delta_t^{j,c}(\theta_t^c, w^{*,j,c}_t) \triangleq \mathbb{E}_{\nu^{\theta_t^c}(s,a)} \Big[ \mathbb{E}_{s'\sim P(\cdot|s,a)} \big[ \nabla_\theta \log \pi_{\theta_t^c}(a|s) \delta^{j, c}_{t}(w^{*,j,c}_t, s, a, s') \big] \Big],
\]
with $\delta^{j, c}_{t}(w^{*,j,c}_t, s, a, s') \triangleq r_{t}^{j,c}(s,a) + \gamma \boldsymbol{\phi}(s')^\top w^{*,j,c}_t - \boldsymbol{\phi}(s)^\top w^{*,j,c}_t$, and $\nu^{\theta_t^c}$ is the state-action visitation distribution induced by policy $\pi_{\theta_t^c}$.

\vspace{1em}
We now analyze the conditional expectation of \eqref{eq:grad_err_decomp_1} with respect to the filtration $\mathcal{F}_t$, which contains all information up to time $t$. The key insight for bounding the term $\mathcal{T}_{1,1,2}$ is that the stochastic gradients are unbiased estimators of $\Delta_t^{j,c}$ and are independent across clients.
\begin{equation}
\label{eq:grad_err_cond_exp}
\begin{aligned}
    \mathbb{E}[ \mathcal{T}_{1,1}^{\text{gradient-error}} \mid \mathcal{F}_t] & \leq 3\,\mathbb{E}[\mathcal{T}_{1,1,1} \mid \mathcal{F}_t] + 3\,\mathbb{E}[\mathcal{T}_{1,1,2} \mid \mathcal{F}_t] + 3\,\mathbb{E}[\mathcal{T}_{1,1,3} \mid \mathcal{F}_t] \\
    & \leq \mathbb{E}[\mathcal{T}_{1,1,1} \mid \mathcal{F}_t] + \frac{3}{C^2} \sum_{c=1}^C \mathbb{E}\left[\left\| \sum_{j=1}^M \lambda_t^j \big(g_t^{j, c}(\theta_t^{w^{*,j,c}_t}) - \Delta_t^{j,c}(\cdot)\big) \right\|_2^2 \mid \mathcal{F}_t\right] + \mathbb{E}[\mathcal{T}_{1,1,3} \mid \mathcal{F}_t].
\end{aligned}
\end{equation}
The second inequality follows because, for client-independent random variables $\{Z_c\}_{c=1}^C$ with $\mathbb{E}[Z_c \mid \mathcal{F}_t] = 0$, the variance of their average is $\mathbb{E}[\|\frac{1}{C}\sum_c Z_c\|_2^2 \mid \mathcal{F}_t] = \frac{1}{C^2}\sum_c \mathbb{E}[\|Z_c\|_2^2 \mid \mathcal{F}_t]$. Here, $Z_c \triangleq \sum_{j=1}^M \lambda_t^j (g_t^{j, c}(\theta_t^{w^{*,j,c}_t}) - \Delta_t^{j,c}(\cdot))$. The factor of $1/C^2$ combined with the sum over $C$ clients results in a $1/C$ scaling, which is the source of the linear speed-up.

To bound the other terms in \eqref{eq:grad_err_cond_exp}, we rely on the following lemmas.

\begin{lemma} \label{lemma:T113_bound}
(Approximation Error) For all $t \in [T]$, the approximation error is bounded by:
\[ \mathbb{E}[\mathcal{T}_{1,1,3} \mid \mathcal{F}_t] \leq 12 \zeta_{\text{approx}}. \]
\begin{proof}
See Appendix D in \cite{zhou2024finite}.
\end{proof}
\end{lemma}

\begin{lemma} \label{lemma:T111_bound}
(Critic Estimation Error) For all $t \in [T]$, the error from critic estimation is bounded by:
\[ \mathbb{E}[\mathcal{T}_{1,1,1} \mid \mathcal{F}_t] \leq \frac{12}{C} \sum_{c=1}^C \max_{j \in [M]} \mathbb{E}[\| w_t^{*,j,c} - w_t^{j,c} \|_2^2 \mid \mathcal{F}_t]. \]
\begin{proof}
See Appendix D in \cite{zhou2024finite}.
\end{proof}
\end{lemma}

\begin{lemma} \label{lemma:T112_bound}
(Gradient Variance) For all $t \in [T]$, the variance of the gradient estimators is bounded by:
\[\mathbb{E}[\mathcal{T}_{1,1,2} \mid \mathcal{F}_t] \leq  \frac{3}{C^2} \sum_{c=1}^C \mathbb{E}\left[\left\| \sum_{j=1}^M \lambda_t^j \big(g_t^{j, c}(\theta_t^{w^{*,j,c}_t}) - \Delta_t^{j,c}(\cdot)\big) \right\|_2^2 \mid \mathcal{F}_t\right] \leq \frac{12}{C} \frac{(r_{\max} + 2R_{w})^2 (1 - \rho + 4\kappa\rho)}{(1 - \rho)B}. \]
\begin{proof}
First inequality is same as before, and here we only need to prove the second inequality. From Appendix D in \cite{zhou2024finite}, we have the per-client variance bound:
\[ \mathbb{E}\left[\left\| \sum_{j=1}^M \lambda_t^j \big(g_t^{j, c}(\theta_t^{w^{*,j,c}_t}) - \Delta_t^{j,c}(\cdot)\big) \right\|_2^2 \mid \mathcal{F}_t\right] \leq 4 \frac{(r_{\max} + 2R_{w})^2 (1 - \rho + 4\kappa\rho)}{(1 - \rho)B}. \]
Substituting this into the left-hand side of the lemma statement gives:
\begin{align*}
\frac{3}{C^2} \sum_{c=1}^C \left( 4 \frac{(r_{\max} + 2R_{w})^2 (1 - \rho + 4\kappa\rho)}{(1 - \rho)B} \right) &= \frac{3}{C^2} \cdot C \cdot \left( 4 \frac{(\dots)}{(\dots)} \right) \\
&= \frac{12}{C} \frac{(r_{\max} + 2R_{w})^2 (1 - \rho + 4\kappa\rho)}{(1 - \rho)B}.
\end{align*}
\end{proof}
\end{lemma}

\vspace{1em}
\noindent By substituting the results from Lemmas~\ref{lemma:T113_bound}, \ref{lemma:T111_bound}, and \ref{lemma:T112_bound} into \eqref{eq:grad_err_cond_exp}, we obtain the final bound on the gradient estimation error:
\begin{equation}
\label{eq:grad_err_final_bound}
\begin{aligned}
    \mathbb{E}[ \mathcal{T}_{1,1}^{\text{gradient-error}} \mid \mathcal{F}_t] \leq 12 \zeta_{\text{approx}} &+ \frac{12}{C} \frac{(r_{\max} + 2R_{w})^2 (1 - \rho + 4\kappa\rho)}{(1 - \rho)B } \\
    &+ \frac{12}{C} \sum_{c=1}^C \max_{j \in [M]} \mathbb{E}[\| w_t^{*,j,c}- w_t^{j,c} \|_2^2 \mid \mathcal{F}_t].
\end{aligned}
\end{equation}

\subsubsection{Bounding the Multi-Objective Disagreement Drif ($\mathcal{T}_{1,1}^{\text{disagr-drift}}$)}
\label{App:SubSection_MOODriftMitigation_Novel_1}

This term captures the error arising from the discrepancy between locally computed MGDA weights ($\lambda_t^c$) across clients. In conventional FL, drift is often bounded by assuming that with a sufficiently large local batch size ($B$), the variance of client gradients diminishes. However, it is not immediately obvious that the solutions to the local MGDA optimization problems, $\lambda_t^c$, will also converge as $B$ increases. A key contribution of our analysis is to formally establish this property, demonstrating that the regularized MGDA formulation ensures $E[\|\lambda_t^{c_1} - \lambda_t^{c_2}\|_2^2 \mid \mathcal{F}_t ] = O(1/B)$.

We begin with the definition of the term, letting $\bar{\lambda}_t^j \triangleq \frac{1}{C}\sum_{c'=1}^C \lambda_t^{j,c'}$ denote the average MGDA weight for objective $j$.
\begin{align}
    \mathbb{E}[\mathcal{T}_{1,1}^{\text{disagr-drift}} \mid \mathcal{F}_t] &= \mathbb{E}\left[\left\| \frac{1}{C}\sum_{c=1}^C \sum_{j=1}^M g_t^{j, c}(\theta_t^c) \left( \lambda_t^{j,c} - \bar{\lambda}_t^j \right) \right\|_2^2 \mid \mathcal{F}_t\right] \nonumber \\
    &\leq \mathbb{E}\left[\left( \frac{1}{C}\sum_{c=1}^C \left\| \sum_{j=1}^M g_t^{j, c}(\theta_t^c) \left( \lambda_t^{j,c} - \bar{\lambda}_t^j \right) \right\|_2 \right)^2 \mid \mathcal{F}_t\right] \tag{Triangle Inequality} \\
    &\leq \frac{1}{C} \sum_{c=1}^C \mathbb{E}\left[\left\| \sum_{j=1}^M g_t^{j, c}(\theta_t^c) \left( \lambda_t^{j,c} - \bar{\lambda}_t^j \right) \right\|_2^2 \mid \mathcal{F}_t\right] \tag{Jensen's Inequality} \\
    &\leq \frac{1}{C} \sum_{c=1}^C \mathbb{E}\left[\left( \sum_{j=1}^M \|g_t^{j, c}(\theta_t^c)\|_2 \left| \lambda_t^{j,c} - \bar{\lambda}_t^j \right| \right)^2 \mid \mathcal{F}_t\right] \tag{Triangle Inequality}\\
    \label{eq:mgda_err_decomp_1} 
\end{align}
To proceed, we first establish that the local gradients are uniformly bounded.

\begin{lemma}[Bounded Gradient]
\label{lemma:bounded_gradient}
Under our assumptions, the local stochastic gradient for any objective $j$ and client $c$ is bounded as $\|g_t^{j,c}\|_2 \le R \triangleq C_{\psi}(r_{\max}+(1+\gamma)R_{w})$.
\begin{proof}
The gradient is $g_t^{j,c} = \frac{1}{B} \sum_{l=1}^{B} \delta_{t,l}^{j,c} \cdot \boldsymbol{\psi}(s_{t,l}, a_{t,l})$, where $\delta_{t,l}^{j,c} = r_{t,l}^{j,c} + \gamma \boldsymbol{\phi}(s_{t,l+1})^\top \mathbf{w}_t^{j,c} - \boldsymbol{\phi}(s_{t,l})^\top \mathbf{w}_t^{j,c}$. Given the assumptions $\|\boldsymbol{\phi}(\cdot)\|_2 \le 1$, $\|\boldsymbol{\psi}(\cdot, \cdot)\|_2 \le C_\psi$, and the critic projection step in Algorithm~\ref{alg:firm} ensuring $\|\mathbf{w}_t^{j,c}\|_2 \le R_w$, we have:
\[
\|g_t^{j,c}\|_2 \le \frac{1}{B} \sum_{l=1}^B \|\delta_{t,l}^{j,c} \boldsymbol{\psi}_{t,l}\|_2 \le \max_l |\delta^{j,c}_{t,l}| \cdot \|\boldsymbol{\psi}_{t,l}\|_2 \le C_\psi \left(|r_{t,l}^{j,c}| + (1+\gamma)\|\boldsymbol{\phi}(\cdot)\|_2 \|\mathbf{w}_t^{j,c}\|_2\right) \le C_{\psi}(r_{\max}+(1+\gamma)R_{w}).
\]
\end{proof}
\end{lemma}

Applying Lemma \ref{lemma:bounded_gradient} to Equation~\eqref{eq:mgda_err_decomp_1}, we get:

\begin{equation}
\label{eq:MGDAErrorIntermediate_1_1}
\begin{aligned}
   & \mathbb{E}[\mathcal{T}_{1,1}^{\text{disagr-drift}} \mid \mathcal{F}_t] &\leq \frac{R^2}{C} \sum_{c=1}^C \mathbb{E}\left[\left(\sum_{j=1}^M \left|\lambda_t^{j,c} - \bar{\lambda}_t^j\right|\right)^2 \mid \mathcal{F}_t\right] \nonumber  
\end{aligned}
\end{equation}

Now,let $S_c \triangleq \sum_{j=1}^M |\lambda_t^{j,c} - \bar{\lambda}_t^j|$. Now, since \( \sum_{j=1}^M|\lambda_t^{j,c}| = 1\) and \( \sum_{j=1}^M|\bar{\lambda}_t^j| = 1 \), by  the triangle inequality, $S_c \leq  \sum_{j=1}^M|\lambda_t^{j,c}| + \sum_{j=1}^M|\bar{\lambda}_t^j| = 1 + 1 = 2$. Moreover, for any non-negative value $S_c \in [0, 2]$, the inequality $S_c^2 \le 2S_c$ holds. Applying this insight, we can bound the term in Equation~\eqref{eq:MGDAErrorIntermediate_1_1} as follows:

\begin{align} \mathbb{E}[\mathcal{T}_{1,1}^{\text{disagr-drift}} \mid \mathcal{F}_t] &\leq \frac{R^2}{C} \sum_{c=1}^C \mathbb{E}\left[\left(\sum_{j=1}^M \left|\lambda_t^{j,c} - \bar{\lambda}_t^j\right|\right)^2 \mid \mathcal{F}_t\right] \nonumber \\ &\leq \frac{2R^2}{C} \sum_{c=1}^C \mathbb{E}\left[\sum_{j=1}^M \left|\lambda_t^{j,c} - \bar{\lambda}_t^j\right| \mid \mathcal{F}_t\right] \tag{Using $S_c^2 \le 2 S_c$ since $S_c \le 1$} \\ &= \frac{2R^2}{C} \sum_{c=1}^C \mathbb{E}\left[ \left\| \lambda_t^c - \bar{\lambda}_t \right\|_1 \mid \mathcal{F}_t \right] \nonumber \\ &\leq \frac{2R^2}{C^2} \sum_{c=1}^C \sum_{c'=1}^C \mathbb{E}\left[ \left\| \lambda_t^c - \lambda_t^{c'} \right\|_1 \mid \mathcal{F}_t \right], \label{eq:mgda_err_decomp_2_alt_0}\\
&   \frac{2R^2 \sqrt{M}}{C^2} \sum_{c=1}^C \sum_{c'=1}^C \mathbb{E}\left[ \left\| \lambda_t^c - \lambda_t^{c'} \right\|_2 \mid \mathcal{F}_t \right]  \tag{Using $ \| v\|_1 \leq \sqrt{M} \| v \|_2$  for  all \(v \in \mathbb{R}^M\)}
\\ \label{eq:mgda_err_decomp_2_alt} \end{align}

Here, we define 
\(\bar{\lambda}_t \triangleq \tfrac{1}{C}\sum_{c=1}^C \bar{\lambda}_t^c \in \mathbb{R}^M\) 
as the average preference vector across clients, with 
\(\bar{\lambda}_t^j\) denoting its $j$-th component.  
The inequality~\eqref{eq:mgda_err_decomp_2_alt_0} follows directly from the triangle inequality for the 
$\ell_1$ norm:
\[
\|\lambda_t^c - \bar{\lambda}_t\|_1 
= \left\|\lambda_t^c - \frac{1}{C}\sum_{c'=1}^C \lambda_t^{c'}\right\|_1
= \left\|\frac{1}{C}\sum_{c'=1}^C (\lambda_t^c - \lambda_t^{c'})\right\|_1 
\le \frac{1}{C}\sum_{c'=1}^C \|\lambda_t^c - \lambda_t^{c'}\|_1.
\]

The bound now depends on the pairwise difference between local MGDA weights. We decompose this difference recursively using the update rule $\lambda_t^c = (1-\eta_t) \lambda_{t-1}^c + \eta_t \lambda_t^{*,c}$:
\begin{align}
    \|\lambda_t^c - \lambda_t^{c'}\|_2 &= \|(1-\eta_t)(\lambda_{t-1}^c - \lambda_{t-1}^{c'}) + \eta_t(\lambda_t^{*,c} - \lambda_t^{*,c'})\|_2 \nonumber \\
    &\le (1-\eta_t)\|\lambda_{t-1}^c - \lambda_{t-1}^{c'}\|_2 + \eta_t\|\lambda_t^{*,c} - \lambda_t^{*,c'}\|_2.
\end{align}
Unrolling this recursion from $t$ down to $1$ and noting that $\lambda_0^c = \lambda_0^{c'}$ for all $c,c'$, we obtain:
\begin{equation}
\label{eq:lambda_recursion_unrolled}
    \mathbb{E}[\|\lambda_t^c - \lambda_t^{c'}\|_2 \mid \mathcal{F}_t] \leq \sum_{i=1}^t \left[ \eta_i \prod_{k=i+1}^t (1-\eta_k) \right] \mathbb{E}[\|\lambda_i^{*,c} - \lambda_i^{*,c'}\|_2 \mid \mathcal{F}_t].
\end{equation}
The following crucial lemma connects the difference in optimal MGDA weights to the difference in client gradients.

\begin{lemma}
\label{lemma:lambda_grad_relation}
For any two clients $c, c' \in \mathcal{C}$ at time $t$, the difference in their optimal MGDA weights is bounded by the difference in their local gradients:
\[
    \|\lambda_t^{*, c} - \lambda_t^{*, c'}\|_2 \leq \frac{4 R M}{\beta} \max_{j\in[M]} \|g_t^{j,c}(\theta_t^c) - g_t^{j,c'}(\theta_t^{c'}) \|_2,
\]
where $R$ is the gradient bound from Lemma \ref{lemma:bounded_gradient}.
\begin{proof}
See Appendix~\ref{Appendix: Proof_of_relating_grad_to_MGDACoeff_1}.
\end{proof}
\end{lemma}

Substituting \eqref{eq:lambda_recursion_unrolled} and Lemma \ref{lemma:lambda_grad_relation} into \eqref{eq:mgda_err_decomp_2_alt}, we arrive at:
\begin{equation}
\label{eq:mgda_err_decomp_5}
\mathbb{E}[\mathcal{T}_{1,1}^{\text{disagr-drift}} \mid \mathcal{F}_t] \leq \frac{8 R^3 \sqrt{M^3}}{ \beta C^2} \sum_{c,c'} \sum_{i=1}^{t} \left[ \eta_i \prod_{k=i+1}^{t} (1 - \eta_k) \right] \max_{j} \mathbb{E} \left[ \|g_i^{j,c} - g_i^{j,c'} \|_2 \mid \mathcal{F}_t \right].
\end{equation}
 This bound demonstrates that the Multi-Objective Disagreement Drift is controlled by the history of client drift, i.e., the difference in gradients across clients. We now bound this drift.

\begin{lemma}[Client Gradient Drift]
\label{lemma:client_drift_bound}
Let $t_0 \le t$ be the last synchronization time-step. For any two clients $c \neq c'$, the expected gradient difference is bounded by:
\begin{align*}
    \mathbb{E}\Big[\| g_t^{j,c} - g_t^{j,c'} \|_2 \mid \mathcal{F}_t\Big] &\leq 4 \sqrt{\zeta_{\text{approx}}} + 2 \sqrt{\mathbb{E}[ \|w_t^{*,j,c} - w_t^{j,c}\|_2 \mid \mathcal{F}_t]} + 2 \sqrt{\mathbb{E}[ \|w_t^{*,j,c'} - w_t^{j,c'}\|_2 \mid \mathcal{F}_t]} \\
    & \quad + 4\sqrt{ \frac{(r_{\max} + R_{w})^2 (1 - \rho + 4\kappa\rho)}{(1 - \rho)B}} + 4 L_J R \alpha (t-t_0) + 2 \zeta.
\end{align*}
\begin{proof}
The proof can be found in Appendix~\ref{App:ProofOfLemmaClientDriftGradBdd_1}.
\end{proof}
\end{lemma}

By substituting the bound from Lemma \ref{lemma:client_drift_bound} into \eqref{eq:mgda_err_decomp_5}, we obtain the final bound for $\mathcal{T}_{1,1}^{\text{disagr-drift}}$. To simplify the expression, we bound the average over client pairs by the maximum over all pairs, which allows us to remove the $\frac{1}{C^2}\sum_{c,c'}$ term. This yields the following comprehensive bound: 

\begin{equation} \label{eq:final_mgda_error_bound}
\begin{aligned} \mathbb{E}[\mathcal{T}_{1,1}^{\text{disagr-drift}} \mid \mathcal{F}_t] &\leq \frac{8 R^3 M^{3/2}}{\beta} \sum_{i=1}^{t} \left( \left[ \eta_i \prod_{k=i+1}^{t} (1 - \eta_k) \right] \times \right. \\ & \quad \max_{(j,c,c') \in [M]\times \mathcal{C}\times \mathcal{C}} \bigg( 4 \sqrt{\zeta_{\text{approx}}} + 2 \sqrt{\mathbb{E}[ \|w_i^{*,j,c} - w_i^{j,c}\|_2 \mid \mathcal{F}_t]} + 2 \sqrt{\mathbb{E}[ \|w_i^{*,j,c'} - w_i^{j,c'}\|_2 \mid \mathcal{F}_t]} \\ & \qquad \left. + 4\sqrt{ \frac{(r_{\max} + R_{w})^2 (1 - \rho + 4\kappa\rho)}{(1 - \rho)B}} + 4 L_J R \alpha K  + 2 \zeta \bigg) \right) \triangleq \mathcal{E}_{\text{MGDA}}(t), \end{aligned} \end{equation}

where remember that \(K\) is the maximum number of local updates. This expression, which we can denote as $\mathcal{E}_{\text{MGDA}}(t)$, provides the final upper bound on the MGDA discrepancy error. It is composed of terms related to function approximation error ($\zeta_{\text{approx}}$), local critic estimation error ($\|w-w^*\|$), gradient variance (inversely proportional to batch size $B$), and divergence due to local updates (proportional to $i-i_0$).

\subsubsection{Final step in bounding \(\mathcal{T}_{1,1}\)}
Now, using Equations(\ref{eq:NeuriPS_FMOORL_Eq_Decomposition_Lemma_1}, \ref{eq:grad_err_final_bound}, and \ref{eq:final_mgda_error_bound}) we have:

   \begin{equation}
\label{eq:NeuriPS_FMOORL_Eq_Decomposition_Lemma_2}
\begin{aligned}
&E[\mathcal{T}_{1,1}\mid \mathcal{F}_t]\leq E[\mathcal{T}_{1,1}^{\text{gradient-error}} \mid \mathcal{F}_t] + E[\mathcal{T}_{1,1}^{\text{disagr-drift}} \mid \mathcal{F}_t]\\
&12 \zeta_{\text{approx}}\; +\; \frac{12}{C} \frac{(r_{\max} + 2R_{w})^2 (1 - \rho + 4\kappa\rho)}{(1 - \rho)B } \;+\; \frac{12}{C} \sum_{c=1}^C \max_{j \in [M]} \mathbb{E}[\| w_t^{*,j,c}- w_t^{j,c} \|_2^2 \mid \mathcal{F}_t] \;+\; \mathcal{E}_{\text{MGDA}}(t)
\end{aligned}
\end{equation}

\subsubsection{Boudning the term \(\mathcal{T}_{1,2}\)}
To bound this term, we can use the proof of Lemma~\ref{lemma:client_drift_bound}. First of all, by applying Proposition~1 from \cite{xu2020improving} we have:

\begin{equation}
    \label{eq:NeuriPS_FMOORL_Eq_1Decomposition_T12_1}
\begin{aligned}
& \mathcal{T}_{1,2} = \| \frac{1}{C}\sum_{c=1}^C \big(\nabla_\theta J(\theta^c_t)   - \nabla_\theta J(\bar{\theta}_t)\big) \lambda_t\|_2^2 \leq  \frac{1}{C} \sum_{c=1}^C \left\| \big(\nabla_\theta \mathbf{J}^c(\theta^c_t) - \nabla_\theta \mathbf{J}^c(\bar{\theta}_t)\big)\lambda_t + \big(\nabla_\theta \mathbf{J}^c(\bar{\theta}_t) - \nabla_\theta \mathbf{J}(\bar{\theta}_t)\big) \lambda_t \right\|_2^2  \\
& \qquad  \leq \frac{2}{C} \sum_{c=1}^C \underbrace{\left\| \nabla_\theta \mathbf{J}^c(\theta^c_t) - \nabla_\theta \mathbf{J}^c(\bar{\theta}_t) \right\|_2^2}_{\text{Parameter Drift}} + \frac{2}{C} \sum_{c=1}^C \underbrace{\left\| \nabla_\theta \mathbf{J}^c(\bar{\theta}_t) - \nabla_\theta \mathbf{J}(\bar{\theta}_t) \right\|_2^2}_{\text{ Heterogeneity}} \\
&\qquad \leq  \frac{L^2_J}{C^2} \sum_{c=1}^C \sum_{c'=1}^C \| \theta^c_t   - \theta_t^{c'} \|_2^2  + 2 \zeta^2  \leq \frac{L^2_J}{C^2}\;\left(\alpha \sum_{\tau = 0}^{t-t_0} E\Big[\|    \sum_{j=1}^M\lambda^{j,c}_{t_0+\tau} \; g_{t_0+\tau}^{j,c} \; -  \;\lambda^{j,c'}_{t_0+\tau} \; g_{t_0+\tau}^{j,c'}\|_2 \mid \mathcal{F}_t\Big] \right)^2 + 2 \zeta^2\\
       & \qquad \leq \frac{L^2_J}{C^2}\;\left(2 \alpha \; \sum_{\tau = 0}^{t-t_0} E\Big[\|    \sum_{j=1}^M\lambda^{j,c}_{t_0+\tau} \; g_{t_0+\tau}^{j,c} \|_2
       \; + \| \sum_{j=1}^M \lambda^{j,c'}_{t_0+\tau} \; g_{t_0+\tau}^{j,c'}\|_2 \mid \mathcal{F}_t\Big]\right)^2+ 2 \zeta^2 \leq \boxed{4 \, L_J^2\, R^2\, \alpha^2\;K^2 + 2 \zeta^2},
\end{aligned}
\end{equation}
where similar to the Lemma~\ref{lemma:client_drift_bound}, \(t_0\) is  the most-recent time before \(t\) that server-aggregation happens, i.e. \(\theta_{t_0}^c = \theta_{t_0}\), for all \(c\in\mathcal{C}\), \(R\) is the upper bound for the norm of the batch-gradients, i.e., \(\| g_{t}^{j,c}\|_2 \leq R\) for all \((t,c) \in [T]\times \mathcal{C}\), and \(|t-t_0|\leq K\), i.e., \(K\) is the maximum number of local steps before a server aggregation happens.

\subsection{Final step in bounding \(\mathcal{T}_1\)}
Now, using Equation\eqref{eq:NeuriPS_FMOORL_Eq_1Decomposition_1}, and the bounds obtained for \(\mathcal{T}_{1,1}\) and \(\mathcal{T}_{1,2}\) in Equations~\eqref{eq:NeuriPS_FMOORL_Eq_Decomposition_Lemma_2} and \eqref{eq:NeuriPS_FMOORL_Eq_1Decomposition_T12_1} we will have:

   \begin{equation}
\label{eq:NeuriPS_FMOORL_Eq_Decomposition_Lemma_2_2}
\begin{aligned}
E[\mathcal{T}_{1}\mid \mathcal{F}_t]&\leq E[\mathcal{T}_{1,1}\mid \mathcal{F}_t] + E[\mathcal{T}_{1,2}\mid \mathcal{F}_t] \leq   \mathcal{Q}_t(B,\alpha, \eta, \zeta_{approx}, M)\\
\end{aligned}
\end{equation}

where,

\begin{equation*}
    \begin{aligned}
        \mathcal{Q}_t(B,\beta, \alpha, \eta, \zeta_{approx}, M) \;&\triangleq \; 12 \zeta_{\text{approx}} +  \frac{12}{C} \frac{(r_{\text{max}} + 2R_{\text{w}})^2 (1 - \rho + 4\kappa\rho)}{(1 - \rho)B } + \\
        &\frac{12}{C} \sum_{c=1}^C \max_{j \in [M]} E[\| w_t^{*,c}- w_t^c  \|_2^2 \mid \mathcal{F}_t] +\mathcal{E}_{\text{MGDA}}(t)+ 4 \, L_J\, R^2\, \alpha^2 K^2 + 2 \zeta^2
    \end{aligned}
\end{equation*}

\subsection{Final Convergence Analysis}

We now consolidate the preceding bounds to derive the main convergence result.

\subsubsection{Rearranging the Descent Lemma}
Thus, using Equation\eqref{eq:NeuriPS_FMOORL_Eq_Decomposition_Lemma_2_2}, now by letting \(\alpha_t = \alpha \) and \(\alpha \in(0, \frac{1}{L_J}]\), we can continue Equation\eqref{eq:NeuriPS_FMOORL_Eq_1Decomposition_0} as follows:

\begin{align}
&\mathbb{E}\bigg[\boldsymbol{\lambda}_t^\top \mathbf{J}(\bar{\theta}_{t+1}) \mid \mathcal{F}_t\bigg] \geq \mathbb{E}\bigg[\boldsymbol{\lambda}_t^\top \mathbf{J}(\bar{\theta}_t) + \frac{\alpha}{2} \|\nabla_\theta \mathbf{J}(\bar{\theta}_t)\boldsymbol{\lambda}_t\|_2^2 - \frac{\alpha}{2} \mathcal{T}_1 \mid \mathcal{F}_t\bigg].\\
& \to \mathbb{E}\left[\|\nabla_\theta \mathbf{J}(\bar{\theta}_t)\boldsymbol{\lambda}_t\|_2^2 \mid \mathcal{F}_t\right] \leq \frac{2}{\alpha} \mathbb{E}\left[\boldsymbol{\lambda}_t^\top \left(\mathbf{J}(\bar{\theta}_{t+1}) - \mathbf{J}(\bar{\theta}_t)\right) \mid \mathcal{F}_t\right] + \mathbb{E}[\mathcal{Q}_t \mid \mathcal{F}_t]. \label{eq:descent_lemma_start}
\end{align}

Now, by   taking expectation of \(\mathcal{F}_t\) on both side of Equation~\eqref{eq:descent_lemma_start} we will have:

\begin{align}
 \mathbb{E}\left[\|\nabla_\theta \mathbf{J}(\bar{\theta}_t)\boldsymbol{\lambda}_t\|_2^2\right] \leq \frac{2}{\alpha} \mathbb{E}\left[\boldsymbol{\lambda}_t^\top \left(\mathbf{J}(\bar{\theta}_{t+1}) - \mathbf{J}(\bar{\theta}_t)\right) \right] + \mathbb{E}[\mathcal{Q}_t ]. \label{eq:descent_lemma_start_1}
\end{align}
\subsubsection{Bounding the Telescopic Sum}
The first term on the right-hand side of Equation~\eqref{eq:descent_lemma_start_1} forms a telescopic sum when averaged over time:
\begin{equation}
\label{eq:NeuriPS_FMOORL_Eq_1Decomposition_3}
\begin{aligned}
& \frac{1}{T} \sum_{t=1}^T E\bigg[\| \nabla_\theta \mathbf{J}(\bar{\theta}_t) \|_2^2\bigg]  \leq \frac{2}{\alpha } \; \frac{1}{T}\; \bigg( \sum_{t=1}^T \lambda_t^\top \big(\mathbf{J}(\bar{\theta}_{t+1}) - \mathbf{J}(\bar{\theta}_t) \big) \bigg) + \; \frac{1}{T}\; \sum_{t=1}^T E\bigg[\mathcal{Q}_t(B, \beta, \alpha, \eta, \zeta_{approx}, M)\bigg]
\end{aligned}
\end{equation}

We can bound the first term on RHS of Equation~\eqref{eq:NeuriPS_FMOORL_Eq_1Decomposition_3} bound this sum using the following lemma:
\begin{lemma}[Telescopic Sum Bound]
\label{lemma:telescopic_sum}
Let the learning rate for the MGDA weights be $\eta_t$. The telescopic sum of objective values is bounded as:
\[
\sum_{t=1}^{T} \mathbb{E}\left[\boldsymbol{\lambda}_t^\top (\mathbf{J}(\bar{\theta}_{t+1}) - \mathbf{J}(\bar{\theta}_t))\right] \leq \frac{r_{\max}}{1 - \gamma} \left(1 + \sum_{t=1}^{T-1} 2\eta_t\right).
\]
\begin{proof}[Proof]
Applying summation by parts, the sum can be rewritten and bounded using Hölder's inequality:
\begin{align*}
    \sum_{t=1}^{T} \mathbb{E}\left[\boldsymbol{\lambda}_t^\top \Delta\mathbf{J}_{t+1}\right] &= \mathbb{E}\left[\sum_{t=1}^{T-1}(\boldsymbol{\lambda}_t - \boldsymbol{\lambda}_{t+1})^\top \mathbf{J}_{t+1} - \boldsymbol{\lambda}_1^\top \mathbf{J}_1 + \boldsymbol{\lambda}_T^\top \mathbf{J}_{T+1}\right] \\
    &\leq \mathbb{E}\left[\sum_{t=1}^{T-1}\|\boldsymbol{\lambda}_t - \boldsymbol{\lambda}_{t+1}\|_1 \|\mathbf{J}_{t+1}\|_\infty + \|\boldsymbol{\lambda}_T\|_1 \|\mathbf{J}_{T+1}\|_\infty\right].
\end{align*}
The objective value is bounded by $\|\mathbf{J}(\cdot)\|_\infty \le \frac{r_{\max}}{1-\gamma}$. The difference $\|\boldsymbol{\lambda}_t - \boldsymbol{\lambda}_{t+1}\|_1$ is bounded by $2\eta_t$ due to the smoothing update and the fact that all $\boldsymbol{\lambda}$ vectors lie on the probability simplex. Combining these yields the result.
\end{proof}
\end{lemma}

\subsubsection{Final Convergence Rate}
By applying Lemma~\ref{lemma:telescopic_sum} on Equation~\eqref{eq:NeuriPS_FMOORL_Eq_1Decomposition_3} we will have:
\begin{align}
\frac{1}{T}\sum_{t=1}^T \mathbb{E}\left[\|\nabla_\theta \mathbf{J}(\bar{\theta}_t)\boldsymbol{\lambda}_t\|_2^2 \right] \leq \frac{2}{T \alpha } \frac{r_{\max}}{1 - \gamma} \left(1 + \sum_{t=1}^{T-1} 2\eta_t\right) +  \frac{1}{T}\sum_{t=1}^T\mathbb{E}[\mathcal{Q}_t]. \label{eq:refBeforeFinishProof_1}
\end{align}
Now let \( E\Big[ \|w_t^{*,j,c} - w_t^{j,c}\|_2 \mid \mathcal{F}_t\Big]\leq \varepsilon_{\text{critic}}\), and \(\eta_t \in (0,1)\), we will have:

\begin{equation}
\label{eq:ICLR_BoundingFinalTerms_1}
    \begin{aligned}
\mathcal{Q}_t(B, \beta, \alpha, \eta, \zeta_{approx}, M) \; &\leq
\;
\frac{8 R^{3} \sqrt{M^3}}{\beta}\;
\Bigl[
      4\sqrt{\zeta_{\mathrm{approx}}}
      +4\sqrt{\varepsilon_{\text{critic}}}
      +4\sqrt{\tfrac{(r_{\max}+R_w)^2(1-\rho+4\kappa\rho)}{(1-\rho)B}}
      +4 L_J R\,\alpha\,K + 2 \zeta
\Bigr]\\
&  +\frac{12}{C} \Bigl[ \frac{(r_{\text{max}} + 2R_{\text{w}})^2 (1 - \rho + 4\kappa\rho)}{(1 - \rho)B }   \Bigl]\; + \; 12\; \varepsilon_{\text{critic}} + 12 \zeta_{\text{approx}} +  4 \, L_J^2\, R^2\, \alpha^2 K^2 + \zeta^2.
    \end{aligned}
\end{equation}

Moreover, if we choose \(\eta_t  = \frac{1}{t}\) we will have

\begin{equation}
\label{eq:NeurIPS_FMOORL_StrongConvexity_Purturn_Object_2}
    \begin{aligned}
        &\frac{2 r_{\max}}{T(1 - \gamma ) \alpha} \left( 1 + \sum_{t=1}^{T} 2\eta_t \right) \leq \frac{2\; \; r_{\max}}{T(1 -\gamma)} \left( 1 + \log(T) \right)
    \end{aligned}
\end{equation}

Substituting the upper bounds from Equation~\eqref{eq:NeurIPS_FMOORL_StrongConvexity_Purturn_Object_2} and Equation~\eqref{eq:ICLR_BoundingFinalTerms_1} into Equation~\eqref{eq:refBeforeFinishProof_1}, we arrive at the final convergence rate:
\begin{equation}
\label{eq:final_rate}
\boxed{
\begin{aligned}
    \frac{1}{T} \sum_{t=1}^T \mathbb{E}\left[\|\nabla_\theta \mathbf{J}(\bar{\theta}_t)\boldsymbol{\lambda}_t\|_2^2 \right] = \mathcal{O}\Bigg(
    &\underbrace{\frac{\log T}{T}}_{\text{Optimization Error}}
    + \underbrace{\frac{1}{C B}}_{\text{Variance}}
    + \underbrace{\sqrt{\zeta_{\text{approx}}} + \sqrt{\varepsilon_{\text{critic}}} + 2 \zeta^2}_{\text{Bias}}
    + \underbrace{\alpha^2 K^2 + \frac{\sqrt{M^3}}{\beta \; \sqrt{B}} \alpha K}_{\text{Client Drift}}
    \Bigg).
\end{aligned}}
\end{equation}
This completes the proof. \qed

\section{Utility Lemmas}
\label{App:Utility_Lemmas_in_Main Proof}
\subsection{Proof of Lemma \ref{lemma:lambda_grad_relation}}
\label{Appendix: Proof_of_relating_grad_to_MGDACoeff_1}
\paragraph{Proof:}
We first note that we have \(\Biggl\|\sum_{j=1}^M \lambda^j\,g_t^{j,c}(\theta_t^c)\Biggr\|_2^2 = \lambda^T G_t^c \lambda\), where \(G_t^c\) is a PSD matrix, i.e., \(G_t^c \triangleq (A_t^c)^\top  \, (A_t^c)\)  and \(A_t^c\) is the matrix obtained by stacking gradienst of the \(M\) objectives, defined as follows:
\begin{equation}
    \begin{aligned}
        & A_t^c\; \triangleq\;\bigl[\,g_t^1(\theta_t^c)\;\bigm|\;g_t^2(\theta_t^c)\;\bigm|\;\cdots\;\bigm|\;g_t^M(\theta_t^c)\bigr]
\in \mathbb{R}^{d\times M}
    \end{aligned}
\end{equation}
Thus, each client \(c\), at time \(t\), is equivalently is solving this optimization problem to obtain their \(\lambda_t^{*,c}\):

\begin{equation}
\label{eq:NeurIPS_FMOORL_RewriteMGDA_Opt_1}
    \lambda_t^{*, c} \triangleq \text{argmin}_{\lambda \in \mathbb{R}^M}
\lambda^T G_{t}^c  \lambda + \frac{\beta}{2} \| \lambda\|_2^2
\quad\text{s.t.}\quad
\lambda \ge 0,\;
\|\lambda\|_1 = 1.
\end{equation}
Now, define function \(f_G: \Delta_d \to  \mathbb{R}^+\) as follows: \(f_G(\lambda) = \lambda^T G \lambda + \beta \|\lambda \|_2^2 \). Now, Since \(\nabla_\lambda^2 f_G(\lambda) \succeq  \beta \), thus \(f_G\) is \(\beta\)-strongly convex, and we will have the following:

\begin{equation}
\label{eq:NeurIPS_FMOORL_StrongConvexity_Purturn_Object_1}
    \begin{aligned}
        &\forall \lambda \in \Delta_d:  f_G(\lambda) \geq f_G(\lambda^{*}) + \frac{\beta}{2} \| \lambda^{*} - \lambda\|_2^2,
    \end{aligned}
\end{equation}
where \(\lambda^*\) is the unique minimizer of the function \(f_G(\lambda)\), i.e., \(\lambda^* = argmin_{\lambda \in \Delta_d} f_G(\lambda)\). Moreover, let \(h(\lambda) = f_G(\lambda) - f_{G'}(\lambda) = \lambda^T (G-G') \lambda\), where \(G'\) is a PSD matrix as well. Now, since \(h(\lambda)\) is continuously differentiable and  \(\|\nabla h(\lambda)\|_2 \leq 2 \|G - G'\|_{op}\), where \(\| \|_{op}\) is the   operator norm (spectral norm).  Thus, we can apply Mean-Value 
theorem to get the following inequlaity for all \(\lambda_1, \lambda_2 \in \Delta_d\):

$$
h(\lambda_1)-h(\lambda_2)
=\int_{0}^{1}\nabla h\bigl(\lambda_2 + t(\lambda_1-\lambda_2)\bigr)^\top(\lambda_1-\lambda_2)\,dt.
$$
Now by applying Cauchy–Schwarz inside the integral gives

\begin{equation}
\label{eq:NeurIPS_FMOORL_inequalityModulos_1}
    \begin{aligned}
        & \bigl|h(\lambda_1)-h(\lambda_2)\bigr|
\;\le\;\int_{0}^{1}\bigl\|\nabla h(\lambda_2 + t(\lambda_1-\lambda_2))\bigr\|_2\,\|\lambda_1-\lambda_2\|_2\,dt
\;\le\;\Bigl(\sup_{\lambda\in\Delta_d}\|\nabla h(\lambda)\|_2\Bigr)\|\lambda_1-\lambda_2\|_2.
    \end{aligned}
\end{equation}

Now, using  \(\|\nabla h(\lambda)\|_2 \leq 2 \|G - G'\|_{op}\) and Equation(\ref{eq:NeurIPS_FMOORL_inequalityModulos_1}) we will have:

\begin{equation}
\label{eq:NeurIPS_FMOORL_inequalityModulos_2}
    \begin{aligned}
        & \bigl|h(\lambda_1)-h(\lambda_2)\bigr|
\;\le\; 2 \|G - G'\|_{op} \,\, \|\lambda_1-\lambda_2\|_2.
    \end{aligned}
\end{equation}

Now, let \( \lambda'^{*}\) be the minizer of function \(f_{G'}(\lambda)\) and recall that \(\lambda^*\) is the minimizer of \(f_G(\lambda)\). Now, consdierng Equation(\ref{eq:NeurIPS_FMOORL_StrongConvexity_Purturn_Object_1}) and Equation(\ref{eq:NeurIPS_FMOORL_inequalityModulos_2}), we can use Proposition 4.32 from \cite{bonnans2013perturbation} to get the following:

\begin{equation}
\label{eq:NeurIPS_FMOORL_Boundinglambda_MatrixGrads_1}
    \begin{aligned}
        & \|\lambda^* - \lambda'^{*}\|_2 \le \frac{4}{\beta} \,\, \| G  - G'\|_{op}
    \end{aligned}
\end{equation}

Now, immediately, using the fact that \(\lambda^{*,c}_t, \lambda^{*,c'}_t\) are the minimzer of \(f_{G_t^c}(\lambda)\) and \(f_{G_t^{c'}}(\lambda)\) respectively, we can use Equation (\ref{eq:NeurIPS_FMOORL_Boundinglambda_MatrixGrads_1}) to get the following:
\begin{equation}
\label{eq:NeurIPS_FMOORL_Boundinglambda_MatrixGrads_2}
    \begin{aligned}
        & \|\lambda^{*,c}_t - \lambda^{*,c'}_t\|_2 \le \frac{4}{\beta} \,\, \| G_t^c  - G_t^{c'}\|_{op}
    \end{aligned}
\end{equation}

Now, give the fact that \(G_t^c = (A_t^c)^\top A_t^c\) and \(G_t^{c'} = (A_t^{c'})^\top A_t^{c'}\), we will have:

\begin{equation}
\label{eq:NeurIPS_FMOORL_Boundinglambda_MatrixGrads_3}
 \begin{aligned}
        \| G_t^c  - G_t^{c'}\|_{op}  &  =  \| (A_t^c)^\top (A_t^c-A_t^{c'}) +   (A_t^c-A_t^{c'})^\top A_t^{c'}\|_{op} \leq  \| (A_t^c)^\top (A_t^c-A_t^{c'})\|_{op} + \| (A_t^c-A_t^{c'})^\top A_t^{c'}\|_{op}\\
        &\leq \|A_t^c\|_{op} \,\, \|A_t^c-A_t^{c'}\|_{op} + \| A_t^c-A_t^{c'}\| \,\,  \|A_t^{c'}\|_{op} = (\| A_t^c\|_{op} + \| A_t^{c'}\|_{op}) \, \, \| A_t^c-A_t^{c'}\|_{op}
    \end{aligned}
\end{equation}

Now, we can bound \( \| A_t^c-A_t^{c'}\|_{op}\) as follows:

\begin{equation}
\label{eq:NeurIPS_FMOORL_Boundinglambda_MatrixGrads_4}
 \begin{aligned}
         &  \| A_t^c-A_t^{c'}\|_{op} \leq \| A_t^c-A_t^{c'}\|_{F} = \big(  \sum_{j=1}^M \| g_t^{j,c}(\theta_t^c) - g_t^{j,c'}(\theta_t^{c'}) \|_2^2 \big)^{\frac{1}{2}} \leq \sqrt{M} \max_{j\in[M]} \| g_t^{j,c}(\theta_t^c) - g_t^{j,c'}\|
    \end{aligned}
\end{equation}

Also to bound \(\| A_t^c\|_{op} \) note that if \(\|g_t^{j,c}\|_2, \; \| g_t^{j,c'}\|_2 \leq R\), then we will have:

\begin{equation}
\label{eq:NeurIPS_FMOORL_Boundinglambda_MatrixGrads_5}
 \begin{aligned}
         &  \| A_t^c\|_{op} , \; \| A_t^{c'}\|_{op} \leq \sqrt{M} R.
    \end{aligned}
\end{equation}

Therefore, combining Equations~(\ref{eq:NeurIPS_FMOORL_Boundinglambda_MatrixGrads_2}-\ref{eq:NeurIPS_FMOORL_Boundinglambda_MatrixGrads_5}) yields:

\begin{equation}
\label{eq:NeurIPS_FMOORL_Boundinglambda_MatrixGrads_2_2}
    \begin{aligned}
        & \|\lambda^{*,c}_t - \lambda^{*,c'}_t\|_2 \le \frac{4 MR}{\beta} \,\,  \max_{j\in[M]} \| g_t^{j,c}(\theta_t^c) - g_t^{j,c'}\|,
    \end{aligned}
\end{equation}
and this concludes the proof \(\square\).

\subsection{Proof of Lemma \ref{lemma:client_drift_bound}}
\label{App:ProofOfLemmaClientDriftGradBdd_1}
Using triangulare inequality we can bound \( \| g_t^{j,c} - g_t^{j, c'} \|_2 \) as follows:

\begin{equation}
\label{eq:boundingClientDriftICMLWorkshop_grads_1}
    \begin{aligned}
         \| g_t^{j,c} - g_t^{j, c'} \|_2 &\leq  \|g_t^{j,c} - g_t^{j,c}(w_t^{*,j,c})\|_2 + \| g_t^{j,c}(w^{*,j,c}_t) - \Delta_t^{j,c}(\theta_t^c, w^{*,j,c}_t)\|_2 + \|\Delta_t^{j,c}(\theta^c_t, w^{*,j,c}_t) - \nabla_{\theta}J^j(\theta_t^c) \|_2 \\
        & \qquad \qquad \qquad\qquad \qquad+  \| \nabla_{\theta}J^{j,c}(\theta_t^c) - \nabla_{\theta}J^{j,c'}(\theta_t^{c'})\|_2\\
        & \qquad + \|\nabla_{\theta}J^j(\theta_t^{c'}) - \Delta_t^{j, c'}(\theta^c_t, w^{*,j,c'}_t) \|_2 
        + \| \Delta_t^{c'}(\theta_t^c, w^{*,j,c'}_t, \theta_t^{c'}) - g_t^{j,c'}(w^{*,j,c'}_t)\|_2  +
        \| g_t^{j,c'}(w_t^{*,j,c'}) - g_t^{j,c'}\|_2  
    \end{aligned}
\end{equation}
Now applying  Lemmas~\ref{lemma:T113_bound}, \ref{lemma:T111_bound}, and \ref{lemma:T112_bound} on Equation~\eqref{eq:boundingClientDriftICMLWorkshop_grads_1} we will have:

\begin{equation}
\label{eq:boundingClientDriftICMLWorkshop_grads_2}
    \begin{aligned}
         E\Big[\| g_t^{j,c} - g_t^{j,c'} \|_2 \mid \mathcal{F}_t\Big] &\leq  4 \sqrt{\zeta_{\text{approx}}} + 2 \sqrt{E\Big[ \|w_t^{*,j,c} - w_t^{j,c}\|_2 \mid \mathcal{F}_t\Big]} + 2 \sqrt{E\Big[ \|w_t^{*,j,c'} - w_t^{j,c'}\|_2 \mid \mathcal{F}_t\Big]}  +\\
         & \qquad4\sqrt{ \frac{(r_{\text{max}} + R_{\text{w}})^2 (1 - \rho + 4\kappa\rho)}{(1 - \rho)B}}  +  E\Big[\| \nabla_{\theta}J^{c,j}(\theta_{t}^c) - \nabla_{\theta}J^{c',j}(\theta_{t}^{c})\|_2 \mid \mathcal{F}_t \Big]+\\
         & \qquad \qquad \qquad
         \qquad \qquad \qquad
         \qquad \qquad 
          E\Big[\| \nabla_{\theta}J^{c,j}(\theta_{t}^{c'}) - \nabla_{\theta}J^{c',j}(\theta_{t}^{c'})\|_2 \mid \mathcal{F}_t \Big]
    \end{aligned}
\end{equation}
Note that the terms \(\sqrt{E\Big[ \|w_t^{*,c'} - w_t^{c'}\|_2 \mid \mathcal{F}_t\Big]}\) and \(\sqrt{E\Big[ \|w_t^{*,c} - w_t^{c}\|_2 \mid \mathcal{F}_t\Big]} \) represents the critic estimation error of each client and can be bounded by proper batch size of the critics. Now,  in order to bound the term \(E\Big[\| \nabla_{\theta}J(\theta_{t}^c) - \nabla_{\theta}J(\theta_{t}^{c'})\|_2 \mid \mathcal{F}_t \Big]\) in Equation(\ref{eq:boundingClientDriftICMLWorkshop_grads_2}) we apply Proposition 1 from \cite{xu2020improving} and will get the following:

\begin{equation}
\label{eq:boundingClientDriftICMLWorkshop_grads_3}
    \begin{aligned}
         E\Big[\| g_t^{j,c} - g_t^{j,c'} \|_2 \mid \mathcal{F}_t\Big] &\leq  4 \sqrt{\zeta_{\text{approx}}} + 2 \sqrt{E\Big[ \|w_t^{*,j,c} - w_t^{j,c}\|_2 \mid \mathcal{F}_t\Big]} + 2 \sqrt{E\Big[ \|w_t^{*,j,c'} - w_t^{j,c'}\|_2 \mid \mathcal{F}_t\Big]}  \\
         & \qquad+  4\sqrt{ \frac{(r_{\text{max}} + R_{\text{w}})^2 (1 - \rho + 4\kappa\rho)}{(1 - \rho)B}} + L_J E\Big[\| \theta_{t}^c - \theta_{t}^{c'} \|_2 \mid \mathcal{F}_t\Big] + 2  \zeta
    \end{aligned}
\end{equation}

where \(L_J = \frac{ r_{\text{max}}}{1-\gamma}\, (4 C_\nu C_\psi + L_\psi) \) and \(C_\nu = \left(\frac{1}{2}\right) C_\pi \left(1 + \lceil \log_\rho \kappa^{-1} \rceil\right) + (1 - \rho)^{-1}\). Now, since we have  \(\theta_t^c = \theta_{t_0}^c + \sum_{\tau = 0}^{t-t_0} \sum_{j=1}^M\lambda^{j,c}_{t_0+\tau} \; g_{t_0+\tau}^{j,c}\), we will have:

\begin{equation}
\label{eq:boundingClientDriftICMLWorkshop_grads_3_1}
    \begin{aligned}
       & E\Big[\| \theta_{t}^c - \theta_{t}^{c'} \|_2 \mid \mathcal{F}_t\Big]
       \leq  E\Big[\| \theta_{t_0}^c + \alpha \sum_{\tau = 0}^{t-t_0} \sum_{j=1}^M\lambda^{j,c}_{t_0+\tau} \; g_{t_0+\tau}^{j,c}  - \theta_{t_0}^{c'} -  \alpha \sum_{\tau = 0}^{t-t_0} \sum_{j=1}^M\lambda^{j,c'}_{t_0+\tau} \; g_{t_0+\tau}^{j,c'}\|_2 \mid \mathcal{F}_t\Big]
    \end{aligned}
\end{equation}
now, since at \(t_0\) we have \(\theta_{t_0}^c = \theta_{t_0}^{c'}\), and applying triangular ineqality we will have:

\begin{equation}
\label{eq:boundingClientDriftICMLWorkshop_grads_3_2}
    \begin{aligned}
       & E\Big[\| \theta_{t}^c - \theta_{t}^{c'} \|_2 \mid \mathcal{F}_t\Big]
       \leq \alpha \sum_{\tau = 0}^{t-t_0} E\Big[\|    \sum_{j=1}^M\lambda^{j,c}_{t_0+\tau} \; g_{t_0+\tau}^{j,c} \; -  \;\lambda^{j,c'}_{t_0+\tau} \; g_{t_0+\tau}^{j,c'}\|_2 \mid \mathcal{F}_t\Big]\\
       & \qquad \leq 2 \alpha \; \sum_{\tau = 0}^{t-t_0} E\Big[\|    \sum_{j=1}^M\lambda^{j,c}_{t_0+\tau} \; g_{t_0+\tau}^{j,c} \|_2
       \; + \| \sum_{j=1}^M \lambda^{j,c'}_{t_0+\tau} \; g_{t_0+\tau}^{j,c'}\|_2 \mid \mathcal{F}_t\Big]
    \end{aligned}
\end{equation}

Now, using the fact that \(\|g_t^{j,c}\|_2 \leq R, \forall t\in[T]\), and  \(\lambda_t^c \in \Delta(R^d)\), we will have:

\begin{equation}
\label{eq:boundingClientDriftICMLWorkshop_grads_3_3}
    \begin{aligned}
       & E\Big[\| \theta_{t}^c - \theta_{t}^{c'} \|_2 \mid \mathcal{F}_t\Big]
       \ \leq 4  R \alpha\, (t-t_0)
    \end{aligned}
\end{equation}

Thus, using Equation(\ref{eq:boundingClientDriftICMLWorkshop_grads_3_3}), we can continue Equation(\ref{eq:boundingClientDriftICMLWorkshop_grads_3}) as follows:

\begin{equation}
\label{eq:boundingClientDriftICMLWorkshop_grads_4}
    \begin{aligned}
         E\Big[\| g_t^{j,c} - g_t^{j,c'} \|_2 \mid \mathcal{F}_t\Big] &\leq  4 \sqrt{\zeta_{\text{approx}}} + 2 \sqrt{E\Big[ \|w_t^{*,j,c} - w_t^{j,c}\|_2 \mid \mathcal{F}_t\Big]} + 2 \sqrt{E\Big[ \|w_t^{*,j,c'} - w_t^{j,c'}\|_2 \mid \mathcal{F}_t\Big]}  \\
         & \qquad+  4\sqrt{ \frac{(r_{\text{max}} + R_{\text{w}})^2 (1 - \rho + 4\kappa\rho)}{(1 - \rho)B}} +4  L_J  R \, \alpha\, (t-t_0) + 2 \zeta \qquad \square
    \end{aligned}
\end{equation}

\section{Preference-Weighted MGDA Subproblem}
\label{appendix:mgda-preference}

To incorporate explicit user preferences, we generalize the local MGDA subproblem by 
replacing the uniform regularizer $\tfrac{\beta}{2} I$ in Equation~\eqref{eq:mgda-regularized-scaled} 
with a diagonal weighting matrix $\text{Diag}(\mathbf{p}^{-1})$, where 
$\mathbf{p} = [p_{1}, \ldots, p_{M}]$ is a vector of positive preference weights. 
The full expression becomes:
\begin{equation}
    \lambda^{*} \in \arg\min_{\lambda \in \Delta_{M}} 
    \lambda^{\top} \left( \widehat{G} + \text{Diag}(\mathbf{p}^{-1}) \right) \lambda ,
    \label{eq:mgda-preference_appendix}
\end{equation}
where $\Delta_{M}$ denotes the probability simplex. A larger preference $p_{j}$ reduces its 
penalty term $1/p_{j}$, encouraging the optimizer to assign greater weight $\lambda_{j}$ 
to the corresponding objective and bias the descent direction accordingly.

\section{Derivation of Gradient Heterogeneity Bound}
\label{App:HeterogeneityDerivation}

In Assumption~\ref{Assump:Heterogeneity}, we introduced the constant $\zeta$ to bound the deviation between the local objective gradient $\nabla_\theta J^c(\theta)$ and the global objective gradient $\nabla_\theta \mathbf{J}(\theta)$. In this section, we formally derive how $\zeta$ depends on the fundamental sources of heterogeneity in the MDPs: the transition dynamics and the reward functions.

\paragraph{Setup.} 
Let the global environment be characterized by a transition kernel $P$ and reward function $r$. Each client $c$ possesses a local environment with transition kernel $P_c$ and reward $r_c$. We define the heterogeneity in dynamics and rewards as follows:
\begin{align*}
    \sup_{s,a} \| P_c(\cdot|s,a) - P(\cdot|s,a) \|_{TV} &\le \epsilon_p, \\
    \sup_{s,a} | r_c(s,a) - r(s,a) | &\le \epsilon_r.
\end{align*}

\begin{proposition}
Under Assumptions~\ref{Assumption:FuncApproxMOMDP_1}-\ref{Assump:Smooth_bounded_AISTAT_1}, the gradient heterogeneity is bounded by:
\[
\| \nabla_\theta J^c(\theta) - \nabla_\theta J(\theta) \|_2 \le C_\psi \left( \frac{\epsilon_r}{1-\gamma} + \frac{r_{\max} \gamma \epsilon_p}{(1-\gamma)^2} \right) \triangleq \zeta.
\]
Thus, $\zeta = \mathcal{O}(\epsilon_r + \epsilon_p)$.
\end{proposition}

\begin{proof}
By the Policy Gradient Theorem, the gradient for client $c$ is:
\[
\nabla_\theta J^c(\theta) = \mathbb{E}_{s \sim d_{\pi, P_c}, a \sim \pi_\theta} [ \nabla_\theta \log \pi_\theta(a|s) Q_c^{\pi}(s,a) ],
\]
where $d_{\pi, P_c}$ is the stationary distribution induced by $\pi$ on $P_c$, and $Q_c^{\pi}$ is the action-value function for client $c$. The difference can be decomposed as:
\begin{align*}
    \| \nabla_\theta J^c - \nabla_\theta J \|_2 &= \| \mathbb{E}_{d_c} [\psi_\theta Q_c] - \mathbb{E}_{d} [\psi_\theta Q] \|_2 \\
    &\le \| \mathbb{E}_{d_c} [\psi_\theta Q_c] - \mathbb{E}_{d_c} [\psi_\theta Q] \|_2 + \| \mathbb{E}_{d_c} [\psi_\theta Q] - \mathbb{E}_{d} [\psi_\theta Q] \|_2 \\
    &\le \mathbb{E}_{d_c} [ \| \psi_\theta \|_2 | Q_c - Q | ] + \| \psi_\theta \|_\infty \| Q \|_\infty \| d_c - d \|_{TV}.
\end{align*}
Using the bound $\|\psi_\theta\|_2 \le C_\psi$ (Assumption 3a), we analyze the two error terms:

\textbf{1. Value Function Difference ($|Q_c - Q|$):}
Using the simulation lemma~\citep{agarwal2019reinforcement}, for any $(s,a)$:
\[
| Q_c^\pi(s,a) - Q^\pi(s,a) | \le \frac{\epsilon_r}{1-\gamma} + \frac{\gamma r_{\max} \epsilon_p}{(1-\gamma)^2}.
\]

\textbf{2. Distribution Difference ($\|d_c - d\|_{TV}$):}
Standard perturbation bounds for Markov chains \citep{xu2020improving} yield:
\[
\| d_{\pi, P_c} - d_{\pi, P} \|_{TV} \le \frac{\gamma \epsilon_p}{1-\gamma}.
\]

Substituting these back:
\begin{align*}
    \| \nabla_\theta J^c - \nabla_\theta J \|_2 &\le C_\psi \left( \frac{\epsilon_r}{1-\gamma} + \frac{\gamma r_{\max} \epsilon_p}{(1-\gamma)^2} \right) + C_\psi \left( \frac{r_{\max}}{1-\gamma} \right) \left( \frac{\gamma \epsilon_p}{1-\gamma} \right) \\
    &= C_\psi \left( \frac{\epsilon_r}{1-\gamma} + \frac{2 \gamma r_{\max} \epsilon_p}{(1-\gamma)^2} \right).
\end{align*}
This confirms that the gradient deviation $\zeta$ scales linearly with the environmental heterogeneity terms $\epsilon_r$ and $\epsilon_p$.
\end{proof}

\end{document}